\documentclass[pdflatex,referee,sn-basic]{sn-jnl}

\usepackage{graphicx}%
\usepackage{multirow}%
\usepackage{amsmath,amssymb,amsfonts}%
\usepackage{amsthm}%
\usepackage{mathrsfs}%
\usepackage[title]{appendix}%
\usepackage{xcolor}%
\usepackage{textcomp}%
\usepackage{manyfoot}%
\usepackage{booktabs}%
\usepackage{algorithm}%
\usepackage{algorithmicx}%
\usepackage{placeins}%

\usepackage{algpseudocode}
\algnewcommand\algorithmicforeach{\textbf{for each}}
\algdef{S}[FOR]{ForEach}[1]{\algorithmicforeach\ #1\ \algorithmicdo}
\algnewcommand{\LineComment}[1]{\State \(\triangleright\) #1}

\usepackage{lscape}

\theoremstyle{thmstyleone}%
\theoremstyle{thmstyletwo}%

\theoremstyle{thmstylethree}%

\usepackage{siunitx}

\usepackage{caption}

\begin{document}

\title[Accelerating FastDEC Clustering with GPUs]{Accelerating FastDEC Clustering Algorithm with GPUs}

\author*{\fnm{Mi} \sur{Li}}\email{ml87@students.waikato.ac.nz}

\author{\fnm{Eibe} \sur{Frank}}\email{eibe.frank@waikato.ac.nz}

\author{\fnm{Bernhard} \sur{Pfahringer}}\email{bernhard.pfahringer@waikato.ac.nz}

\affil{\orgdiv{Department of Computer Science}, \orgname{University of Waikato}, \city{Hamilton}, \country{New Zealand}}

\abstract{
Fast Dominance Estimation Clustering (FastDEC) is an efficient variant of Density Peak Clustering (DPC), offering high clustering quality through dominance-based density estimation, while still being constrained by the computational cost of both exact $k$-nearest-neighbor ($k$-NN) construction and dominance component (DC) processing. In this paper, we present \textit{SymWarp-FastDEC}, a graphics processing unit (GPU)-accelerated implementation of FastDEC. The proposed framework employs a set of GPU-specialized optimizations. At its core is \textit{SymTri-$k$NN}, an exact self-$k$-NN kernel that exploits the symmetry of the search setting through upper-triangular traversal and warp-cooperative top-$k$ selection. In addition, we introduce \textit{WP-DC}, a warp-parallel dominance component (DC) reduction algorithm specifically designed to exploit the architectural characteristics of modern GPUs. Combined with a memory-adaptive execution planner and stream scheduling strategy, WP-DC effectively eliminates DC reduction as a runtime bottleneck. 

We evaluate SymWarp-FastDEC on synthetic datasets and a diverse collection of real-world benchmarks. Experimental results demonstrate substantial end-to-end acceleration, and on all evaluated equivalence benchmarks SymWarp-FastDEC produces clustering outputs identical to the original FastDEC implementation. Relative to the FastDEC variant employing a widely used, highly optimized exact GPU $k$-NN baseline, FAISS GpuIndexFlatL2, our approach delivers approximately $1.7\times$ to $4.9\times$ higher end-to-end performance, with both performing exhaustive (exact) $k$-NN search and the advantage generally increasing as data dimensionality grows. Relative to the original single-threaded Python FastDEC reference implementation, speedups reach up to approximately $2.5\times10^4$; this extreme figure arises in a very high-dimensional setting where that baseline's $k$-d-tree-based search degrades, and is not representative of the typical acceleration. 
}




\maketitle

\section{Introduction}
\label{sec:intro}

The proliferation of big data and the increasing complexity of computational tasks have created a growing demand for efficient and scalable data analysis algorithms. Clustering, a fundamental technique in data mining and machine learning, aims to group similar data objects into meaningful clusters based on their intrinsic characteristics~\citep{fahad2014survey,sajana2016survey,bhattacharjee2021survey,yin2026rapid}. Among the various clustering paradigms, density-based methods have attracted considerable attention because of their ability to identify clusters with arbitrary shapes while effectively handling noise and outliers~\citep{yin2026rapid}. Density Peak Clustering (DPC), introduced by Rodriguez and Laio in 2014, is a representative density-based clustering algorithm that identifies cluster centers as data points exhibiting high local density and large distances from other points with higher densities~\citep{rodriguez2014clustering}. It is suitable for a wide range of real-world applications~\citep{wei2023overview} but is computationally expensive. The calculation of local densities and pairwise distances limits scalability as dataset size increases~\citep{wang2024density}. 

Fast Dominance Estimation Clustering (FastDEC)~\citep{yang2022fastdec} advances density peak clustering by integrating k-Nearest Neighbors (k-NN) graphs with dominance estimation to form dominance components (DC), significantly improving the efficiency of the clustering process. Nevertheless, opportunities for further improvements remain, particularly by exploiting parallel processing.
Parallel computing has proven to be a practical solution for handling big data~\citep{fang2008parallel,langdon2013large,shirkhorshidi2014big,xu2024faithful}. Traditionally, supercomputers were required for high-performance parallel computing to solve complex scientific problems~\citep{upadhyaya2013parallel} but modern graphics processing units (GPUs) offer an alternative for general-purpose high-performance computing~\citep{owens2008gpu}. 
In particular, they have been used to significantly accelerate algorithms for clustering problems~\citep{li2022review}---for example, by applying the general-purpose CUDA programming model on NVIDIA GPU architectures~\citep{li2016paralleled}. In this paper, we show that FastDEC's performance can be further enhanced by leveraging the parallel processing capabilities of GPUs.

Our primary contributions are summarized as follows: 
\begin{enumerate} 
    \item To the best of our knowledge, we introduce the first GPU-accelerated FastDEC framework capable of scaling to large datasets, high-dimensional feature spaces, and a large number of cluster centers. 
    \item We propose a high-performance GPU-based $k$-NN algorithm specifically designed for FastDEC. Leveraging query-reference symmetry via tiled upper-triangular traversal, the method achieves efficient nearest-neighbor computation on large-scale, high-dimensional datasets. 
    \item We propose an efficient index-free algorithm for Dominance Component (DC) Inter-Dominance Estimation and Reduction on GPUs. Through warp-parallel reduction, memory-adaptive slab planning, and streamed ping-pong execution, the algorithm eliminates the nearest-neighbor constraints imposed by most existing GPU-based $k$-NN methods. 
    \item We decouple DC Inter-Dominance Estimation and Reduction from the spatial indexing structures employed during Density Estimation, thereby enhancing the flexibility, scalability, and general applicability of the FastDEC framework. 
\end{enumerate}

The remainder of this paper is structured as follows. Section 2 provides background on GPU architectures and reviews related work on Density Peak Clustering (DPC) and GPU-accelerated DPC methods. Section 3 introduces our GPU-accelerated FastDEC framework and describes its design, implementation, parallelization strategies, and optimization techniques. Section 4 presents a comprehensive experimental evaluation across diverse datasets and benchmarks the proposed method against FastDEC variants employing similarity-search algorithms from well-established libraries. Finally, Section 5 summarizes the main findings and outlines avenues for future work.

\section{Background and Related Work}
\label{sec:relwork}
First, we briefly review essential concepts from general-purpose computation on GPUs. Subsequently, we review research addressing (a) Density Peak Clustering and its variants, with an emphasis on the FastDEC algorithm, and (b) GPU-based implementations of DPC and its variants. 

\subsection{GPGPU}
\label{sec:gpgpu}

From a hardware perspective, and using NVIDIA terminology, a GPU consists of an array of streaming multiprocessors (SMs). Each SM contains multiple execution cores. These cores execute threads according to a Single-Instruction Multiple-Thread (SIMT) execution model, in which groups of threads typically execute the same instruction stream while operating on different data elements. Each SM possesses its own register file and a block of shared memory that is accessible only to the threads executing on that SM. In GPGPU programming, three principal memory types are commonly utilised: registers, shared memory, and global memory (also known as device memory). Registers are the fastest form of on-chip storage available to a thread, but the available register space is extremely limited. Shared memory is an on-chip memory resource that offers significantly lower latency and higher bandwidth than global memory. Shared memory typically ranges from tens to hundreds of kilobytes per SM depending on the GPU architecture. In contrast, device memory resides physically on the graphics card but outside the SMs themselves. It is referred to as global memory because it is accessible to all threads executing on the GPU. Device memory may also be accessed from the host through CUDA-managed memory mechanisms or explicit data-transfer operations. Because it is located off-chip, access latency may require hundreds of clock cycles, making it significantly slower than register or shared-memory access. Effective GPU programming therefore requires techniques that minimise or hide memory latency. Common approaches include tiled data loading from device memory into shared memory and pipelined execution strategies \citep{cuda2026guide}.

GPU threads are lightweight and incur minimal creation and context-switching overhead. Thousands of threads can be executed concurrently to help hide memory access latency. Threads are organised into groups known as warps, whose threads execute the same instruction stream on different data elements. Warps are further grouped into thread blocks. Each thread block is guaranteed to reside entirely on a single SM, enabling threads within the block to cooperate through shared memory and synchronisation primitives. In contrast, communication and synchronisation between different thread blocks are limited. Finally, thread blocks are organised into a computational grid, which is processed collectively by the GPU's array of streaming multiprocessors.

\subsection{Density Peak Clustering (DPC)}
\label{sec:dpc}

Density Peak Clustering (DPC) is a density-based clustering algorithm proposed by Rodriguez and Laio in 2014 in their seminal Science paper, Clustering by Fast Search and Find of Density Peaks (FSFDP) \citep{rodriguez2014clustering}. The fundamental idea of DPC is to identify cluster centers as data points that exhibit both high local density and a relatively large distance from any other point of higher density. These points, referred to as density peaks, are assumed to represent the centers of the underlying clusters. Similar to other density-based clustering algorithms, such as DBSCAN and MeanShift, DPC can identify clusters with arbitrary shapes by estimating the density distribution of data points and exploiting the connectivity of high-density regions. 

DPC and its numerous variants have been successfully applied across a wide range of domains. In bioinformatics, DPC has been employed for gene expression analysis, protein classification, and biological network analysis. In image processing and computer vision, it has been used for image segmentation, feature extraction, and object recognition. Applications have also been reported in social network analysis, anomaly detection, market basket analysis, and other data mining tasks \citep{guan2021peak,li2022review,wei2023overview,wang2024density,ren2025network,wang2025improved}. 

The FSFDP algorithm for DPC follows three key steps:
\begin{enumerate}
    \item Density Estimation: For each point $i$, the local density $\rho_i$ is estimated using a cutoff distance $d_c$, where all neighbors within $d_c$ contribute to the density.
    \item Distance to Higher Density: For each point $i$, the distance $\delta_i$
 to the nearest point of higher density is computed. Points with both large $\rho$ and large $\delta$ are identified as cluster centers, a.k.a. density peaks.
    \item Cluster Assignment: Remaining points are assigned to the same cluster as their nearest denser neighbor, ensuring propagation of cluster membership.
\end{enumerate}

By combining local density and relative distance to identify density peaks, FSFDP provides a mechanism for identifying cluster centers without prior knowledge of the number of clusters. It inspired many subsequent advancements in density peak clustering, including FastDEC~\citep{yang2022fastdec}, FastDPeak~\citep{chen2020fast}, and the GPU-accelerated FSFDP~\citep{li2016paralleled}.

\subsection{Variants of the DPC Algorithm}
\label{sec:dpc-variants}

FSFDP, like other well-known density-based clustering algorithms, suffers from high computational complexity, primarily due to the pairwise distance calculations it requires~\citep{wang2024density}. In particular, it relies on brute-force search to find the nearest denser (ND) object for each data point, making it impractical for handling large datasets.
Several variants of DPC have been proposed that aim to improve its efficiency and scalability. They explore different strategies, including: (1) algorithmic optimizations; (2) hardware acceleration through parallelization, and (3) data volume reduction. Below, we survey representative publications with a focus on those that do not rely on approximations and thus do not cover the algorithms in the third category~\cite{amagata2021fast,ma2023ultra,ma2024up}. The discussion of FastDEC, the algorithm we aim to improve in this paper, is deferred to a separate subsection.

\subsubsection{Algorithmic optimizations}
\label{sec:algorithmic-optimisations}

Algorithmic optimizations employ efficient data structures and algorithms for density estimation, local density peak identification, and cluster formation. They primarily apply modifications along two axes: (1) in the method for finding density peaks, and (2) in the method for combining local density peaks into final clusters.

\citet{liu2018shared} propose SNN-DPC (Shared Nearest Neighbor DPC) to improve scalability by constructing a Shared Nearest Neighbor (SNN) graph from a k-NN graph. The edges are reweighted based on the number of shared neighbors, and DPC is applied to this compact graph.

\citet{chen2020fast} propose FastDPeak to accelerate DPC for large-scale data, replacing traditional cutoff-based density estimation with kNN-density, which can be computed more efficiently using spatial indexing (e.g., cover trees). This reduces computational complexity from $O(N^2)$ to $O(N \log N)$. The algorithm distinguishes local and non-local density peaks, applying tailored strategies to compute dependent distances ($\delta$): triangle inequality optimizations for local peaks and neighbor-based searches for non-local peaks. 

\citet{rasool2020index} introduce an index-based approach, named Index-DPC, to accelerate density and distance computations. Several lists buffer object-wise significance information, and an efficient query algorithm avoids redundant comparisons.

\citet{chena2021improved} propose C-DP to solve the problem of traditional DPC having difficulties to determine the correct number of clusters when dealing with datasets containing a single cluster with multiple density peaks. They apply the CURE (Clustering Using Representatives) algorithm on the density peak points identified by the traditional DPC. It iteratively combines sub-clusters with the closest representative point pairs using a Contraction Factor ($\alpha$) to control cluster shape during CURE merging. C-DP requires tuning of the contraction factor $\alpha$ and the merging strategy also adds computational overhead. 

\citet{zheng2021quickdsc} proposed QuickDSC (Quick Density Subgraph Clustering), which partitions the input k-NN graph into subgraphs based on mutual k-NN connections and then applies DPC to each subgraph, thereby reducing computational costs. Subsequently, \citet{jia2024adaptive} introduced the Adaptive Density Subgraph Clustering (ADSC) algorithm to overcome the limitations of QuickDSC, such as constraints on cluster size arising from the use of density subgraphs, which may result in an excessive number of small clusters and hinder accurate clustering, particularly when the true number of clusters is unknown.

\subsubsection{Hardware Acceleration}
\label{sec:hardware-acceleration}

Numerous studies have demonstrated the effectiveness of GPU acceleration in data clustering applications. For example, G-DBSCAN, a GPU-accelerated implementation of the density-based clustering algorithm DBSCAN, achieves significant performance improvements over conventional CPU-based implementations \citep{andrade2013g}. Similarly, GPU-accelerated implementations of the k-means algorithm have demonstrated substantial reductions in execution time compared with CPU implementations \citep{li2023large}.

\citet{li2016paralleled} accelerate the DPC algorithm using CUDA (Paralleled FSFDP), optimizing thread/block models and shared memory usage to maximize GPU resource utilization. The authors use CUDA to parallelize the computation of distance matrices and potentials, achieving 7–15.75× speedup compared to CPU implementations. 

\citet{ge2017efficient} speed up the DPC algorithm by accelerating all-pairs distance computation via massive GPU parallelism (CUDA-DP). It uses a structure-of-arrays data layout for coalesced memory access and parallel reduction for local density ($\rho$) and dependent distance ($\delta$) calculations. Experimental results show that it achieves 7-45x speedup vs. single-threaded CPU-based DPC on mid-sized datasets of 10k-100k points. Although it scales poorly on datasets with more than 500k points, CUDA-DP, together with Paralleled FSFDP above, demonstrates the feasibility of GPU-based DPC acceleration.

\citet{su2020gdpc} present GPU-Accelerated Density Peaks Clustering (GDPC) to accelerate DPC. The core idea here is leveraging GPU parallelism to speed up the computationally intensive parts of the DPC algorithm. It relies on a spatial index structure called vectorized VP-tree to enable coalesced memory access, with nodes stored in GPU memory. Redundant distance calculations are reduced via VP-tree pruning. CUDA streams are used for overlapped VP-tree construction and distance computations. It also parallelizes the query process, using warp-centric work assignment to minimize divergence and ballot-counting optimization for efficient density computation within cutoff distance $d_c$. However, the VP-tree construction may become a bottleneck for high-dimensional data ($>1k$ dimensions), and the cutoff distance $d_c$ also requires tuning.

Multicore CPU computing provides an alternative approach to parallel clustering algorithms. \citet{amagata2021fast} propose several algorithms (Ex-DPC, Approx-DPC, S-Approx-DPC) in this category. Ex-DPC is an exact DPC accelerated by multicore parallelism. It uses an in-memory $k$-d tree to accelerate local density ($\rho_i$) computation via range searches and dependent distance ($\delta_i$) computation via incremental nearest-neighbor searches. The local density computation is parallelized across CPU cores using dynamic load balancing. Approximation strategies are applied to Approx-DPC and S-Approx-DPC: Approx-DPC uses grid-based joint range searches for coalesced memory access, and S-Approx-DPC employs grid sampling to trade precision for speed via cell-level clustering. Similar to GDPC, relying on a tree-based structure limits the exact algorithm's applicability to high-dimensional data.

\subsection{Fast Dominance Estimation Clustering (FastDEC)}
\label{sec:fastdec}

FastDEC leverages fast dominance estimation on $k$-Nearest Neighbors ($k$-NN) graphs~\citep{yang2022fastdec} to improve the efficiency and robustness of DPC. The core concept of FastDEC revolves around dominance, where a point is considered to dominate another if it has a higher density and is within a certain distance threshold. By estimating dominance relationships, FastDEC effectively identifies cluster centers and assigns points to the nearest dominant center, forming clusters. Experimental results in both artificial and real-world datasets demonstrate that FastDEC consistently achieves top-tier results in clustering quality metrics (ARI and NMI). It outperforms or matches state-of-the-art methods such as Quickshift++, FINCH, QuickDSC, and SNN-DPC, all while maintaining competitive computational runtime.

Traditional DPC algorithms, such as the one proposed by Rodriguez and Laio~\citep{rodriguez2014clustering}, estimate the significance of data points using density and geometric distance factors. However, these algorithms often suffer from high computational complexity and require tuning multiple parameters. FastDEC addresses these challenges by integrating the nearest density-higher (NDH) relation with the $k$-NN graph to form dominance-components (DCs). The algorithm introduces the concept of direct $k$-NN dominators (DkD) and uses a novel significance estimation that incorporates the global rank of the dominator, in addition to density and geometric distance. It offers several distinct advantages:
\begin{enumerate}
    \item Efficiency: It uses $k$-NN graphs to accelerate the identification of dominance components (LDPs - local Density Peaks), reducing computational complexity to near-linear in $N$ and $k$ via space indexing, avoiding the quadratic time complexity of traditional DPC.
    \item Simplicity: It requires only one parameter, the number of nearest neighbors ($k$), which controls neighborhood granularity, simplifying configuration and tuning. It replaces DPC's fixed---and often difficult to choose---radius-based density estimation with kernel-based density estimation over the $k$ nearest neighbors.
    \item Robustness: Its clustering quality remains superior to competing methods across a wide range of $k$ values ($10 \le k \le 200$)~\citep{yang2022fastdec}, and it is more robust to noise due to reliance on $k$-NN graphs and dominance estimation.
    \item Interpretability: It introduces dominance estimation, evaluating the significance of data points using the global rank of the dominator, in addition to density and geometric distance.
\end{enumerate}

The three main steps of the FastDEC algorithm can be broadly described as follows:
\begin{enumerate}
    \item Density Estimation: For each data point, the algorithm estimates its local density using a fast, $k$-NN-based density estimation technique, which can employ different kernel functions.
    \item Dominance Estimation: The algorithm then estimates the dominance relationships between points. A point $p$ is considered to dominate another point $q$ if $p$ has a higher density and is within the distance threshold of $q$.
    \item Cluster Formation: Based on the dominance relationships, the algorithm identifies cluster centers as points that are not dominated by any other point. This step relies on the global ranking factor of DCs to determine inter-relationships between DCs, so that DCs can be reduced to form the cluster centers. Each data point is then assigned to the nearest dominant center, forming clusters.
\end{enumerate}

We now consider the steps of the algorithm in detail. Algorithm \ref{alg:fastdec} outlines the overall framework. Lines 1 and 2 estimate the density of each data point using a $k$-NN-based density estimator, as presented in Algorithm \ref{alg:density-estimation} and explained in detail in Section \ref{sub:density-estimation}. Direct $k$-NN Dominator (DkD) detection, shown in Line 3, is described in Algorithm \ref{alg:dominance-detection}. The extraction of Dominance Components (DCs), performed in Lines 10 and 11, is presented in Algorithm \ref{alg:dc-reduction}. Finally, Line 12 assigns all remaining data points to the identified cluster centers using Algorithm \ref{alg:cluster-formation}, thereby completing the clustering process.

\begin{algorithm}[t]
\caption{FastDEC}
\label{alg:fastdec}
\begin{algorithmic}[1]
\Require Dataset $X$, initial neighborhood size $k$, target number of clusters $c$
\Ensure Cluster labels $y$

\State Build space index for $X$ and construct the $k$-NN graph for each point
\State Estimate the local density of each point from its $k$-NN distances
\State Identify dominance links and extract the initial DCs

\If{the number of clusters is not specified}
    \State Treat all DCs as cluster centers
    \State Assign labels by dominance-based cluster formation
    \State \Return $y$
\EndIf

\State Adjust the neighborhood size, if necessary, until at least $c$ DCs are obtained
\State If more than $c$ DCs are detected, reduce them by inter-dominance significance
\State Select the final $c$ DCs as cluster centers
\State Assign all remaining points by following closest density-higher (CDH) links
\State \Return $y$

\end{algorithmic}
\end{algorithm}

\subsubsection{Density Estimation}
\label{sub:density-estimation}

FastDEC estimates local density by first constructing a $k$-nearest-neighbor ($k$-NN) graph for the input dataset. Let $X=\{x_1,x_2,\ldots,x_N\}$ denote a dataset containing $N$ samples. For each sample $x_i$, the algorithm searches for its $k$ nearest distinct neighbors under the selected distance metric, where distinct means a different index $j \neq i$; points with coordinates identical to $x_i$ but a different index remain eligible neighbors. In the reference implementation of Yang et al., the $k$-NN search is performed using an exact $k$-d tree by default, although FastDEC itself only requires a $k$-NN graph and is not tied to a particular nearest-neighbor search method. The query retrieves $k+1$ neighbors, including the point itself, after which the self-match is removed to obtain the $k$ nearest distinct neighbors.

The $k$-NN search produces two matrices: a neighbor-index matrix $I \in \mathbb{N}^{N \times k}$ and a neighbor-distance matrix $\Delta \in \mathbb{R}^{N \times k}$. Each entry $I_{ij}$ stores the index of the $j$-th nearest neighbor of $x_i$, and each entry $\Delta_{ij}=d_{ij}$ stores the corresponding distance.

The local density of each point is then estimated from its $k$-NN distances using an exponential kernel density estimator\footnote{The original FastDEC paper describes a Gaussian kernel density estimator as the default density estimator. However, the reference implementation used in our experiments employs the exponential kernel density estimator defined below. To ensure fidelity to the published source code and reproducibility of results, we follow the implementation rather than the textual description in the original paper.} with a globally determined bandwidth based on the average nearest-neighbor distance. Specifically, FastDEC first computes the global mean nearest-neighbor distance over all points and their retrieved neighbors as

\[
\bar{d}
=
\frac{1}{Nk}
\sum_{i=1}^{N}
\sum_{j=1}^{k}
d_{ij}.
\]

The density of point $x_i$ is then defined as

\[
\rho_i
=
\sum_{j=1}^{k}
\exp\!\left(
-\frac{d_{ij}}{2\bar{d}}
\right).
\]

This estimator assigns high density to points located in compact neighborhoods, since small neighbor distances contribute larger exponential weights. Conversely, points in sparse regions or near cluster boundaries tend to have larger $k$-NN distances and therefore receive smaller density values.

The resulting density vector $\rho=(\rho_1,\rho_2,\ldots,\rho_N)$ induces the density ordering used in the subsequent dominance detection stage, where each point is compared with its neighbors to identify nearby samples of higher density.

\begin{algorithm}[t]
\caption{Density Estimation}
\label{alg:density-estimation}
\begin{algorithmic}[1]
\Require Dataset $X=\{x_1,\ldots,x_N\}$, neighborhood size $k$, metric $d(\cdot,\cdot)$
\Ensure Density vector $\rho=(\rho_1,\ldots,\rho_N)$ and $k$-NN graph $\mathcal{N}_k$

\State Construct the $k$-NN graph $\mathcal{N}_k$ and obtain the neighbor-index matrix $I=[I_{ij}]\in\mathbb{N}^{N\times k}$ and neighbor-distance matrix $\Delta=[\Delta_{ij}]\in\mathbb{R}^{N\times k}$; let $\Delta_{ij}=d_{ij}$ denote the distance from $x_i$ to its $j$-th nearest neighbor.
\State Compute the mean $k$-NN distance $\bar{d}=\frac{1}{Nk}\sum_{i=1}^{N}\sum_{j=1}^{k} d_{ij}$.
\State For each point $x_i$, compute $\rho_i=\sum_{j=1}^{k}\exp\!\left(-\frac{d_{ij}}{2\bar{d}}\right)$.
\State \Return $\rho$ and $\mathcal{N}_k$
\end{algorithmic}
\end{algorithm}

\subsubsection{Dominance Estimation}
\label{sub:dominance-estimation}

Dominance estimation, also known as Dominance Component (DC) detection, identifies points that are locally dominated by denser neighboring points. Let $\rho_i$ denote the density of point $x_i$, and let $\mathcal{N}_k(x_i)$ be the ordered set of its $k$ nearest neighbors, sorted by increasing distance from $x_i$. For each point $x_i$, FastDEC scans $\mathcal{N}_k(x_i)$ and selects the first neighbor $x_j$ satisfying 
$\rho_j > \rho_i$. This neighbor is defined as the closest density-higher (CDH) point of $x_i$ and is denoted $\operatorname{cdh}(i) = j$.

A point without such a neighbor is a \emph{$k$-NN mode}. A dominance component (DC) is the set of points that reach a given $k$-NN mode by repeatedly following $\operatorname{cdh}$ links~\citep{yang2022fastdec}; each DC contains exactly one mode, which is its densest point. For brevity, we identify each DC with its mode throughout this paper, and the modes form the pool of candidate cluster centers. Thus, the initial DC set is defined as

\[
\mathcal{C}
=
\left\{
x_i \mid
\nexists x_j \in \mathcal{N}_k(x_i)
\text{ such that }
\rho_j > \rho_i
\right\}.
\]

This step converts the density-ordered $k$-NN graph into a directed dominance structure: every point other than a $k$-NN mode points toward a nearby higher-density sample.

An adaptive refinement step is introduced to handle the scenario where the number of detected DCs is smaller than the requested number of clusters: the neighborhood size $k$ is iteratively halved and DC detection is re-invoked on the truncated (already-computed) indexing graph until enough DCs are found. This is a cheap way of enlarging the pool of density-peak candidates without any additional nearest-neighbor search. This halving is bounded: if $k$ is reduced to $1$ and the number of detected DCs is still below the requested number of clusters, the procedure terminates and reports failure rather than halving further, since $k=0$ would be meaningless.

Because the $k$-NN neighborhood is a strict subset of the full dataset, some points that appear to be local density maxima under this truncated search are not true global peaks---they are "false DCs" produced when a cluster's true higher-density mode lies just outside the $k$-NN radius (e.g., on a plateau, saddle region, or at cluster boundaries with insufficient $k$). DC detection therefore over-approximates the true set of density peaks: a point is retained as a DC exactly when none of its own $k$ nearest neighbors has higher density. This candidate set therefore over-approximates the set of cluster centers: it may also contain spurious peaks. This over-approximation is deliberately cheap and is what allows the subsequent, more expensive step, DC reduction (also known as DC inter-dominance estimation), to perform an exhaustive ($O(N)$ per DC) dominance re-check restricted only to the small DC set---recovering the equivalent of DPC's global $\delta$/decision-graph score without paying the full $O(N^2)$ cost on the entire dataset.

\begin{algorithm}[t]
\caption{Dominance Detection}
\label{alg:dominance-detection}
\begin{algorithmic}[1]
\Require Ordered $k$-NN graph $\mathcal{N}_k$, density vector $\rho$
\Ensure Dominance links $\operatorname{cdh}$, DCs $\mathcal{C}$

\ForAll{points $x_i \in X$}
    \State Find the nearest point $x_j \in \mathcal{N}_k(x_i)$ such that $\rho_j > \rho_i$
    \If{such a point exists}
        \State Define $x_j$ as the CDH point of $x_i$: $\operatorname{cdh}(i) \gets j$
    \Else
        \State Treat $x_i$ as a DC
    \EndIf
\EndFor

\State $\mathcal{C} \gets \{x_i \mid \operatorname{cdh}(i) \text{ is undefined}\}$
\State \Return $\operatorname{cdh}$ and $\mathcal{C}$
\end{algorithmic}
\end{algorithm}

\subsubsection{DC Reduction}
\label{sub:dc-reduction}

Given the set of DCs obtained from local dominance detection, FastDEC applies a DC reduction step to select the final cluster centers. Let $\mathcal{C}=\{c_1,c_2,\ldots,c_m\}$ denote the set of DCs, where $m$ is the number of DCs identified by the local dominance step. For each DC $c_i$, the algorithm queries its complete nearest-neighbor ordering over the dataset and searches for the nearest point $x_j$ whose density exceeds that of the DC, i.e., $\rho(x_j)>\rho(c_i)$. 

This search estimates the inter-dominance relation between DCs and higher-density regions. For each DC $c_i$, two quantities are recorded: $g_i = d(c_i,x_j)$, the geometric distance from $c_i$ to its nearest higher-density point, and $r_i=\operatorname{rank}_{c_i}(x_j)$, the rank at which that higher-density point appears in the global nearest-neighbor ordering. If no higher-density point exists, the DC is treated as globally dominant and is assigned the maximum observed distance and rank values.

The algorithm then normalizes the DC density, geometric distance, and dominance rank using min-max normalization. A significance score is computed for each DC as
\[
S_i
=
\tilde{\rho}(c_i)
\cdot
\tilde{g}_i
\cdot
\tilde{r}_i,
\]
where $\tilde{\rho}(c_i)$ is the normalized density of DC $c_i$, $\tilde{g}_i$ is its normalized distance to the nearest higher-density point, and $\tilde{r}_i$ is the normalized global dominance rank. DCs with high $S_i$ are simultaneously dense, well separated from higher-density regions, and difficult to dominate in the global neighbor ordering. The DC reduction step selects the top $c$ DCs with the largest significance scores, where $c$ is the desired number of clusters:

\[
\mathcal{C}^{*}
=
\operatorname*{arg\,top\text{-}\mathit{c}}_{c_i \in \mathcal{C}}
\left(S_i\right).
\]

The selected set $\mathcal{C}^{*}$ is used as the final set of cluster centers. In addition, the nearest higher-density point found for each DC is stored as its CDH link, $\operatorname{cdh}(c_i) = j$, completing the dominance links that local dominance detection left undefined; this applies to every DC, including those not selected as centers, so that they and the points dominated by them can be labeled during cluster formation.

\begin{algorithm}[t]
\caption{DC Reduction}
\label{alg:dc-reduction}
\begin{algorithmic}[1]
\Require DCs $\mathcal{C}$, densities $\rho$, dominance links $\operatorname{cdh}$, target number of clusters $c$
\Ensure Final cluster centers $\mathcal{C}^{*}$, completed dominance links $\operatorname{cdh}$

\ForAll{$c_i \in \mathcal{C}$}
    \State Find the nearest higher-density point $x_j$ to $c_i$ in the global neighbor ordering
    \State Record its distance $g_i$ and dominance rank $r_i$
    \State $\operatorname{cdh}(c_i) \gets j$ \Comment{global CDH link; left undefined if no such $x_j$ exists}
\EndFor

\State Min-max normalize $\rho(c_i)$, $g_i$, and $r_i$
\State Score each DC by $S_i = \tilde{\rho}(c_i)\tilde{g}_i\tilde{r}_i$.
\State Select the top $c$ DCs with the largest $S_i$ as $\mathcal{C}^{*}$
\State \Return $\mathcal{C}^{*}$ and $\operatorname{cdh}$
\end{algorithmic}
\end{algorithm}

\subsection{Cluster Formation}
\label{sub:cluster-formation}

After the final cluster centers have been selected, FastDEC forms clusters by propagating labels along the dominance structure. Let $\mathcal{C}^{*} = \{c^{*}_1, c^{*}_2, \ldots, c^{*}_{c}\} \subseteq \mathcal{C}$ denote the final set of selected cluster centers, and let $\operatorname{cdh}(i)$ denote the CDH point associated with sample $x_i$ (for a DC, the global CDH link recorded during DC reduction). The algorithm first initializes all sample labels as unassigned. Each selected center $c^{*}_\ell$ is then assigned a unique cluster label $\ell$.

The remaining points are processed in descending order of density. For each unassigned point $x_i$, FastDEC assigns it the same label as its CDH point, i.e., $y_i = y_{\operatorname{cdh}(i)}$.

Because points are visited from high density to low density, the CDH point of $x_i$ has already received a label when $x_i$ is processed. This procedure defines a label propagation process over the directed dominance graph, where every non-center point inherits the cluster membership of the denser point that dominates it. Consequently, each cluster consists of a selected density center and all samples that reach that center by repeatedly following CDH links.

\begin{algorithm}[t]
\caption{Cluster Formation}
\label{alg:cluster-formation}
\begin{algorithmic}[1]
\Require Selected centers $\mathcal{C}^{*}$, dominance links $\operatorname{cdh}$, densities $\rho$
\Ensure Cluster labels $y$

\State Assign distinct labels to all centers in $\mathcal{C}^{*}$
\ForAll{points $x_i$ in descending order of $\rho_i$}
    \If{$x_i$ is unlabeled}
        \State Assign $x_i$ the label of $\operatorname{cdh}(i)$
    \EndIf
\EndFor
\State \Return $y$
\end{algorithmic}
\end{algorithm}

Of the four steps, Step~1 (Density Estimation) and Step~3 (DC Reduction) dominate the runtime, largely because they require a substantial number of pairwise distance computations. Accordingly, this paper concentrates on accelerating these two stages.

\section{GPU-accelerated FastDEC}
\label{sec:cuda-fastdec-algorithms}

FastDEC delivers high clustering quality with strong robustness to parameter choice and native support for arbitrary-shaped clusters, but its CPU-based implementation does not scale to datasets with millions of points. The dominant cost lies in the construction and maintenance of a spatial index---typically a $k$-d tree---used to obtain the self-$k$-nearest-neighbor graph that underlies both density estimation and Dominance Component (DC) reduction. Because indexing overhead grows superlinearly with the number of points $N$, and because DC reduction is tightly coupled to the index structure it queries, both stages become the dominant cost well before $N$ reaches millions of points. Closing this gap requires more than tuning the existing CPU pipeline: the self-$k$-NN and DC reduction stages must be rethought from the ground up.

A second, largely orthogonal limitation is FastDEC's failure to scale with dimensionality. The exact variant, which relies on a $k$-d tree for density estimation, degrades sharply once dimensionality exceeds roughly 10--20---a direct consequence of the curse of dimensionality, under which the pruning power of tree-based indices collapses as $D$ grows~\citep{weber1998quantitative}. Substituting an approximate index such as NNDescent extends density estimation to higher dimensions, but only relocates the bottleneck: DC reduction, no longer backed by the accuracy guarantees of an exact index, incurs substantial additional computation even on moderately sized datasets. Recent GPU-native exact $k$-NN methods are similarly oriented toward low-dimensional data: CLOVER~\citep{kamel2025clover}, for example, prunes distance computations using a graph over Voronoi cells and is explicitly optimized for 3D data; its evaluation is limited to 3- and 4-dimensional datasets, and its effectiveness on high-dimensional embeddings has not been demonstrated.

We address both limitations by re-architecting FastDEC's self-$k$-NN-based density estimation and DC reduction stages for GPU execution rather than porting the existing CPU pipeline. First, we remove redundant computation inherent to the self-$k$-NN setting: since the query and reference sets are identical, every off-diagonal pairwise distance benefits two points at once, so we compute only the upper-triangular block of the distance matrix and apply a bidirectional update, roughly halving the work of a generic $k$-NN formulation. Second, we replace index-based traversal with a family of FastDEC-specific, index-free kernels: self-$k$-NN distances are expressed as block-tiled matrix multiplication over precomputed squared norms, with warp-cooperative top-$k$ maintenance filtering candidates before they reach the heap, and DC reduction is likewise recast as a dense, index-free reduction over the same GPU-resident data. Because this formulation never builds or maintains an index, it avoids the branch divergence and irregular memory access patterns associated with tree traversal on GPUs while eliminating the dependency between indexing and DC reduction that constrains existing GPU-accelerated $k$-NN approaches. Crucially, the resulting pipeline remains exact: both self-$k$-NN construction and DC reduction are performed without algorithmic approximation, preserving the clustering quality, parameter robustness, and arbitrary-shape support of FastDEC. Although the redesign preserves the asymptotic complexity of these stages, it substantially lowers practical execution costs through redundant computation elimination, efficient GPU parallelization, and enhanced data reuse and memory locality. In particular, SymTri-kNN performs $\Theta(N^2D)$ arithmetic with $O(ND+Nk+B^2)$ working memory, where $B$ is the traversal tile size chosen by the memory-adaptive planner, while WP-DC performs $\Theta(MND)$ arithmetic, where $M \ll N$ is the number of candidate DC points. Because the memory-adaptive traversal never materializes the $N \times N$ distance matrix, memory consumption grows approximately linearly with the stored dataset and neighbor lists, and no index construction is required. This enables datasets substantially larger and higher-dimensional than those practical under the original CPU implementation, although runtime remains quadratic in the number of points.

The resulting workflow performs exact self-$k$-NN and exact DC reduction on the GPU, improving scalability while, by design, computing the same density, dominance, and reduction criteria as the original FastDEC pipeline (up to floating-point precision). The proposed algorithm offers the following key features: 
\begin{enumerate}  
    \item An index-free, purpose-built self-$k$-nearest-neighbor ($k$-NN) algorithm, termed \textit{SymTri-$k$NN}, is used for density estimation. We show that SymTri-$k$NN scales favorably with dimensionality over the tested range (up to 32,768 dimensions): the dominant $D$-dependent work is performed by highly optimized general matrix-matrix multiplication (GEMM) calls, one per visited tile. Top-$k$ heap maintenance, self-exclusion, and upper-triangular traversal are strictly independent of $D$. We demonstrate empirically that this problem-specific design outperforms a widely used general-purpose GPU $k$-NN baseline: the FastDEC variant built on SymTri-kNN is approximately $1.7\times$–$4.9\times$ faster end-to-end than the FastDEC variant built on FAISS' exact GPU $k$-NN, with both variants performing exhaustive $k$-NN search and the speed advantage generally widening on higher-dimensional datasets.
    \item A new DC reduction algorithm, Warp-Parallel DC Reduction (WP-DC), is introduced that does not depend on the index structure generated during density estimation. It features a dominance component reduction kernel that fuses closest density-higher (CDH) selection with dominance-count accumulation in a single GPU pass. 
\end{enumerate}
 By decoupling DC reduction from density estimation, the proposed design removes restrictions on the choice of $k$ in GPU-based similarity search algorithms and enables a wider range of GPU-accelerated $k$-NN methods to be integrated into the FastDEC framework. Consequently, future advances in $k$-NN algorithms can be incorporated directly into the framework. 
 
We combine SymTri-$k$NN with WP-DC to form a GPU-accelerated end-to-end FastDEC pipeline, SymWarp-FastDEC, in which the computationally dominant stages execute on the GPU, and present the end-to-end speedups in Section~\ref{sec:exp-results}.

In the following sections, we detail each step of the FastDEC framework, with particular emphasis on the two most computationally intensive steps, and describe the enhancements and optimizations that adapt the algorithm for efficient GPU acceleration when clustering large, high-dimensional datasets.

\subsection{SymTri-kNN: Symmetric Upper-Triangular Exact Self-kNN on the GPU}
\label{sec:symtri-knn}

FastDEC operates in the self-$k$-NN regime, where the query and reference sets are identical ($Q = R$). General-purpose GPU $k$-NN libraries---including FAISS (GpuIndexFlatL2)---typically still treat $Q$ and $R$ as independent. As a result, they compute both $d(i,j)$ and $d(j,i)$ and then apply additional filtering to remove the trivial self-match $i=j$.

We introduce \textbf{SymTri-kNN}, a CUDA-based exact self-$k$-NN algorithm tailored to FastDEC. SymTri-kNN exploits the symmetry of the pairwise distance matrix to eliminate redundant distance computation and to update the neighbor lists of both endpoints from each off-diagonal tile. Specifically, it leverages the $Q = R$ assumption on three levels:

\begin{itemize}
   \item \textbf{Triangular traversal.} Squared-L2 distances are obtained from a tiled Single-precision General Matrix-Matrix Multiplication (SGEMM) of the form $G = XX^{\top}$, but only tiles $(i,j)$ with $j \ge i$ are launched, halving the GEMM work, and the input reads it performs, relative to a $Q$-vs-$R$ schedule.
  \item \textbf{Symmetric bidirectional top-$k$ update.} Each off-diagonal tile $(i,j)$ is computed once and consumed by two stream-ordered, warp-cooperative selection passes: a forward pass that updates the top-$k$ heaps of block $i$, and a reverse pass that updates those of block $j$ from the same tile, without a second GEMM. From the second tile row onward, the forward pass also marks, in a bitmap, the tile entries that beat the current $k$-th-best distance of their reference point, so that the reverse pass reads only those entries rather than the whole tile. Each point's heap persists in global memory between kernel launches and is staged into shared memory by a single warp for the duration of an update; since every heap is owned by exactly one warp at a time and successive tile kernels are stream-ordered, insertions require no cross-warp locking. Ballot-filtered candidate gating ensures that only candidates beating the heap's current $k$-th-best distance are inserted, so only candidates that can actually improve a heap pay the update cost.
  \item \textbf{Index-based self-exclusion.} A candidate whose global index equals that of the query point is rejected inside the same warp-wide predicate used for ballot-filtered candidate gating, so self-exclusion adds no separate branch to the insertion loop. Because exclusion is by index rather than by distance, exact duplicates of a point (identical coordinates, different indices) remain eligible neighbors, and no host-side post-processing of the neighbor lists is required to remove self-matches.
\end{itemize}

Squared point norms are precomputed once, so that each squared distance is reconstructed from a single dot product produced by the SGEMM; arbitrary dimensionality $D$ is therefore handled by cuBLAS rather than by custom shared-memory tiling.

\subsubsection{Symmetric Upper Triangle Traversal}
\begin{figure}[t]
    \centering
    \includegraphics[width=\linewidth]{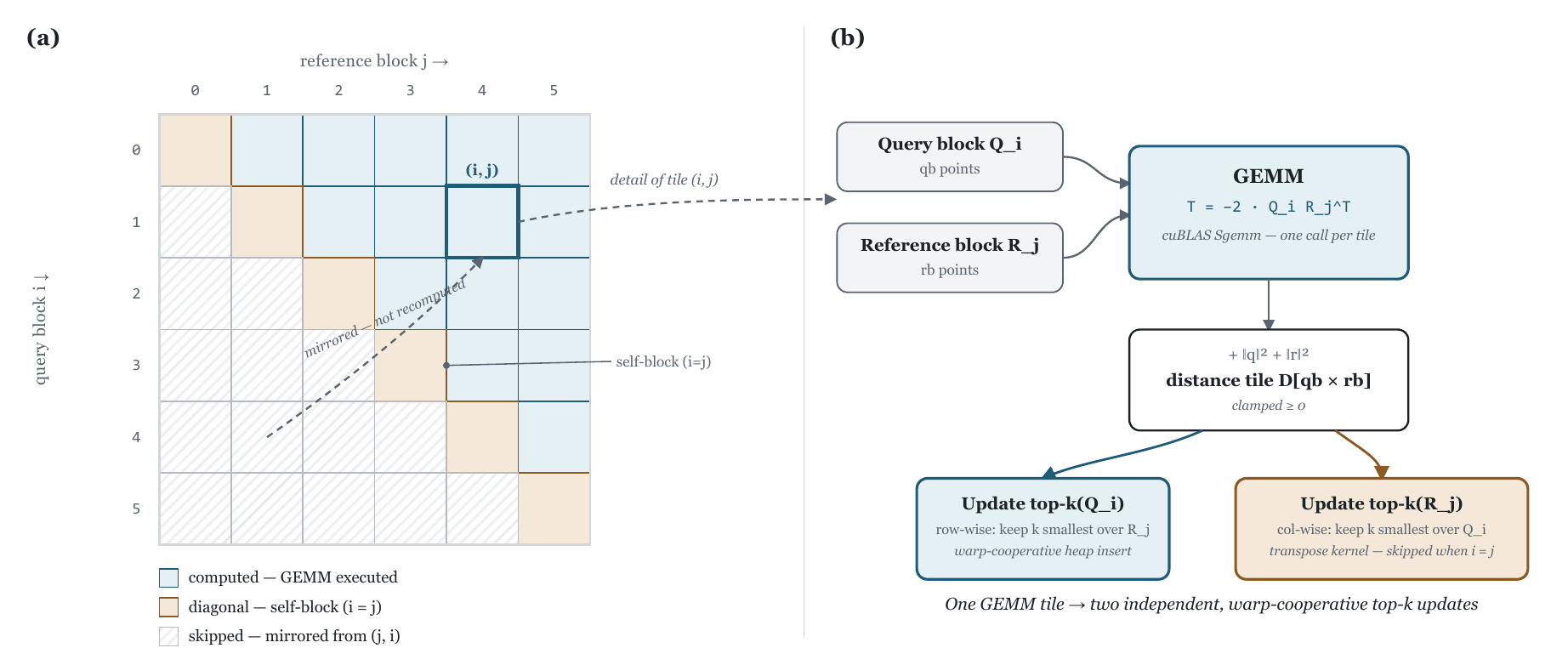}
    \caption{The grid of point blocks used by the exact self-$k$-NN kernel. 
    \textbf{(a)} Because the reference set and query set are identical, only the upper-triangular and diagonal tiles ($j \ge i$) are computed. Each computed tile is reused, through a second (reverse) selection pass, for its transposed position rather than recomputed, reducing the number of GEMM calls by approximately one half. Diagonal tiles ($i=j$) require updates only once. 
    \textbf{(b)} Detail of a computed tile. A single GEMM computes the $-2QR^{\top}$ term for the entire tile, which is combined with precomputed squared norms to form the distance tile. The resulting distances are then processed by two independent warp-cooperative kernels that update the top-$k$ heaps of both the query block and the reference block from the same computation.}
    \label{fig:selfknn-tiling}
\end{figure}

This algorithmic component exploits the symmetry of the pairwise distance matrix through tiled upper-triangular traversal. Figure \ref{fig:selfknn-tiling} illustrates the idea. Panel (a) depicts the block-tile structure of the query–reference distance matrix, organized as a $6\times6$ grid of query blocks (rows) against reference blocks (columns). Tiles on and above the diagonal are explicitly computed via GEMM: off-diagonal tiles (shaded blue) pair two distinct blocks, while diagonal tiles (shaded orange) correspond to self-blocks where $i = j$. Tiles below the diagonal (hatched) are skipped entirely, since their distances are reused from the already-computed transpose tile $(j, i)$ rather than recomputed. An inset highlights a representative tile $(i, j)$ to indicate the detail expanded in panel (b).

Panel (b) details the per-tile computation pipeline. A query block $Q_i$ ($qb$ points) and reference block $R_j$ ($rb$ points) are passed to a single SGEMM call that computes $T = -2 \cdot Q_i R_j^{\top}$. The squared point norms $\lVert q \rVert^2$ and $\lVert r \rVert^2$ are then added to $T$ and the result clamped to be non-negative, yielding the distance tile $D[qb \times rb]$. This single distance tile drives two independent, warp-cooperative top-$k$ updates: a row-wise update maintaining the $k$ smallest distances for $Q_i$, and a column-wise update maintaining the $k$ smallest distances for $R_j$ — the latter reading the tile column-wise (by a strided read, an explicit transpose, or, from the second tile row onward, only the entries marked in the survivor bitmap) and skipped for diagonal tiles ($i = j$). This dual-update design allows a single GEMM tile to advance the top-$k$ neighbor lists of both its query and reference blocks, which is what enables the upper-triangular traversal in panel (a) to avoid redundant computation.

\begin{algorithm}
\caption{\textsc{SymTri-kNN}: Exact Self-$k$-NN via Tiled Upper-Triangular Computation}
\label{alg:symtri-knn}
\begin{algorithmic}[1]
\Require point set $X = \{x_1,\dots,x_N\} \subset \mathbb{R}^D$, neighbor count $k$
\Ensure for every point $i$: its $k$ nearest neighbors $(\mathrm{dist}_i, \mathrm{idx}_i)$
\State $\mathrm{norm}_i \gets \lVert x_i \rVert^2$ \quad for all $i = 1,\dots,N$ \Comment{precompute squared norms once}
\State $H_i \gets \textsc{EmptyMaxHeap}(k)$ \quad for all $i = 1,\dots,N$ \Comment{per-point top-$k$ candidate heap}
\State choose block sizes $QB = RB$ from available device memory \Comment{square tiling is required for symmetry}
\For{$q \gets 0, QB, 2QB, \dots < N$} \Comment{iterate query blocks}
  \State $Q \gets X[q : q + QB)$
  \For{$r \gets q, q + RB, \dots < N$} \Comment{reference blocks: upper triangle only}
    \State $R \gets X[r : r + RB)$
    \State $T \gets -2 \cdot Q R^{\top}$ \Comment{one SGEMM call per tile}
    \State $D[a,b] \gets \max\!\big(\mathrm{norm}_{q+a} + \mathrm{norm}_{r+b} + T[a,b],\ 0\big)$ \quad for all tile entries $(a,b)$
    \State \Call{UpdateTopK}{$H$, $\mathrm{idx}(Q)$, $\mathrm{idx}(R)$, $D$} \Comment{row-wise: updates query block $Q$}
    \If{$r \neq q$}
      \State \Call{UpdateTopK}{$H$, $\mathrm{idx}(R)$, $\mathrm{idx}(Q)$, $D^{\top}$} \Comment{mirrored update of $R$ — no recomputation}
    \EndIf
  \EndFor
\EndFor
\For{$i \gets 1$ to $N$}
  \State $(\mathrm{dist}_i, \mathrm{idx}_i) \gets$ entries of $H_i$ sorted ascending by distance
  \State $\mathrm{dist}_i \gets \sqrt{\mathrm{dist}_i}$ \quad (elementwise)
\EndFor
\State \Return $\{(\mathrm{dist}_i, \mathrm{idx}_i)\}_{i=1}^{N}$
\end{algorithmic}
\end{algorithm}

\begin{algorithm}
\caption{\textsc{UpdateTopK}: Candidate-Filtered Heap Update for One Distance Tile}
\label{alg:updatetopk}
\begin{algorithmic}[1]
\Procedure{UpdateTopK}{$H$, $\mathrm{targets}$, $\mathrm{sources}$, $D$}
  \For{each $t \in \mathrm{targets}$ \textbf{in parallel}} \Comment{one warp per target point}
    \State $w \gets H_t[0]$ \Comment{current $k$-th-best distance}
    \For{each $s \in \mathrm{sources}$}
      \If{$s \neq t$ \textbf{and} $D[t,s] < w$}
        \State insert $(D[t,s],\, s)$ into $H_t$; \quad $w \gets H_t[0]$
      \EndIf
    \EndFor
  \EndFor
\EndProcedure
\end{algorithmic}
\end{algorithm}

Algorithm~\ref{alg:symtri-knn} formalizes the block-tiled traversal outlined above. It begins by precomputing the squared norm of every point once (line 1) and giving each point an empty, bounded-capacity max-heap to hold its current $k$ best candidates (line 2). The $N$ points are then partitioned into query and reference blocks, and the algorithm sweeps only the upper-triangular block pairs $r \ge q$ (lines 4–7): since the query and reference sets are identical, computing pair $(q,r)$ and its mirror $(r,q)$ separately would be redundant, so each unordered pair of blocks is visited exactly once. For the visited pair, a single dense matrix multiplication produces the $-2\cdot QR^\top$ cross term for the entire tile in one call (line 8), which is then combined with the precomputed norms to reconstruct the squared Euclidean distance tile $D$ (line 9)---turning the dominant cost of the algorithm into a small number of large, hardware-optimized GEMM calls rather than a pairwise distance loop.

The resulting tile is consumed by the \textsc{UpdateTopK} subroutine (Algorithm~\ref{alg:updatetopk}), which the driver calls twice whenever $q \ne r$: once on $D$ to update the query block's heaps, and once on $D^\top$---at no extra distance computation cost---to update the reference block's heaps, so one computed tile resolves both directions of the pair; diagonal tiles ($q = r$) require only the first call, since they have no mirror. Inside \textsc{UpdateTopK}, each target point's heap is updated independently and in parallel: a candidate is only inserted if it beats the point's current $k$-th-best distance, so points whose neighbor lists are already unlikely to change do most of their work in a cheap comparison rather than a heap update.

The block size used for GEMM tiling is chosen adaptively from the amount of free device memory at runtime, with a fallback search over candidate sizes if the preferred allocation fails. This allows a single implementation to adapt its tile size to the size of the set of points and the available memory of the device without manual retuning per data set or GPU; the underlying cost of pairwise-distance $\Theta(N^2 D)$ is not affected by this adaptivity.

\subsubsection{Tile Size Planner and Traversal Scheduler}

The planner is responsible for choosing the dot-product tile size and allocating the reusable workspace. The block size that parameterizes this traversal is selected once, before the main loop begins, through a two-stage procedure that separates prediction from verification. First, given the amount of free device memory reported at runtime, the implementation reserves a fixed fraction of it as a workspace budget and searches a small set of candidate block sizes---coarser for large point sets, finer for small ones---returning the largest candidate whose quadratic memory footprint fits within that budget and does not exceed $N$. Preferring the largest feasible size is deliberate: larger GEMM tiles push the underlying dense linear algebra kernel closer to peak throughput and amortize launch overhead over more useful work. Second, because a free-memory query only estimates whether a contiguous allocation will actually succeed, the implementation attempts the allocation at the predicted size and, if it fails, falls back through a further list of decreasing candidate sizes until one succeeds. The resulting size is applied uniformly to both the query and reference block dimensions, which is what preserves the square, symmetric tiling the traversal scheduler relies on. Because this workspace is allocated once and reused across the entire traversal, the cost of the search is incurred a single time per invocation, independent of how many block pairs are subsequently processed, allowing the same implementation to adapt its tile size to point sets of very different sizes without manual retuning---though this adaptivity to dataset size does not change the $\Theta(N^2 D)$ arithmetic complexity of the underlying pairwise-distance computation.

The planner therefore optimizes for three goals:
\begin{itemize}
  \item \textbf{Large SGEMM efficiency.} Larger tiles give the BLAS backend more work per call and generally improve GPU utilization.
  \item \textbf{Memory feasibility.} The tile size is selected according to available memory, avoiding hard failure on smaller GPUs.
  \item \textbf{Symmetry preservation.} Query and reference tile sizes are kept equal, so that upper-triangular scheduling remains valid.
\end{itemize}

The scheduler traverses the pairwise tile matrix produced by the planned square tiling. The traversal scheduler restricts the double loop over query and reference blocks to the upper triangle of the $N/B \times N/B$ block grid (Algorithm~\ref{alg:symtri-knn}, lines 4--13; Figure~\ref{fig:selfknn-tiling}a). Because a query or reference block is simply a pointer offset into a single resident device buffer, forming a block requires no data movement; the scheduler instead controls only which block pairs are visited, and in what order the resulting distance tile is consumed. For an off-diagonal pair $(q,r)$ with $r > q$, a single GEMM call produces the pairwise dot-product tile, which is then read twice under two different traversal orders rather than recomputed: once row-wise, to update the top-$k$ heaps of the query block, and once more, to update the top-$k$ heaps of the reference block---read directly in column order for small $k$, or first transposed on-chip into a coalesced layout for larger $k$---so that one tile resolves both directions of the pair at the cost of a single matrix multiplication. Diagonal pairs ($r = q$) require only the row-wise pass, since scanning a block against itself already gives every point in it a turn as query against every other point, with self-matches excluded explicitly. This traversal order lets the scheduler visit each unordered block pair exactly once while still fully updating every block it touches, halving the number of GEMM calls relative to a naive dense traversal without any loss of correctness.

Within each update kernel, scheduling is warp-based: a CUDA block consists of 256 threads (8 warps), with each warp assigned to the top-$k$ update of a single point, as described in Section~\ref{sec:wc-topk}.

\subsubsection{Warp-Cooperative Top-\texorpdfstring{$k$}{k} Selection}
\label{sec:wc-topk}

SymTri-$k$NN uses a streaming warp-cooperative top-$k$ selection algorithm to avoid materializing and sorting full pairwise distance lists, converting exact $k$-NN search from full-list sorting into streaming top-$k$ maintenance. Each warp is assigned to one data point and maintains the current best $k$ nearest-neighbor set of that point as a small max-heap in shared memory. For each distance tile, the warp proceeds as follows:
\begin{enumerate}
  \item Load the point's current top-$k$ heap from global memory to shared memory.
  \item Scan a tile of candidate points cooperatively across warp lanes.
  \item Use warp ballots to identify candidates that improve the heap.
  \item Insert only improving candidates into the shared-memory max-heap.
  \item Write the updated heap back to global memory.
\end{enumerate}

\begin{algorithm}[t]
\caption{Warp-Cooperative Top-$k$ Selection}
\label{alg:warp_cooperative_topk}
\begin{algorithmic}[1]
\Require Distance tile represented by dot products $G$;
         squared norms $\nu$;
         tile offsets $q_0,r_0$;
         current top-$k$ heaps $\{H_i\}$;
         neighbor count $k$
\Ensure Updated top-$k$ heaps

\State Assign one warp to each query point $q$ in the query tile
\State Load $H_q$ into shared memory
\State $\tau \gets$ distance of the root of $H_q$ \Comment{Lexicographically largest retained pair}

\ForAll{candidate batches of 32 reference points}
    \State Each warp lane $\ell$ evaluates one candidate $r$
    \State Compute
    \[
        d^2(q,r) = \nu_q + \nu_r - 2x_q^\top x_r .
    \]
    \State Clamp $d^2(q,r) \gets \max(d^2(q,r),0)$
    \State Mark lane $\ell$ active iff
    \[
        r \neq q \quad \text{and} \quad d^2(q,r) < \tau .
    \]
    \State $M \gets \mathrm{Ballot}(\text{active lanes})$

    \While{$M \neq 0$}
        \State Select one active lane $\ell^\star$ from $M$
        \State Broadcast $(d^\star, r^\star) \gets (d^2(q,r), r)$ from $\ell^\star$
        \If{$d^\star < \tau$} \Comment{recheck: $\tau$ may have dropped since the ballot}
            \State Lane $0$ replaces the root of the shared-memory max-heap $H_q$ (ordered lexicographically by $(d^2,\mathrm{index})$) with $(d^\star, r^\star)$ and restores the heap
            \State Lane $0$ updates $\tau \gets$ distance of the root of $H_q$
            \State Broadcast $\tau$ to all lanes
        \EndIf
        \State Remove $\ell^\star$ from $M$
    \EndWhile
\EndFor

\State Write $H_q$ back to global memory

\end{algorithmic}
\end{algorithm}

Algorithm~\ref{alg:warp_cooperative_topk} details the warp-cooperative procedure used to update the top-$k$ neighbor heaps from a computed distance tile. Each warp is assigned to a single query point $q$ and maintains that point's top-$k$ heap $H_q$ in shared memory, using the distance $\tau$ of the current worst retained neighbor (the heap root) as a rejection threshold.

Reference points are processed in batches of 32, with each of the warp's 32 lanes evaluating one candidate $r$ in parallel. Each lane computes the squared Euclidean distance $d^2(q,r)$ from the precomputed dot product and squared norms, clamping the result to be non-negative to guard against floating-point error. A lane is marked active if its candidate is a valid, sufficiently close neighbor---that is, $r \neq q$ and $d^2(q,r) < \tau$---and a warp-wide ballot produces a 32-bit mask identifying the set of active lanes. If the mask is empty, which is the common case once the heap holds good neighbors, the warp skips the heap entirely and proceeds to the next batch.

Because concurrent lanes cannot safely update a shared heap simultaneously, active candidates are then inserted serially: one active lane at a time is extracted from the ballot mask with \texttt{\_\_ffs}, and its distance and index are broadcast to the warp via warp-shuffle instructions. Because an earlier insertion in the same batch may have lowered $\tau$, lane 0 first rechecks the candidate against the current $\tau$ and discards it if it no longer qualifies; otherwise it performs the heap insertion and recomputes $\tau$, which is broadcast back to all lanes before the next candidate is processed. This continues until all active candidates in the batch have been consumed, after which the warp proceeds to the next batch of 32 reference points. Once all candidates in the tile have been evaluated, the updated heap $H_q$ is written back to global memory.

The key GPU optimization is thus the combination of warp-cooperative execution with ballot-filtered candidate insertion. Warp-level ballot and shuffle operations replace slow thread-synchronization primitives, distance computation and threshold filtering exploit full warp-wide parallelism, and the sequential heap update is confined to a single lane and performed only for candidates that can actually change the top-$k$ set---every other candidate is rejected by a single parallel comparison.

The core data structure is a max-heap over distances, with the root storing the worst neighbor retained so far. Top-$k$ selection repeatedly asks one question: \emph{is this new candidate closer than the worst neighbor currently retained?} With the worst neighbor at the root, this comparison is $O(1)$; only a candidate that improves the top-$k$ list replaces the root and restores the heap---implemented as a hole-based sift-down that shifts entries down without pairwise swapping---at a cost of $O(\log k)$. Therefore, processing all candidates for one point costs approximately
\begin{equation}
  O(N) \text{ comparisons} \;+\; O(m \log k) \text{ heap updates},
  \label{eq:heapcost}
\end{equation}
where $m$ is the number of candidates that actually enter the heap. Usually $m \ll N$, especially once the heap has been filled with good neighbors.

Keeping the heap in shared memory for the duration of a tile avoids repeated global-memory accesses during heap updates. Once all tiles have been processed, the retained $k$ neighbors for each point are sorted with an insertion sort for small $k$ or a GPU radix sort for larger $k$. Since $k$ is small in FastDEC relative to $N$, sorting these $k$ elements incurs negligible cost. A final kernel then converts the retained squared distances to Euclidean distances by taking their square roots, yielding values suitable for downstream processing.

\label{sec:tie-break}
SymTri-$k$NN breaks distance ties deterministically in favor of the smaller point index. Because each point receives its candidates in increasing index order and the heap treats the larger index as worse on equal distances, the distance-only admission test (line~8 of Algorithm~\ref{alg:warp_cooperative_topk}) already retains the $k$ lexicographically smallest $(d^2,\mathrm{index})$ pairs, and the final sort uses the same order. This makes the neighbor lists, and hence DC detection, reproducible. On MNIST, ties affected about 2\% of rows at $k=256$ without changing any selected CDH point in the cases examined.

\subsection{Density Estimation}
\label{sec:gpu-density-estimation}

Density estimation is integrated into the end-to-end \textit{SymWarp-FastDEC} pipeline and operates directly on the self-$k$-NN distances produced by the GPU stage. Consistent with the original reference implementation, it uses the exponential kernel density estimator and procedure described in Section~\ref{sub:density-estimation} (Algorithm~\ref{alg:density-estimation}).

The computational cost of density estimation is $O(Nk)$, which is negligible relative to the $\Theta(N^2D)$ self-$k$-NN computation and the $\Theta(mND)$ DC reduction stage. Consequently, density estimation is executed on the host without a dedicated GPU kernel, as GPU acceleration of this stage would have minimal impact on overall runtime.

\subsection{Dominance Component Detection}
\label{sec:fastdec-dc-detection}

Dominance Component detection is integrated into the end-to-end pipeline and uses the same procedure described in Section~\ref{sub:dominance-estimation} (Algorithm~\ref{alg:dominance-detection}). For each point, the ordered $k$-NN list is scanned to identify the first neighbor with strictly higher density; points for which no such neighbor exists are returned as DCs.

Like density estimation, this stage has computational complexity $O(Nk)$ and is therefore inexpensive relative to the dominant self-$k$-NN and DC reduction stages. Consequently, it is executed on the host.

Retaining the complete $k$-column neighbor table produced during density estimation, rather than only the nearest higher-density match, enables the adaptive $k$-halving fallback (Section~\ref{sub:dominance-estimation}) to reuse subsets of the already-computed neighborhoods. Density estimation and Dominance Component detection can therefore be recomputed without requiring an additional self-$k$-NN invocation.

To verify that these host-side stages do not limit end-to-end performance, we also evaluated a fully GPU-resident variant that keeps the neighbor graph on the GPU and executes density estimation, DC detection, and cluster assignment there as well. At matched precision, it selected the same DC counts and ordered centers as the default pipeline in all tested configurations. However, in FP32 it saved only about 0.6\,s per fit on the 500k-point synthetic sweep (a $1.008\times$ speedup at 4096d) and gave at most $1.08\times$ on the million-point workloads; a substantial relative gain appeared only on sub-second fits such as mnist\_70k\_2d. Because self-$k$-NN dominates the runtime, we retain the simpler host-side implementation of these stages.

\subsection{Dominance Component Reduction}
\label{sec:gpu-dc-reduction}

DC reduction is the process that reduces the set of dominance components (DCs) to the final cluster centers. For each DC, this requires computing a dominance score, defined by three factors: the DC's normalized density, its distance to its CDH point, and its global rank---the number of reference points closer to the DC than its CDH. Once dominance scores are computed, DCs are sorted in descending order by score, and the top $c$ are selected as cluster centers.

Computing the dominance score is expensive: it requires the distance from each DC to every point in the reference dataset, sorted in ascending order so that the rank can be read off as the number of points closer than the CDH. The original CPU-based FastDEC obtains these distances by reusing the spatial index graph already built during density estimation, performing an $(N-1)$-nearest-neighbor search for each DC, where $N$ is the size of the reference dataset. This coupling is efficient on CPU, where the index can be built once and cached in memory for reuse across the DC reduction step. It is impractical on GPU for several reasons. First, GPUs are inefficient at constructing complex index structures such as $k$-d trees and at sorting large arrays of distances. Second, most GPU-based $k$-NN libraries do not support queries for as many neighbors as $(N-1)$ requires; FAISS, for example, caps $k$ at 2048, far below $N-1$ for datasets of practical size. Third, GPU memory is a scarce resource, and an index graph large enough to support this search would consume memory better spent on computation, degrading parallel performance. Finally, coupling DC reduction to the density-estimation index limits FastDEC's flexibility, restricting which $k$-NN algorithms and hardware configurations it can exploit.

To address these limitations, we propose a fast and efficient DC reduction algorithm that decouples dominance-score computation from density estimation and eliminates the need for any pre-built index or large-scale distance sorting. This decoupling allows DC reduction to instead rely on a $k$-NN algorithm purpose-built for GPU execution, improving computational efficiency and scalability on large datasets while allowing FastDEC to benefit directly from future advances in GPU-based $k$-NN methods. Our experimental results demonstrate that the proposed DC reduction algorithm performs consistently well across all datasets evaluated, accounting for less than 1\% of the total runtime on 11 of the 15 datasets evaluated and for at most 7.36\% on any of them. The algorithm effectively eliminates DC reduction as a computational bottleneck.

\subsubsection{Warp-Parallel DC Reduction (WP-DC)}
\label{sec:wp-dc}

We propose \textsc{WP-DC} (Warp-Parallel DC Reduction), a GPU algorithm that casts distance evaluation between DC points and the reference dataset as a single tiled SGEMM operation and performs CDH selection (Stage A) and rank counting (Stage B) using warp-level reductions. A fast path fuses the two reduction stages when the dataset fits within a single slab. The implementation further employs a memory-adaptive slab planner and streamed double-buffered (ping-pong) execution to improve GPU utilization across varying dataset sizes.

Algorithm~\ref{alg:wp-dc-reduction} summarizes the proposed DC reduction algorithm on the GPU. The expensive DC-to-dataset dot products are evaluated by SGEMM, while the two remaining operations are expressed as warp-level reductions. Stage A (Algorithm~\ref{alg:stagea}) selects the closest higher-density point for each DC, and Stage B (Algorithm~\ref{alg:stageb}) counts the number of points lying strictly inside the corresponding CDH radius.

\begin{algorithm}[!htbp]
\caption{Warp-Parallel DC Reduction (high-level workflow)}
\label{alg:wp-dc-reduction}
\begin{algorithmic}[1]

\Require Dataset $X \in \mathbb{R}^{N \times dim}$ with densities $\rho_X$;
         DCs $C \in \mathbb{R}^{m \times dim}$ with densities $\rho_C$
\Ensure CDH distance $g[\cdot]$, CDH index $idx[\cdot]$, and rank $r[\cdot]$
        for every DC

\State $n_X \gets \textsc{SquaredNorms}(X)$
\State $n_C \gets \textsc{SquaredNorms}(C)$

\State $g[1..m] \gets +\infty$, $idx[1..m] \gets -1$, $r[1..m] \gets 0$

\State $(T_c,T_d,fused) \gets
    \textsc{PlanTiling}(N,m,dim,\text{memory budget})$
\Comment{$fused=\mathrm{true}$ iff one full dataset block fits in the GEMM buffer}

\For{each DC tile $C_t \subseteq C$ of size $\leq T_c$
      \textbf{(streamed/pipelined)}}

    \If{$fused$}
        \State $D \gets C_t X^{\top}$
        \Comment{single SGEMM, $\lvert C_t\rvert \times N$}

        \State \Call{StageA-Reduce}
            $(D,n_X,\rho_X,n_C[C_t],\rho_C[C_t])$
        \Comment{update $g[C_t]$ and $idx[C_t]$}

        \State \Call{StageB-Reduce}
            $(D,n_X,n_C[C_t],g[C_t])$
        \Comment{reuse the same $D$; atomic accumulation into $r[C_t]$}

    \Else
        \For{each dataset block $X_b \subseteq X$ of size $\leq T_d$}
            \State $D \gets C_t X_b^{\top}$
            \Comment{SGEMM tile}

            \State \Call{StageA-Reduce}
                $(D,n_X[X_b],\rho_X[X_b],n_C[C_t],\rho_C[C_t])$
            \Comment{merge across blocks via atomic packed min}
        \EndFor

        \For{each dataset block $X_b \subseteq X$ of size $\leq T_d$}
            \State $D \gets C_t X_b^{\top}$
            \Comment{SGEMM tile, recomputed}

            \State \Call{StageB-Reduce}
                $(D,n_X[X_b],n_C[C_t],g[C_t])$
            \Comment{atomic accumulation into $r[C_t]$}
        \EndFor
    \EndIf
\EndFor

\For{$i = 1$ \textbf{to} $m$}
    \If{$idx[i] = -1$}
        \State $g[i] \gets -1$, $r[i] \gets -1$
    \Else
        \State $g[i] \gets \sqrt{g[i]}$
        \If{$r[i] = 0$}
            \State $r[i] \gets -1$
        \EndIf
    \EndIf
\EndFor

\State \Return $(g,idx,r)$

\end{algorithmic}
\end{algorithm}

WP-DC computes the inter-dominance quantities used to rank DC points and identify the final cluster centers. For each DC point, it determines (i) its CDH point, searched over the entire dataset rather than only its $k$ nearest neighbors, (ii) the corresponding distance, and (iii) a global rank defined as the number of data points closer than the CDH distance. To improve computational efficiency, WP-DC reformulates the dominant distance computation from explicit pairwise loops as matrix multiplications and employs fused GPU reductions to efficiently extract the CDH information and global ranks.
Given a dataset $X=\{x_i\}_{i=1}^{N}$ with densities $\rho_i$ and a set of DC points $C=\{c_j\}_{j=1}^{m}$ with densities $\rho^C_j$, the algorithm computes, for each DC $c_j$, three quantities: the index $h_j$ of its CDH point, the corresponding CDH distance $g_j$, and the rank $r_j$. The CDH point is defined as
\begin{equation}
    h_j =
    \arg\min_{i:\rho_i>\rho^C_j}
    \|x_i-c_j\|_2^2 ,
\end{equation}
and the rank is the number of dataset points closer to $c_j$ than its CDH point:
\begin{equation}
    r_j =
    \left\lvert\left\{
    i : \|x_i-c_j\|_2^2 < \|x_{h_j}-c_j\|_2^2
    \right\}\right\rvert
\end{equation}
In exact arithmetic, and whenever $g_j>0$, $r_j$ equals the CDH's zero-based position in the reference implementation's full sorted neighbor list, since the DC itself lies strictly inside the CDH radius at distance zero and is counted. The two can differ only under distance ties: the reference's positional rank may then depend on how its search index orders equidistant points, whereas $r_j$ counts only points strictly inside the CDH radius and is independent of that ordering.
If no higher-density point exists for $c_j$, the algorithm assigns sentinel values to indicate that $c_j$ has no valid CDH.

As in SymTri-$k$NN (Section~\ref{sec:symtri-knn}), explicit coordinate-wise distance computation is replaced with dense matrix multiplication: for a DC tile $Q\subseteq C$ and a dataset slab $S\subseteq X$, a single SGEMM call computes the dot-product matrix $Z_{Q,S}=QS^\top$, which combines with precomputed squared norms $\beta_i=\|c_i\|_2^2$ and $\alpha_j=\|x_j\|_2^2$ to reconstruct squared distances as $d_{ij}^2=\beta_i+\alpha_j-2z_{ij}$, where $z_{ij}=c_i^\top x_j$. The norm vectors $\alpha$ and $\beta$ are computed once and reused in both reduction stages.

The computation is organized over DC tiles and dataset slabs. A DC tile contains at most $T$ DC points, and each dataset slab contains at most $B$ data points. For each tile-slab pair, SGEMM produces the corresponding dot-product matrix, which is consumed by two warp-parallel reducers. The algorithm uses a warp-per-DC mapping: each GPU warp is responsible for one DC point in the current tile, while the lanes of the warp partition the dataset slab.

\begin{algorithm}[!htbp]
\caption{\textsc{StageA-Reducer}: Fused Warp-Parallel CDH Selection}
\label{alg:stagea}
\begin{algorithmic}[1]

\Require Dot-product tile $D \in \mathbb{R}^{|C_t|\times|B|}$; norms $n_X,n_C$; densities $\rho_X,\rho_C$
\Ensure Per-DC best squared distance $g$ and index $idx$ for this tile

\For{each DC $c \in C_t$ \textbf{in parallel} (one warp per DC)}
    \State $(best_d,best_i) \gets (+\infty,-1)$

    \For{each point $x \in B$ processed by this lane}
        \If{$\rho_C[c] < \rho_X[x]$}
            \Comment{$x$ must be strictly denser than $c$}
            \State $d \gets \max\big(n_X[x] + n_C[c] - 2\,D[c,x],\ 0\big)$
            \If{$d < best_d$ \textbf{or} ($d = best_d$ \textbf{and} $x < best_i$)}
                \State $(best_d,best_i) \gets (d,x)$
            \EndIf
        \EndIf
    \EndFor

    \State $(best_d,best_i) \gets
        \textsc{WarpReduceArgMin}(best_d,best_i)$

    \If{$best_i \neq -1$}
        \State $\textsc{AtomicPackedMin}\big(g[c],idx[c],best_d,best_i\big)$
        \Comment{lock-free 64-bit atomicMin on packed $(dist,index)$}
    \EndIf
\EndFor

\end{algorithmic}
\end{algorithm}

The first reducer, Stage A, described in Algorithm~\ref{alg:stagea}, identifies the closest higher-density point for each DC point. Each lane scans a strided subset of the slab and considers only points satisfying $\rho_i>\rho^C_j$. For all admissible points, the lane reconstructs $d_{ij}^2$ and keeps its best candidate pair $(d_{ij}^2,i)$ in registers. The lane-local candidates are then reduced within the warp using shuffle instructions. The comparison is lexicographic:
\begin{equation}
    (d_{ij}^2,i) \prec (d_{i'j}^2,i')
    \iff
    d_{ij}^2 < d_{i'j}^2
    \;\;\text{or}\;\;
    \left(d_{ij}^2=d_{i'j}^2 \;\text{and}\; i<i'\right).
\end{equation}
Thus, the smallest distance is selected, with deterministic tie-breaking by the smaller dataset index.

The Stage A warp reduction is lock-free within the warp. Candidate distances and indices are kept in registers, and warp shuffle primitives exchange these register values directly between lanes. No shared-memory reduction, block-level barrier, or fine-grained atomic operation is required during the scan. After the warp-level reduction, lane 0 holds the best candidate for that DC point over the current slab.

To merge slab-level winners into the persistent CDH state, the algorithm packs each candidate pair into a 64-bit word:
\begin{equation}
    p(d^2,i)
    =
    \operatorname{bits}(d^2) \ll 32
    \;\; \vert \;\;
    i .
\end{equation}
Because all squared distances are non-negative FP32 values, their unsigned IEEE-754 bit patterns preserve numeric ordering. Consequently, an unsigned atomic minimum on $p(d^2,i)$ implements an ordered minimum over $(d^2,i)$. This reduces global synchronization to one compact atomic update per warp-level winner rather than one atomic update per candidate point.

\begin{algorithm}[!htbp]
\caption{\textsc{StageB-Reducer}: Fused Warp-Parallel Rank Counting}
\label{alg:stageb}
\begin{algorithmic}[1]

\Require Dot-product tile $D \in \mathbb{R}^{|C_t|\times|B|}$; norms $n_X,n_C$; CDH squared distance $g$ per DC
\Ensure Partial rank count for each DC in this tile

\For{each DC $c \in C_t$ \textbf{in parallel} (one warp per DC)}
    \If{$g[c] = +\infty$}
        \State \textbf{continue}
        \Comment{no CDH found: no rank contribution}
    \EndIf

    \State $cnt_{\mathrm{lane}} \gets 0$

    \For{each point $x \in B$ processed by this lane}
        \State $d \gets \max\big(n_X[x] + n_C[c] - 2\,D[c,x],\ 0\big)$
        \If{$d < g[c]$}
            \State $cnt_{\mathrm{lane}} \gets cnt_{\mathrm{lane}} + 1$
        \EndIf
    \EndFor

    \State $cnt \gets \textsc{WarpReduceSum}(cnt_{\mathrm{lane}})$

    \If{$cnt > 0$}
        \State $\textsc{AtomicAdd}\big(r[c],cnt\big)$
    \EndIf
\EndFor

\end{algorithmic}
\end{algorithm}

After Stage A, the packed CDH state of each DC point is decoded into the squared CDH distance
\begin{equation}
    \delta_j = \|x_{h_j}-c_j\|_2^2
\end{equation}
and the CDH index $h_j$. The second reducer, Stage B, illustrated in Algorithm~\ref{alg:stageb}, then computes the rank associated with each DC point. Using the same distance reconstruction, each lane counts its assigned points satisfying $d_{ij}^2 < \delta_j$. The lane-local counts are summed within the warp using shuffle reduction. Only the resulting warp-level count is added to the global rank accumulator:
\begin{equation}
    r_j \leftarrow r_j + p_j(S),
\end{equation}
where
\begin{equation}
    p_j(S)
    =
    \sum_{x_i\in S}
    \mathbf{1}\!\left[d_{ij}^2 < \delta_j\right].
\end{equation}
As in Stage A, the hot loop uses only register-local updates and warp-level shuffle communication. The only global synchronization is one atomic addition per DC point and slab.

When the dataset slab can cover the full dataset in memory, the algorithm uses a fused single-slab execution. In this case, the SGEMM result for a DC tile is consumed first by Stage A and then immediately by Stage B. This avoids recomputing the dot-product matrix for rank counting. Otherwise, the algorithm uses a two-pass tiled execution: Stage A first scans all dataset slabs to find the CDH state of every DC point, and Stage B then scans the slabs again to compute ranks from the selected CDH distances.

Finally, the squared CDH distances are converted to Euclidean distances, i.e., $g_j = \sqrt{\delta_j}$. If a DC point has no valid CDH, the output values are set to sentinels. If the rank count is zero, the rank is also assigned a sentinel value. The final outputs of the reduction are therefore
\begin{equation}
    \{(h_j,g_j,r_j)\}_{j=1}^{m},
\end{equation}
which are subsequently used by FastDEC to score and select the final cluster centers. Final square-root conversion and sentinel handling are fused in one kernel pass (fused finalize kernel).

Before entering the significance-score computation described in Section~\ref{sub:dc-reduction}, sentinels ($g_j=-1$, $r_j=-1$) are replaced with the maximum $g$ and $r$ values observed across all DCs, consistent with the original CPU FastDEC convention for globally dominant DCs (those with no valid CDH). WP-DC also assigns the rank sentinel when the rank count is zero. Since the DC itself is counted whenever $g_j>0$, this occurs only when a higher-density point coincides with the DC ($g_j=0$), a degenerate case that requires exact duplicate points with different densities. Such a DC receives the same maximum-rank fill value as a globally dominant DC.

The runtime advantage of this design comes from three sources. First, the dominant arithmetic is expressed as SGEMM, enabling highly optimized GPU matrix-multiplication kernels. Second, both CDH selection and rank counting are performed by warp-level register reductions, avoiding shared-memory reductions and block-wide synchronization in the inner loops. Third, global atomic operations are applied only after each warp has reduced a whole slab of candidates or counts, greatly reducing contention relative to a point-wise atomic design.

\subsubsection{Planner and scheduler}

WP-DC uses a memory-adaptive planner to choose between tiled two-pass execution and a fused single-slab execution. For a DC tile of size $T$ and a dataset slab of size $B$, the SGEMM output requires $4TB$ bytes. The planner estimates the available memory after fixed allocations and a 1024 MiB reserve, applies a 0.85 safety factor, divides the budget across two ping-pong buffers, and caps the per-buffer budget at 2560 MiB. It then computes the largest full-dataset DC tile,
\[
T_{\max}=\left\lfloor \frac{M_{\mathrm{buf}}}{4N}\right\rfloor.
\]
If $\min\{1024,T_{\max},m\}$ is at least the minimum tile size, the planner sets $B=N$ and enables the fused path, allowing Stage~B to reuse the Stage-A SGEMM output and eliminate an additional matrix multiplication. Otherwise, WP-DC uses dataset slabs of size 32768 and performs separate Stage-A and Stage-B passes. The planner therefore reduces redundant SGEMM work whenever sufficient memory is available while retaining a bounded-memory execution mode for large datasets.

In addition to the default single-stream mode, DC reduction can optionally be scheduled across two CUDA streams. The scheduler distributes tile-slab work between the streams using two ping-pong buffers for dot-product output. When stream interleaving is enabled, consecutive work units are assigned alternately to the two streams. Operations that depend on the same buffer are kept within the same stream; for example,
\[
\mathrm{SGEMM}
\rightarrow
\mathrm{Stage\ A}
\rightarrow
\mathrm{CDH\ decode}
\rightarrow
\mathrm{Stage\ B}.
\]
In the fused path, CUDA stream ordering guarantees that the reducer kernels consume a buffer only after SGEMM has produced it, and that Stage~B executes only after the CDH distance has been decoded. In the two-pass path, Stage~A and Stage~B are separated by a stream synchronization because rank counting requires the final CDH distance. Atomic minimum updates in Stage~A and atomic additions in Stage~B make partial results from different slabs safe under stream interleaving. The two buffers prevent a subsequent SGEMM from overwriting data that a reducer is still consuming.

The goal is to expose overlap between SGEMM and reduction kernels while preserving the dependency ordering required by each tile/slab pipeline. The primary benefit is the potential to overlap parts of $\mathrm{SGEMM}_{k+1}$ with the reducer for slab $k$, thereby reducing idle gaps between SGEMM and custom kernels. Potential benefits include:
\begin{enumerate}
    \item Better GPU utilization. SGEMM is compute-intensive, whereas the reducer kernels are lighter and primarily memory-bound. Interleaving can exploit these complementary resource profiles.

    \item Reduced serialization and pipelined execution. Independent tile/slab units need not wait behind a single stream queue, allowing one stream to consume \texttt{dot\_buf[0]} while the other produces into \texttt{dot\_buf[1]}.

    \item Safe cross-slab accumulation. Stage~A uses \texttt{atomicMin} and Stage~B uses \texttt{atomicAdd}, allowing concurrent updates from multiple streams.
\end{enumerate}

The tradeoff is that ping-pong scheduling is not always faster. If SGEMM already saturates the GPU, there may be little opportunity for overlap. WP-DC therefore treats stream interleaving as a configurable policy: single-stream scheduling is simpler and more reproducible, while ping-pong scheduling aims for higher throughput when useful overlap exists.

\subsubsection{Warp-level reduction vs block-wide shared-memory reduction}

In the proposed DC reduction, we employ warp-level reduction rather than block-wide shared-memory reduction to match the native execution granularity of the GPU. Each warp (32 lanes) is assigned to one DC, and lanes process candidate data points in a stride-32 pattern, yielding fully regular SIMT execution. Partial minima (Stage A) and partial counts (Stage B) are merged via warp shuffle instructions in a logarithmic tree, so inter-thread exchange occurs in registers without block-wide barriers. In contrast, block-level reduction typically requires repeated shared-memory read/write phases and global synchronization via thread-block barriers, which introduces stall time and under-utilization as the active thread set shrinks across reduction levels. Consequently, warp-level reduction reduces synchronization cost, lowers shared-memory pressure, and minimizes reduction latency in the hot path.

This design also improves scalability under tiled/slabbed processing. Let B denote threads per block and W the warp width; moving from thread-wise global updates to warp-aggregated updates reduces global atomic traffic approximately from O(B) to O(B/W) per reduction event, while preserving correctness through packed argmin and atomic accumulation. Because each warp independently resolves one DC reduction, scheduler mapping is straightforward and occupancy remains robust as the number of DCs grows, provided slab sizes keep per-warp work sufficiently large. Overall, the warp-centric formulation is expected to yield higher throughput than block-wide reductions for this workload by co-optimizing synchronization, memory movement, and hardware-aligned parallel decomposition. We therefore use the warp-parallel reduction throughout this work.

\subsection{Cluster Assignment}
\label{sec:cluster-assignment}

Algorithm~\ref{alg:cluster_assignment} follows the label-propagation procedure of Algorithm~\ref{alg:cluster-formation} (Section~\ref{sub:cluster-formation}): the $c$ centers receive distinct labels, and the remaining points, in descending density order, inherit the label of their CDH point. The only addition is a guard (line~9) for a point that has no higher-density point anywhere in the dataset but was not selected as a center. This can occur only when more than $c$ points tie exactly at the global maximum density; such a point is then assigned to the cluster of a selected center of the same density. The case did not arise in any experiment reported in this paper.

\begin{algorithm}[!htbp]
    \caption{Final Cluster Assignment} \label{alg:cluster_assignment}
    \begin{algorithmic}[1]
    \Require $\rho[1..N]$: density array, $CDH\_id[1..N]$: closest density-higher links, $CenterIdx$: indices of the $c$ centers $\mathcal{C}^{*}$
    \Ensure $Label[1..N]$: cluster assignment for all points
    \State $Label[1..N] \gets -1$ \Comment{Initialize all points as unlabeled}
    \State $cluster\_id \gets 0$
    \ForAll{$i \in CenterIdx$}
        \State $Label[i] \gets cluster\_id$ \Comment{Assign unique label to each selected center}
        \State $cluster\_id \gets cluster\_id + 1$
    \EndFor
    \ForAll{points $i$ in descending order of $\rho[i]$}
        \If{$Label[i] = -1$} \Comment{Process unlabeled (non-center) points}
            \If{$CDH\_id[i] = -1$} \Comment{Tied global density maximum, not selected as a center}
                \State $Label[i] \gets 0$ \Comment{Cluster of a selected center of equal density}
            \Else
                \State $Label[i] \gets Label[CDH\_id[i]]$ \Comment{$CDH\_id[i]$ has strictly higher density and is already labeled}
            \EndIf
        \EndIf
    \EndFor
    \end{algorithmic}
\end{algorithm}

\FloatBarrier

\section{Experimental Results}
\label{sec:exp-results}

This section describes the experimental setup, including the system configuration, datasets, and algorithms employed in the evaluation. It subsequently presents the experimental results and analyzes the performance of the proposed algorithm.

\subsection{Experimental Configuration}
\label{sec:exp-config}
The computing environment used to execute and test our algorithm is equipped with an AMD EPYC 7543 32-Core CPU (base clock 2.8 GHz), 256GB of RAM and a NVIDIA RTX A6000 GPU, featuring 10,752 CUDA cores, delivering up to 38.7 TFLOPS of single-precision (FP32) performance, with 48 GB of GDDR6 ECC memory and 768 GB/s of memory bandwidth. The operating system is Rocky Linux 9.7 (Blue Onyx) with the CUDA Toolkit 13.2.2 GPU computing platform installed. All GPU computations in all implementations, including every cuBLAS SGEMM call, use strict FP32 arithmetic; TF32 tensor-core math is disabled.

All reported runtimes are measured as wall-clock time for the FastDEC pipeline, including host-to-device transfers and all computation from $k$-NN construction through cluster assignment, but excluding the one-time cost of loading the dataset from disk into host memory. For each configuration, we perform four runs, discard the first as a warm-up iteration to exclude one-time CUDA context initialization and JIT compilation overheads, and report the mean of the remaining three runs. Across the more than 400 measured implementation--configuration combinations, the coefficient of variation (CV) of the three timed runs was at most 2.35\%, below 1.5\% in 95.7\% of cases, and 0.28\% at the median. This run-to-run variability is far smaller than the narrowest speedup of SymWarp-FastDEC over any baseline reported in this paper ($1.70\times$).

All implementations are compiled for compute capability 8.6 (sm\_86). The \textit{faissgpu} baseline links against FAISS 1.12.0, built from source with its classic GPU kernels (cuVS disabled) using nvcc 13.0 and GCC 11, with Intel MKL 2023.2 as the CPU BLAS backend. DC reduction uses the default single-stream schedule (the optional stream interleaving described in Section~\ref{sec:wp-dc} is disabled), and host-side stages run on a single CPU thread. Every timed stage returns control to the host only after its GPU work has completed, so the host wall-clock timers capture all GPU execution without additional synchronization.
\subsection{Datasets}
\label{sec:datasets}
In our experiments, we evaluate the proposed approach using five different types of datasets.

\subsubsection{Image Patch Datasets}
The image patch datasets are composed of small grayscale image patches, which are collected from unlabelled imagery by cropping out random $n$-by-$n$ chunks. Each $n$-by-$n$ pixel grayscale patch is represented as a vector of $n \times n$ pixel intensities (i.e., $ x(i) \in R^{(n \times n)} $). The method described in the paper \citep{Coates2012learning} is used to normalize the brightness and contrast of the patches: for each $ x(i) $, we subtract out the mean of the intensities and divide by the standard deviation. A small value is added to the variance before division to avoid dividing by zero and also suppress noise. The images used to create the image patches datasets are animal photographs drawn from the Open Images Dataset \citep{kuznetsova2020open}, a large-scale, publicly available collection of web images released by Google. 

\subsubsection{WikiGiga GloVe Datasets}
The GloVe datasets (glove.2024.wikigiga) are downloaded from the Global Vectors for Word Representation (GloVe) project \citep{pennington2014glove} established by the Natural Language Processing Group at Stanford University. GloVe is an unsupervised learning algorithm that produces vector representations of words by leveraging aggregated global word-word co-occurrence statistics from a corpus. The resulting representations exhibit meaningful linear substructures within the word vector space. The datasets used in our experiments contain pre-trained word vectors derived from the 2024 Wikipedia and Gigaword 5 datasets, comprising 11.9 billion tokens and a 1.2 million uncased vocabulary ~\citep{carlson2025new}. These vectors are available in dimensionalities of 50, 100, 200, and 300. We use all four to evaluate how runtime scales with dimensionality at a fixed dataset size, and the dimensionality used in each of the remaining experiments is stated with its results.

\subsubsection{MNIST Datasets}
The MNIST dataset is a standard benchmark consisting of 70,000 grayscale images of handwritten digits (0–9), including 60,000 training images and 10,000 testing images \cite{lecun1998gradient}. In our experiments, the training and testing sets were combined into a single dataset of 70,000 images. The MNIST8M dataset is a large-scale extension of MNIST generated using the InfiMNIST data augmentation procedure \cite{loosli2007training} and contains approximately 8.1 million handwritten digit images. In both datasets, each image has a resolution of (28 $\times$ 28) pixels and is represented as a 784-dimensional feature vector belonging to one of 10 digit classes. MNIST8M is widely used for benchmarking the performance and scalability of large-scale classification and nearest neighbor search algorithms \cite{loosli2007training} due to its size and high dimensionality.

\subsubsection{Deep1B Datasets}
The Deep1B dataset is a widely used benchmark for evaluating billion-scale nearest neighbor search algorithms in machine learning and database systems \cite{babenko2016efficient}. It consists of one billion 96-dimensional image embedding vectors extracted from the ImageNet dataset using a pre-trained GoogLeNet model. The datasets used in our experiments are subsets sampled from the Deep1B dataset.

\subsubsection{Synthetic Datasets}
The synthetic datasets are created by using a random data generator producing a uniform data distribution, so that no natural clustering structure exists. This avoids favoring methods that exploit clustered data.

\subsection{Algorithms Used}
FastDEC provides a framework for density peak clustering in which different $k$-NN search algorithms can be employed at the density estimation step. Our GPU-based implementation retains this flexibility by design. We evaluate the performance of several exact $k$-NN algorithms from well-known open-source libraries against our GPU-accelerated self-$k$-NN algorithm, which is specifically optimized and tailored to the requirements of FastDEC.

We compare the following variants of FastDEC:

\begin{enumerate}
    \item FastDEC-pycpu (\textit{pycpu}): the original CPU-based FastDEC implementation in Python~\citep{yang2022fastdec}.
    \item FastDEC-faissgpu (\textit{faissgpu}): uses the \texttt{GpuIndexFlatL2} index of the GPU-based FAISS library for density estimation.
    \item FastDEC-knncuda (\textit{knncuda}): uses an adapted version of the open-source KnnCuda library for density estimation.
    \item SymWarp-FastDEC (\textit{symwarp}): the algorithm proposed in this paper, which incorporates optimizations specifically designed to accelerate FastDEC on modern GPUs.
\end{enumerate}

Because \textit{pycpu} is a single-threaded, unoptimized Python/$k$-d-tree reference, the large speedups reported against it mainly reflect the cost of that baseline rather than the efficiency of SymWarp-FastDEC itself. We report them to establish that the GPU pipeline is a practical drop-in replacement (same interface and parameters) for the original implementation. The comparisons against the GPU baselines, \textit{faissgpu} and \textit{knncuda}, which are described below, are the more informative measure of the proposed method's efficiency.

FAISS (Facebook AI Similarity Search) is an open-source library for large-scale similarity search and clustering of dense vectors, supporting both exact and approximate nearest-neighbor search on CPUs and GPUs~\citep{faiss,johnson2019billion}. Designed to support similarity search at billion-vector scales, FAISS provides highly optimized GPU implementations that have demonstrated substantial performance improvements through efficient distance computation and $k$-selection~\citep{johnson2019billion}. We use its \texttt{GpuIndexFlatL2} index, which performs exhaustive GPU-accelerated search using squared Euclidean distance, as the exact $k$-NN backend for FastDEC's density estimation stage. To adapt FAISS to FastDEC's self-$k$-NN workload, we retrieve $k+1$ neighbors per point, discard the trivial self-match by index (so that exact duplicates remain eligible), and retain the $k$ nearest remaining points. Since FAISS's \texttt{METRIC\_L2} returns squared Euclidean distances, we apply the square root before writing the results to the transposed $k \times N$ output buffers consumed by the rest of FastDEC. FAISS therefore serves as an established, highly optimized GPU baseline against which we evaluate our proposed self-$k$-NN algorithm for accelerating FastDEC on large-scale datasets.

FastDEC-knncuda uses an adapted version of KnnCuda~\citep{garcia2008fast,garcia2010knn}, an open-source library for exact GPU-based $k$-NN search, in the density estimation stage. KnnCuda identifies the $k$ nearest neighbors in two steps: it first computes all pairwise distances between the query and reference points and stores them in a distance matrix in GPU global memory; it then applies a modified insertion sort to each column of the matrix to extract the $k$ nearest neighbors of each query point. We chose KnnCuda for its conceptual simplicity and exact $k$-NN search. However, its quadratic memory requirement limits its scalability to large datasets. The distance matrix has dimensions $N_{\mathrm{ref}} \times N_{\mathrm{query}}$, which, for FastDEC's self-query workload ($N_{\mathrm{ref}} = N_{\mathrm{query}} = N$), requires $O(N^2)$ memory. For example, at $N = 10^6$, the matrix alone would occupy 4\,TB in single precision, far exceeding the memory capacity of any current GPU. Furthermore, KnnCuda does not explicitly handle self-matches: when the reference and query sets are identical, each point is its own nearest neighbor at distance zero, and this match must be excluded from its own $k$-NN list.

To overcome these limitations, we retain KnnCuda's shared-memory-tiled distance computation kernel and integrate it into a memory-adaptive, batched pipeline for large-scale self-query $k$-NN search. The pipeline performs top-$k$ selection within each batch and then merges the intermediate results using a linear-time merge step. The modifications are as follows:

\begin{enumerate}

\item Native self-query handling\label{knncuda:selfquery}

Rather than duplicating the dataset into separate reference and query buffers, the reference set is used as both reference and query after a single host-to-device transfer. The distance-selection kernel is further enhanced with index-based self-exclusion, so a point cannot be selected as its own nearest neighbor while exact duplicates remain eligible.

\item Memory-adaptive batched distance computation\label{knncuda:batching}

The tiled shared-memory distance kernel is preserved unmodified in its per-tile computation, but it is never invoked against the full $N \times N$ matrix. Instead, the reference set is partitioned into batches of $B$ points, and only a single $B \times N$ distance tile is materialized in device memory at any time:

\begin{itemize}
\item Adaptive batch sizing. Before processing, the implementation derives the largest batch size that fits in the available device memory, aiming to maximize the data processed in each batch. This makes batch size a function of the actual GPU's available memory at runtime, rather than a fixed constant.

\item Per-batch pipeline. For each batch of reference points: (i) the tiled shared-memory kernel computes a $B \times N$ squared-distance tile against all $N$ query points; (ii) the modified insertion-sort kernel (point~\ref{knncuda:selection} below) reduces each column to its $k$ smallest entries within the batch; (iii) a linear-merge kernel (point~\ref{knncuda:merge} below) folds the batch's top-$k$ list into the running, globally accumulated top-$k$ list per query point, offsetting batch-relative indices to global reference indices.
\end{itemize}

This reduces device memory for the distance structure from $O(N^2)$ to $O(B \times N + k \times N)$, where $B$ is bounded only by available free memory, allowing the same kernel to remain memory-feasible for larger $N$ by shrinking $B$; this resolves the memory bottleneck but does not change the underlying $O(N^2)$ time complexity of the distance computation.

\item Top-$k$ selection within a batch\label{knncuda:selection}

The top-$k$ candidates of each batch are selected with KnnCuda's original modified insertion sort, extended only with the self-match exclusion described in point~\ref{knncuda:selfquery}. For each query column, one thread scans the batch's candidates in reference-index order and maintains a sorted list of the $k$ smallest distances seen so far. A candidate that is not smaller than the current $k$-th entry is rejected with a single comparison; otherwise, larger entries are shifted down and the candidate is written into its sorted position. The scan requires $O(B)$ threshold tests, plus up to $O(k)$ additional comparisons and shifts per accepted candidate, giving $O(Bk)$ work per query column in the worst case. Because the sorted list resides in global memory and each accepted candidate may shift up to $k$ entries, the selection cost grows with $k$. We deliberately retain this original selection kernel rather than replacing it with an asymptotically faster alternative such as heap-based selection, so that FastDEC-knncuda remains faithful to the published KnnCuda algorithm and its modifications are confined to what is needed to scale it to large self-query workloads. FastDEC-faissgpu, which builds on the highly optimized FAISS library, serves as the complementary state-of-the-art GPU baseline.

\item Linear merge of accumulated and batch results\label{knncuda:merge}

After each batch's top-$k$ candidates are selected, they must be merged with the top-$k$ list accumulated from all previously processed batches. Since both the batch result and the running accumulated result are already sorted in ascending order per query column, the merge is performed as a single $O(k)$ linear merge of two sorted lists (analogous to the merge step of merge-sort) rather than by re-inserting each batch candidate one at a time — the latter would cost up to $O(k^2)$ per query column in the worst case. The merge writes its output into a separate pair of buffers, and these buffers are then pointer-swapped (``ping-ponged'') with the accumulated-result buffers before the next batch, avoiding a copy step.

\end{enumerate}

After applying these optimizations, FastDEC-knncuda preserves KnnCuda's original tiled distance kernel (widening only its index arithmetic to 64 bits for large tiles), while removing its $O(N^2)$ memory ceiling through adaptive batching combined with an efficient batch-accumulation pipeline (per-batch insertion-sort selection paired with linear merging), and further incorporates the self-match exclusion required for a correct self-query $k$-NN.

The results presented in this paper were obtained by running the FastDEC-pycpu, FastDEC-knncuda, FastDEC-faissgpu, and SymWarp-FastDEC implementations on datasets of varying sizes and dimensionalities. FastDEC-pycpu, the original implementation, was evaluated only on relatively small datasets, as its runtime becomes prohibitive on medium and large datasets.

\subsection{Performance Evaluation}
We now present the empirical results obtained using the aforementioned algorithms across four real-world domains and one synthetic domain, comparing them against relevant baselines. Performance is evaluated using several metrics, including execution time, speedup, and scalability (with respect to size and dimensionality).

We first present the results of our DC Reduction acceleration algorithm on different datasets. Next, we compare our algorithm against the FastDEC variants described above - implemented using the FAISS and KnnCuda libraries - in terms of execution time across various real-world and synthetic datasets. Finally, we examine the impact of different values of k and c on these FastDEC variants. To ensure a fair comparison, all data used were generated under the same experimental configuration; that is, all algorithms within a given experiment use identical values of k and c, the same DC reduction algorithm, and the same L2 (Euclidean) distance metric.

\subsubsection{Effectiveness of DC Reduction}
Results are presented in this section to evaluate the effectiveness and efficiency of the proposed Dominance Component (DC) reduction algorithm (i.e., DC inter-dominance estimation). Table~\ref{tab:fastdec-dcr-large-ds} reports the runtime of the DC reduction step and its proportion of the total runtime for the SymWarp-FastDEC implementation—shown in later experiments to be the fastest variant—across nine large-scale datasets exceeding one million instances. Table~\ref{tab:fastdec-dcr-small-ds} reports the corresponding results on six smaller datasets spanning a broad range of dimensionalities, from 64 to 16,384, allowing the algorithm's sensitivity to dimensionality to be assessed independently of dataset size. Across both tables, the DC reduction step consistently accounts for only a small fraction of the total runtime, ranging from 0.12\% to 7.36\% on the large-scale datasets and from 0.23\% to 1.68\% on the smaller, higher-dimensional datasets. The two large MNIST-derived datasets exhibit the highest shares, 2.47\% on mnist\_2m\_784d and 7.36\% on mnist\_8m\_784d, the largest overhead among all tested datasets; even in this case, however, the DC reduction step remains a minor component of the overall runtime.

\begin{table*}[t]
    \centering
    \caption{Runtime of \textit{SymWarp-FastDEC} and time spent in Dominance Component (DC) reduction across large-scale datasets (all with at least 1M instances), ordered by number of points. DC reduction (\%) denotes the proportion of total runtime consumed by the DC reduction step.}
    \label{tab:fastdec-dcr-large-ds}
    \setlength{\tabcolsep}{4pt}
    \renewcommand{\arraystretch}{0.90}
    \scriptsize
    \resizebox{\textwidth}{!}{%
        \begin{tabular}{l rrrrrrrrr}
            \toprule
            Dataset & \texttt{syn\_1m\_1536d} & \texttt{glove\_1.2m\_300d} & \texttt{deep\_2m\_96d} & \texttt{mnist\_2m\_784d} & \texttt{syn\_2m\_1536d} & \texttt{deep\_5m\_96d} & \texttt{mnist\_8m\_784d} & \texttt{syn\_10m\_64d} & \texttt{syn\_10m\_512d} \\
            \midrule
            Runtime (s) & 66.66 & 30.82 & 41.57 & 128.79 & 240.78 & 236.77 & 2121.55 & 814.36 & 2278.88 \\
            DC reduction (s) & 0.11 & 0.04 & 0.20 & 3.18 & 0.31 & 0.66 & 156.14 & 1.55 & 2.84 \\
            DC reduction (\%) & 0.17 & 0.13 & 0.48 & 2.47 & 0.13 & 0.28 & 7.36 & 0.19 & 0.12 \\
            \bottomrule
        \end{tabular}%
    }
    \vspace{-1.2mm}
\end{table*}

\begin{figure}[t]
\centering
\begin{minipage}[t]{.48\textwidth}
  \centering
  \includegraphics[width=1\linewidth]{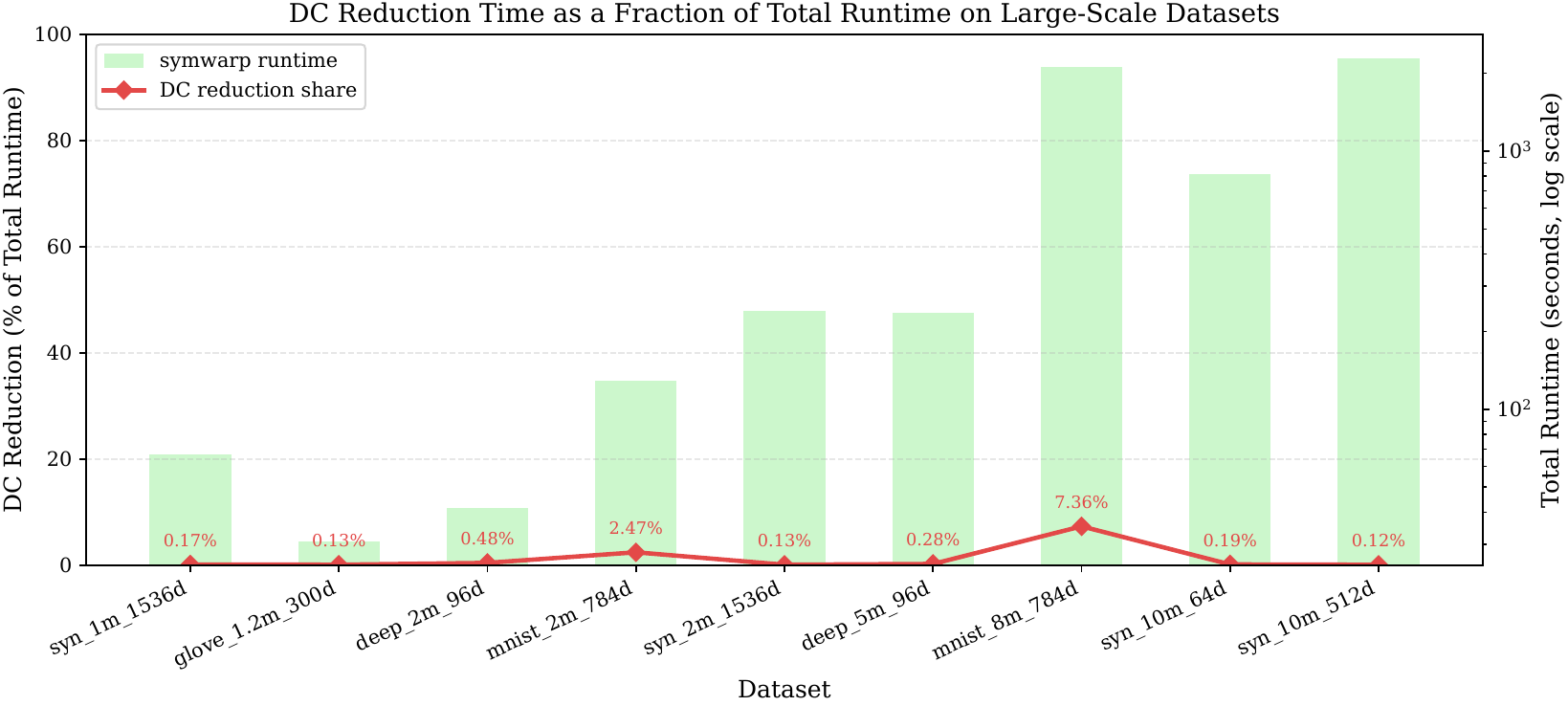}
  \caption{DC reduction time as a fraction of the total \textit{SymWarp-FastDEC} runtime on large-scale datasets, each containing more than one million instances. The line (left axis) shows the percentage of total runtime spent on DC reduction, while the bars (right axis, logarithmic scale) show the corresponding total runtimes.}
  \label{fig:dcr-large-datasets}
\end{minipage}%
\hfill
\begin{minipage}[t]{.48\textwidth}
  \centering
  \includegraphics[width=1\linewidth]{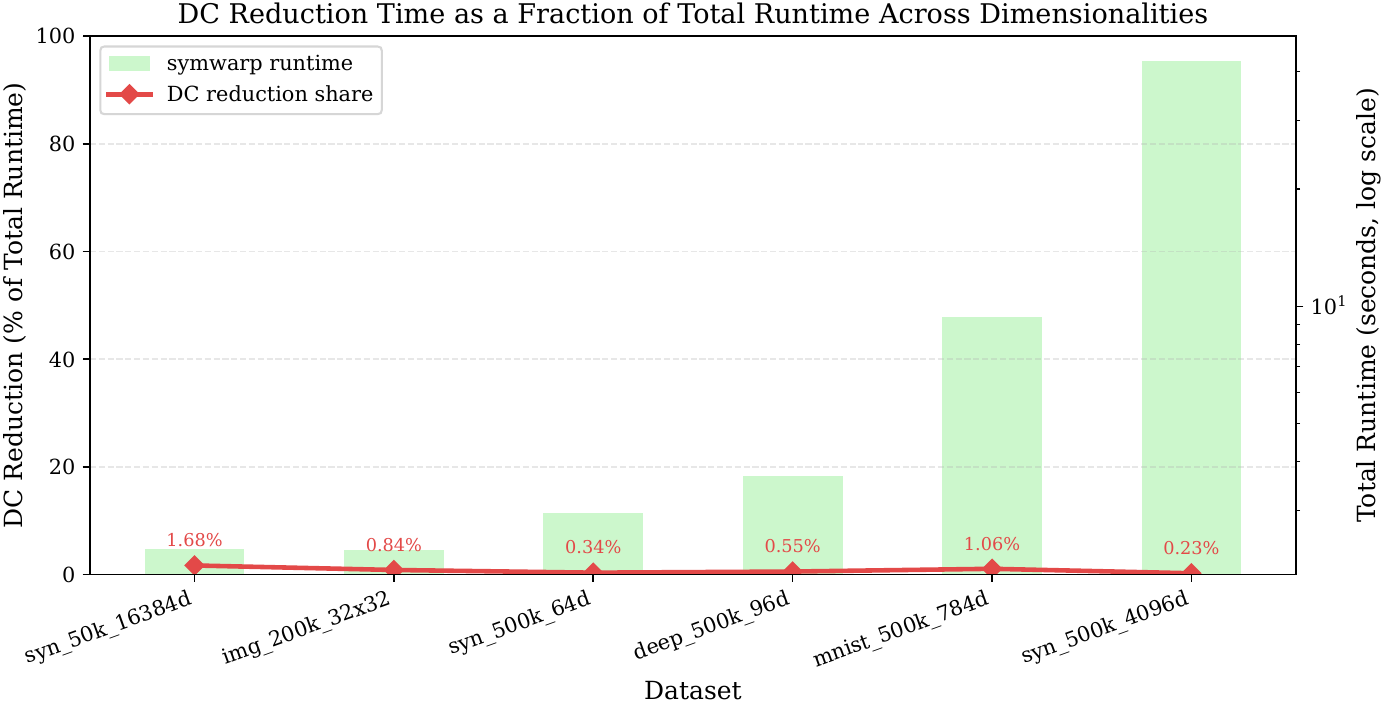}
  \caption{DC reduction time as a fraction of the total \textit{SymWarp-FastDEC} runtime on datasets with fewer instances and spanning a wide range of dimensionalities (64--16,384). The line (left axis) shows the percentage of total runtime spent on DC reduction, while the bars (right axis, logarithmic scale) show the corresponding total runtimes.}
  \label{fig:dcr-small-datasets}
\end{minipage}
\end{figure}

Figures~\ref{fig:dcr-large-datasets} and \ref{fig:dcr-small-datasets} visualize these results for the large-scale and smaller datasets, respectively, plotting the DC reduction share (line, left axis) alongside the total runtime (bar, right axis, log scale) for each dataset. In both figures, the DC reduction share stays close to the bottom of the axis even as the runtime bars span roughly three orders of magnitude across the two figures (2.37\,s to 2,278.88\,s), visually reinforcing that DC reduction is not a runtime bottleneck regardless of dataset size or dimensionality.

\begin{table*}[t]
    \centering
    \caption{Runtime of \textit{SymWarp-FastDEC} and time spent in Dominance Component (DC) reduction across smaller datasets of varying dimensionality, ordered by number of points and then by dimensionality. DC reduction (\%) denotes the proportion of total runtime consumed by the DC reduction step.}
    \label{tab:fastdec-dcr-small-ds}
    \setlength{\tabcolsep}{4pt}
    \renewcommand{\arraystretch}{0.90}
    \footnotesize
    \resizebox{\textwidth}{!}{%
        \begin{tabular}{l rrrrrr}
            \toprule
            Dataset & \texttt{syn\_50k\_16384d} & \texttt{img\_200k\_32x32} & \texttt{syn\_500k\_64d} & \texttt{deep\_500k\_96d} & \texttt{mnist\_500k\_784d} & \texttt{syn\_500k\_4096d} \\
            \midrule
            Runtime (s) & 2.38 & 2.37 & 2.95 & 3.66 & 9.42 & 42.75 \\
            DC reduction (s) & 0.04 & 0.02 & 0.01 & 0.02 & 0.10 & 0.10 \\
            DC reduction (\%) & 1.68 & 0.84 & 0.34 & 0.55 & 1.06 & 0.23 \\
            \bottomrule
        \end{tabular}%
    }
    \vspace{-1.2mm}
\end{table*}

To further demonstrate the effectiveness of the proposed GPU-accelerated DC reduction algorithm, WP-DC (labelled \textit{wp-dc} in the table and figure), Table~\ref{tab:dcr-comparison} compares its runtime proportion against that of the original CPU-based DC reduction (py-dc) across six datasets spanning both low (2–8) and high (512–784) dimensionality. For each dataset, the proportion of total runtime consumed by DC reduction is reported for the corresponding FastDEC implementation — SymWarp-FastDEC (GPU-accelerated) and FastDEC-pycpu (CPU-based).

\begin{figure*}[t]
\centering
\includegraphics[width=\textwidth]{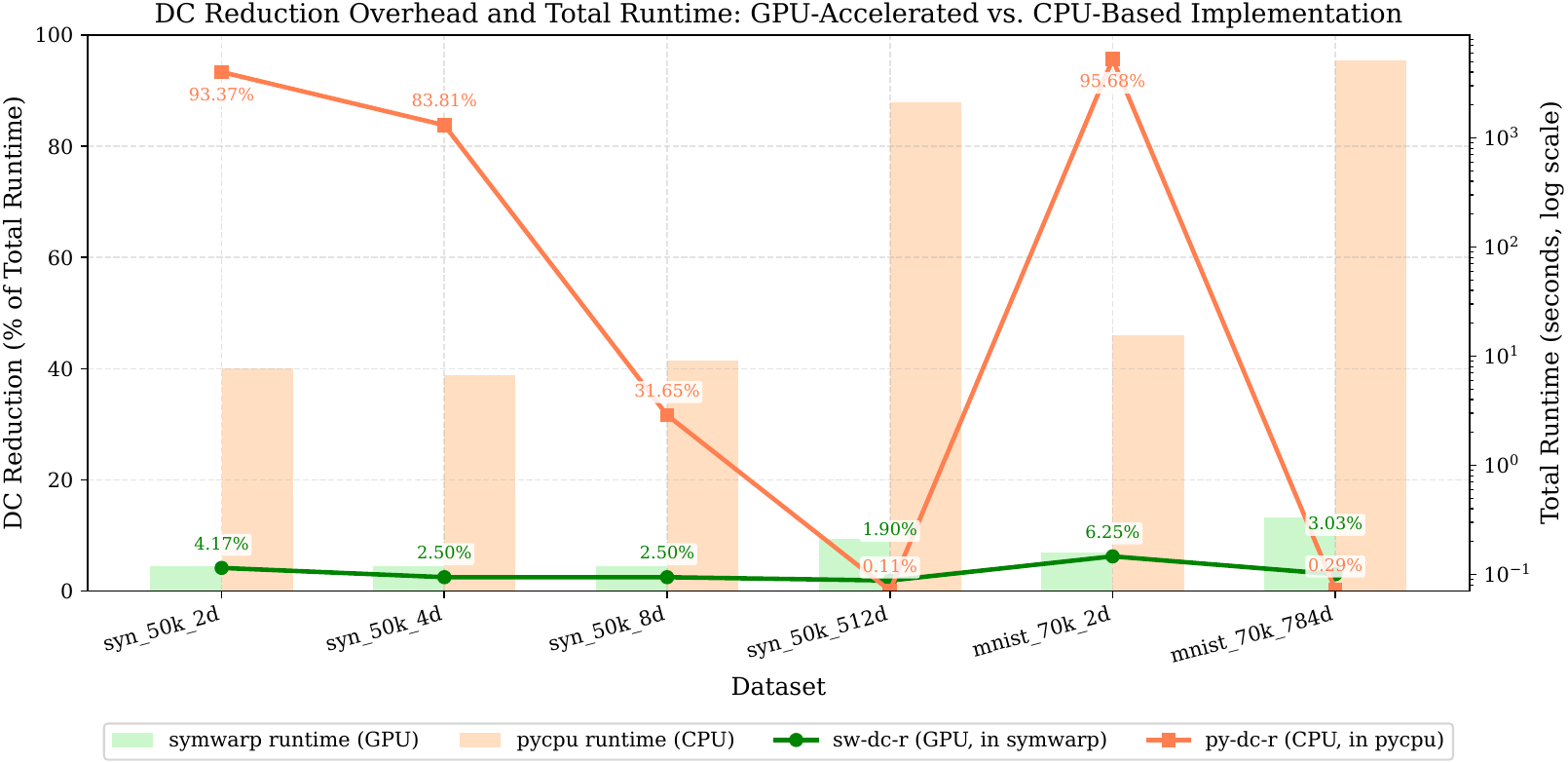}
\caption{DC reduction overhead and total runtime for the GPU-accelerated implementation (\textit{wp-dc}, in \textit{SymWarp-FastDEC}) and the CPU-based implementation (\textit{py-dc}, in \textit{FastDEC-pycpu}) across six datasets with varying dimensionalities. The lines (left axis) show the percentage of total runtime spent on DC reduction, while the bars (right axis, logarithmic scale) show the corresponding total runtimes.}
\label{fig:dcr-gpu-vs-cpu}
\end{figure*}

The results show that WP-DC consistently accounts for a small share of the total SymWarp-FastDEC runtime, ranging from 1.90\% to 6.25\%, regardless of dimensionality. These shares are somewhat higher than on the larger datasets above because SymWarp-FastDEC completes each of these datasets in only 0.12--0.33\,s, so that even a WP-DC time of 3--10\,ms registers as a few percent; in absolute terms, WP-DC is roughly three orders of magnitude faster than py-dc on every dataset (e.g., 0.010\,s versus 14.84\,s on mnist\_70k\_2d). In contrast, py-dc constitutes a substantially larger share of the FastDEC-pycpu runtime on low-dimensional datasets, reaching as high as 95.68\% on mnist\_70k\_2d, indicating that DC reduction is the dominant computational bottleneck in the CPU-based implementation at low dimensionality. On the two high-dimensional datasets (syn\_50k\_512d and mnist\_70k\_784d), this proportion drops sharply to 0.11\% and 0.29\%, respectively; however, this decrease reflects a shift in bottleneck rather than an improvement in DC reduction efficiency, consistent with the earlier observation that the exact, $k$-d-tree-based density estimation step becomes prohibitively expensive beyond roughly 10--20 dimensions, causing the total pycpu runtime (2,098 and 5,125 seconds, respectively) to be dominated by density estimation rather than DC reduction. Taken together, these results confirm that the proposed WP-DC algorithm effectively eliminates DC reduction as a runtime bottleneck across all tested dimensionalities, whereas the original CPU-based approach remains bottlenecked by DC reduction whenever density estimation itself remains tractable. Figure~\ref{fig:dcr-gpu-vs-cpu} visualizes these results, plotting each implementation's DC reduction share (lines, left axis) alongside its total runtime (bars, right axis, log scale). The figure makes the contrast concrete: wp-dc's share stays low (at most 6.25\%) regardless of dimensionality, while py-dc's share is high precisely on the four low-dimensional datasets where pycpu's total runtime is small, and collapses only on the two high-dimensional datasets where the total runtime instead balloons by orders of magnitude.

\begin{table*}[t]
    \centering
    \caption{Contribution of Dominance Component (DC) reduction to total runtime in the GPU-accelerated (\textit{wp-dc}, within \textit{SymWarp-FastDEC}) and CPU-based (\textit{py-dc}, within \textit{FastDEC-pycpu}) implementations. Runtime and the proportion of total runtime consumed by DC reduction are reported, with datasets ordered by number of points and then by dimensionality.}
    \label{tab:dcr-comparison}
    \setlength{\tabcolsep}{4pt}
    \renewcommand{\arraystretch}{0.90}
    \footnotesize
    \resizebox{\textwidth}{!}{%
        \begin{tabular}{l rrrrrr}
            \toprule
            Dataset & \texttt{syn\_50k\_2d} & \texttt{syn\_50k\_4d} & \texttt{syn\_50k\_8d} & \texttt{syn\_50k\_512d} & \texttt{mnist\_70k\_2d} & \texttt{mnist\_70k\_784d} \\
            \midrule
            \textit{symwarp} runtime (s) & 0.12 & 0.12 & 0.12 & 0.21 & 0.16 & 0.33 \\
            \textit{wp-dc} runtime (s) & 0.005 & 0.003 & 0.003 & 0.004 & 0.010 & 0.010 \\
            \textit{wp-dc} proportion (\%) & 4.17 & 2.50 & 2.50 & 1.90 & 6.25 & 3.03 \\
            \midrule
            \textit{pycpu} runtime (s) & 7.69 & 6.67 & 9.10 & 2098.24 & 15.51 & 5125.48 \\
            \textit{py-dc} runtime (s) & 7.18 & 5.59 & 2.88 & 2.29 & 14.84 & 14.91 \\
            \textit{py-dc} proportion (\%) & 93.37 & 83.81 & 31.65 & 0.11 & 95.68 & 0.29 \\
            \bottomrule
        \end{tabular}%
    }
    \vspace{-1.2mm}
\end{table*}

These results demonstrate that the proposed GPU-accelerated DC reduction algorithm performs consistently and efficiently across datasets of varying size and dimensionality. Accordingly, the same DC reduction algorithm is adopted across all GPU-accelerated FastDEC implementations in the experiments that follow, enabling a fair comparison of these variants with respect to density estimation performance.

\subsubsection{Evaluation on MNIST Subsets}

\begin{figure}[t]
\centering
\begin{minipage}[t]{.48\textwidth}
  \centering
  \includegraphics[width=1\linewidth]{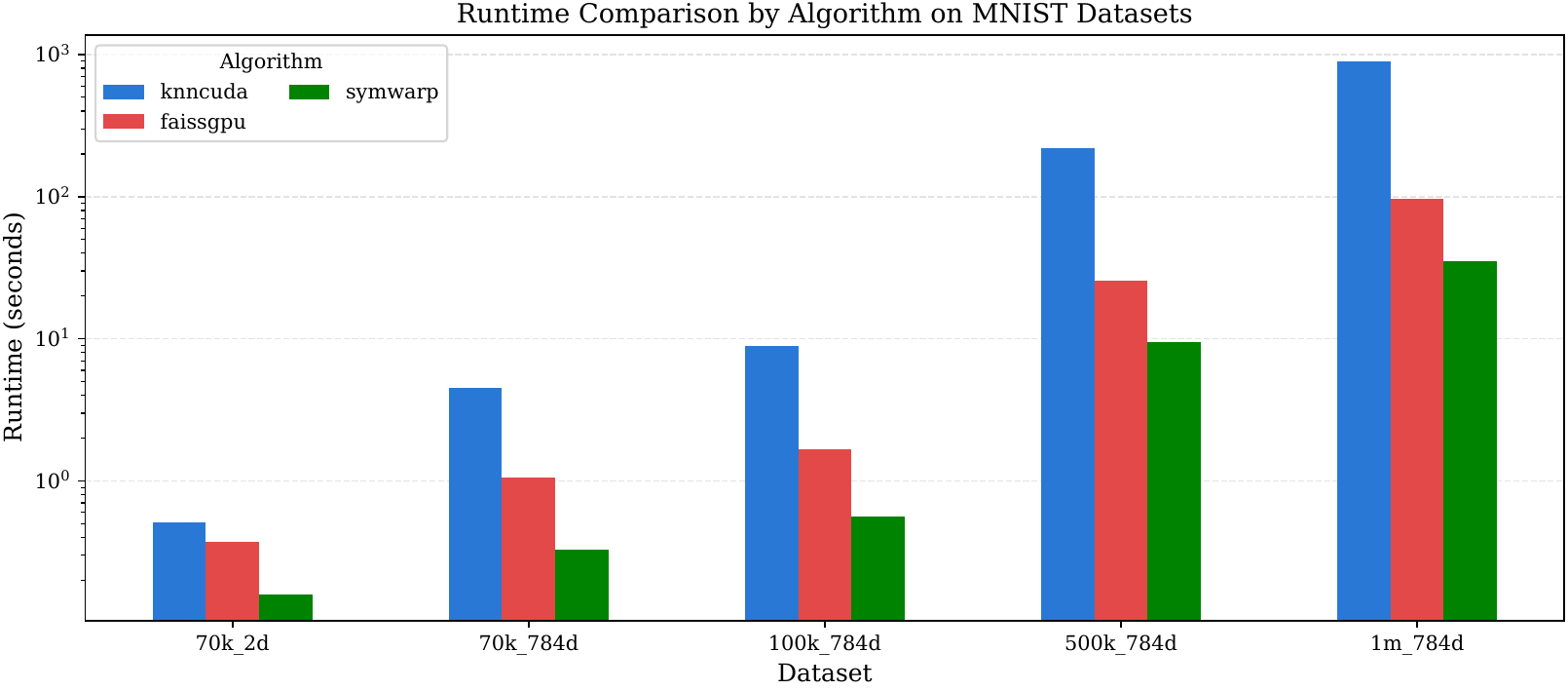}
  \caption{Runtime comparison of \textit{knncuda}, \textit{faissgpu}, and \textit{symwarp} on MNIST datasets of increasing size ($c=10$, $k=25$). Runtime is shown on a logarithmic scale.}
  \label{fig:mnist-runtime}
\end{minipage}%
\hfill
\begin{minipage}[t]{.48\textwidth}
  \centering
  \includegraphics[width=1\linewidth]{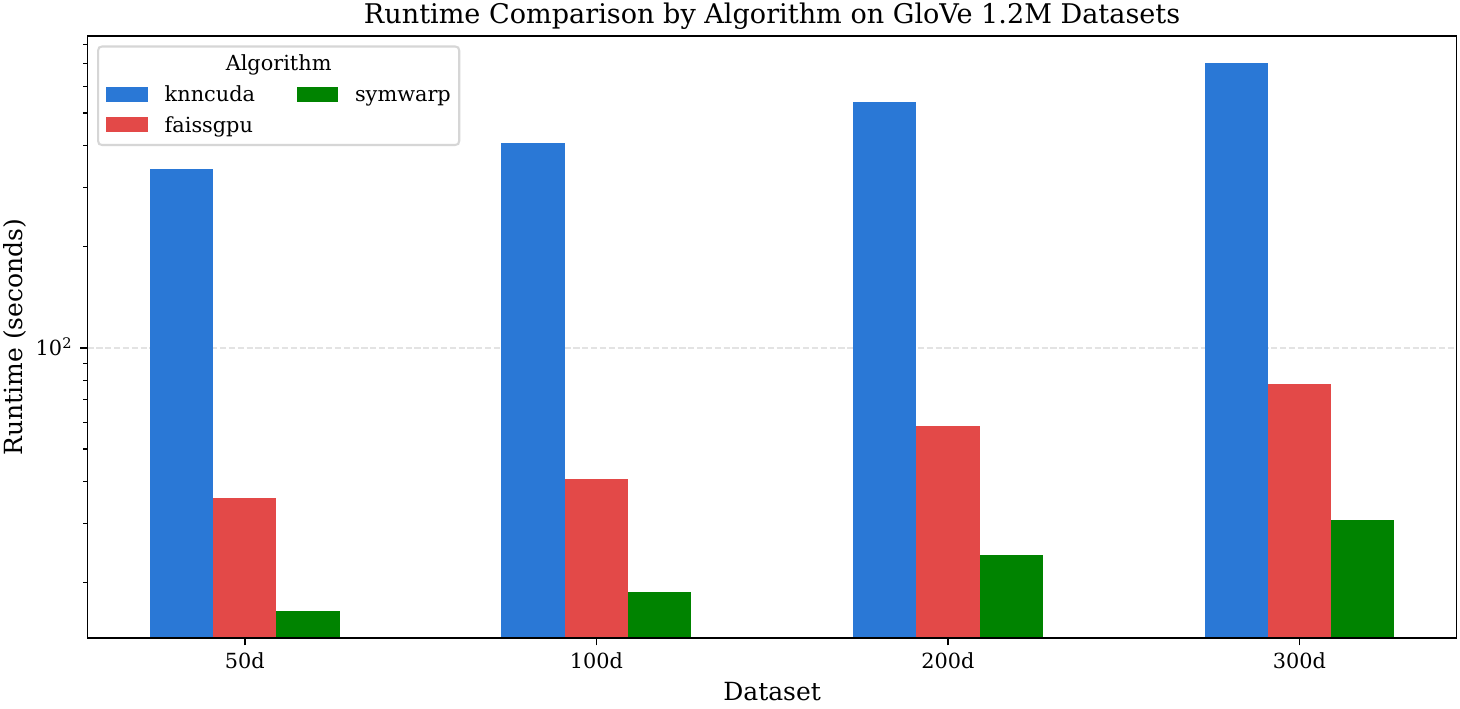}
  \caption{Runtime comparison of \textit{knncuda}, \textit{faissgpu}, and \textit{symwarp} across GloVe datasets of increasing embedding dimensionality ($c=20$, $k=45$). Runtime is shown on a logarithmic scale.}
  \label{fig:glove-dim-runtime}
\end{minipage}
\end{figure}

We additionally evaluate five MNIST-derived datasets ranging from 70k to 1m instances: MNIST2D~\cite{ma2024up}, a UMAP-based 2-dimensional projection of the original MNIST dataset, and four native 784-dimensional datasets of increasing size. Table~\ref{tab:mnist-runtime-speedup} and Figure~\ref{fig:mnist-runtime} show that \textit{symwarp} achieves the lowest runtime across all datasets. Relative to the \textit{knncuda} baseline, its speedup increases monotonically from $3.19\times$ on 70k\_2d to $25.16\times$ on 1m\_784d, demonstrating increasingly favorable scaling behavior across the evaluated datasets. \textit{faissgpu} is also faster than \textit{knncuda} on every dataset, but by a smaller margin, from $1.38\times$ on 70k\_2d to $9.23\times$ on 1m\_784d. Relative to \textit{faissgpu}, \textit{symwarp} is $2.31$--$3.18\times$ faster; at a fixed 70k instances, moving from 2d to 784d widens this margin from $2.31\times$ to $3.18\times$, whereas on the two largest datasets it settles at $2.73\times$. These results suggest that the advantage of \textit{symwarp} over \textit{knncuda} grows with both dataset size and dimensionality, while its margin over \textit{faissgpu} remains comparatively stable across the evaluated sizes.

\begin{table*}[t]
\centering
\caption{Runtime ($T$, in seconds) comparison across MNIST subsets. Speedup (SUP) is computed relative to the \textit{knncuda} baseline. The fastest runtime for each dataset is shown in bold.}
\label{tab:mnist-runtime-speedup}
\setlength{\tabcolsep}{2.6pt}
\renewcommand{\arraystretch}{0.90}
\footnotesize
\resizebox{\textwidth}{!}{%
\begin{tabular}{l rr rr rr rr rr}
\toprule
\multirow{2}{*}{Algorithm} & \multicolumn{2}{c}{70k\_2d} & \multicolumn{2}{c}{70k\_784d} & \multicolumn{2}{c}{100k\_784d} & \multicolumn{2}{c}{500k\_784d} & \multicolumn{2}{c}{1m\_784d} \\
\cmidrule(lr){2-3}\cmidrule(lr){4-5}\cmidrule(lr){6-7}\cmidrule(lr){8-9}\cmidrule(lr){10-11}
& T & SUP & T & SUP & T & SUP & T & SUP & T & SUP \\
\midrule
\textit{knncuda} & 0.51 & $1.00\times$ & 4.50 & $1.00\times$ & 8.87 & $1.00\times$ & 220.66 & $1.00\times$ & 889.33 & $1.00\times$ \\
\textit{faissgpu} & 0.37 & $1.38\times$ & 1.05 & $4.29\times$ & 1.66 & $5.34\times$ & 25.67 & $8.60\times$ & 96.38 & $9.23\times$ \\
\textit{symwarp} & \textbf{0.16} & \textbf{$3.19\times$} & \textbf{0.33} & \textbf{$13.64\times$} & \textbf{0.56} & \textbf{$15.84\times$} & \textbf{9.42} & \textbf{$23.42\times$} & \textbf{35.34} & \textbf{$25.16\times$} \\
\bottomrule
\end{tabular}%
}
\vspace{-1.2mm}
\end{table*}

\subsubsection{Evaluation on GloVe Datasets}

Across four GloVe dimensionalities at a fixed dataset size of approximately 1.2M vectors (Table~\ref{tab:glove-dim-speedup}, Figure~\ref{fig:glove-dim-runtime}), \textit{symwarp} achieves the lowest runtime at every dimensionality, with its runtime increasing from 16.49\,s to 30.82\,s ($1.9\times$) as dimensionality grows from 50d to 300d. Over the same range, the runtimes of \textit{knncuda} and \textit{faissgpu} increase slightly more, from 340.68\,s to 702.07\,s ($2.1\times$) and from 35.76\,s to 78.24\,s ($2.2\times$), respectively. Consequently, \textit{symwarp}'s speedup over \textit{knncuda} rises only slightly, from $20.66\times$ at 50d to $22.78\times$ at 300d, while its advantage over \textit{faissgpu} increases from $2.17\times$ to $2.54\times$. Within this comparatively narrow range of dimensionality, the effect of dimensionality on the relative performance of the three methods is therefore modest; the image-patch and synthetic experiments below examine much wider dimensionality ranges.

\begin{table*}[t]
    \centering
    \caption{Runtime ($T$, in seconds) comparison across GloVe datasets. Speedup (SUP) is computed relative to the \textit{knncuda} baseline. The fastest runtime for each dataset is shown in bold.}
    \label{tab:glove-dim-speedup}
    \setlength{\tabcolsep}{2.6pt}
    \renewcommand{\arraystretch}{0.90}
    \footnotesize
    \resizebox{\textwidth}{!}{%
        \begin{tabular}{l rr rr rr rr}
            \toprule
            \multirow{2}{*}{Algorithm} & \multicolumn{2}{c}{50d} & \multicolumn{2}{c}{100d} & \multicolumn{2}{c}{200d} & \multicolumn{2}{c}{300d} \\
            \cmidrule(lr){2-3} \cmidrule(lr){4-5} \cmidrule(lr){6-7} \cmidrule(lr){8-9}
                & T & SUP & T & SUP & T & SUP & T & SUP \\
            \midrule
            \textit{knncuda}  & 340.68 & $1.00\times$ & 405.68 & $1.00\times$ & 539.57 & $1.00\times$ & 702.07 & $1.00\times$ \\
            \textit{faissgpu} & 35.76  & $9.53\times$ & 40.63  & $9.98\times$ & 58.59  & $9.21\times$ & 78.24  & $8.97\times$ \\
            \textit{symwarp}  & \textbf{16.49} & \textbf{$20.66\times$} & \textbf{18.72} & \textbf{$21.67\times$} & \textbf{24.14} & \textbf{$22.35\times$} & \textbf{30.82} & \textbf{$22.78\times$} \\
            \bottomrule
        \end{tabular}%
    }
    \vspace{-1.2mm}
\end{table*}

\subsubsection{Evaluation on Deep1B Subsets}

\begin{figure}[t]
\centering
\begin{minipage}[t]{.48\textwidth}
  \centering
  \includegraphics[width=1\linewidth]{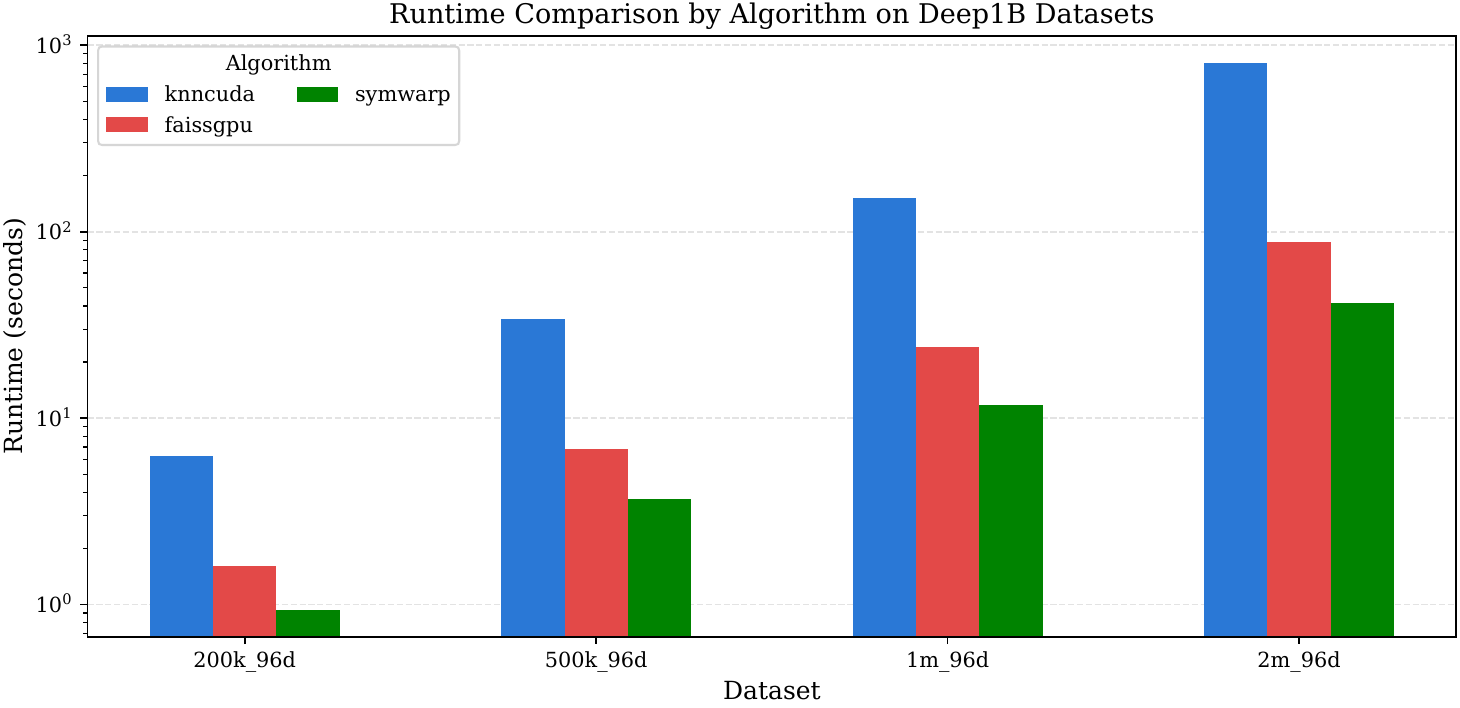}
  \caption{Runtime comparison of \textit{knncuda}, \textit{faissgpu}, and \textit{symwarp} on Deep1B subsets of increasing size ($c=80$, $k=50$). Runtime is shown on a logarithmic scale.}
  \label{fig:deep1b-runtime}
\end{minipage}%
\hfill
\begin{minipage}[t]{.48\textwidth}
  \centering
  \includegraphics[width=1\linewidth]{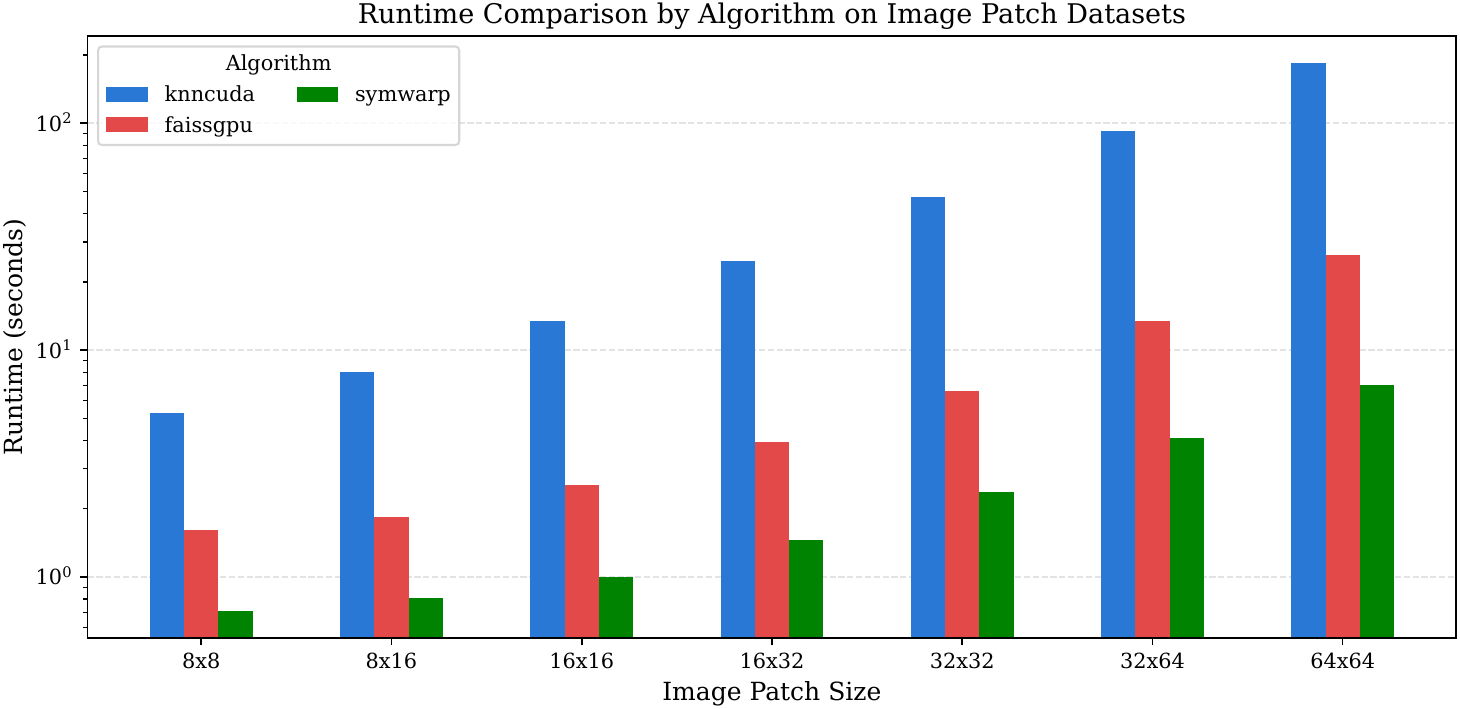}
  \caption{Runtime comparison of \textit{knncuda}, \textit{faissgpu}, and \textit{symwarp} across image-patch datasets of increasing dimensionality ($c=64$, $k=25$). Runtime is shown on a logarithmic scale.}
  \label{fig:image-patch-runtime}
\end{minipage}
\end{figure}

On the Deep1B subsets, which maintain a fixed dimensionality of 96 while varying in size from 200k to 2m instances (Table~\ref{tab:deep1b-runtime-speedup}, Figure~\ref{fig:deep1b-runtime}), \textit{symwarp} consistently achieves the lowest runtime. As dataset size increases, the runtime of \textit{symwarp} grows from 0.94\,s to 41.57\,s, compared with increases from 6.25\,s to 799.65\,s for \textit{knncuda} and from 1.60\,s to 88.03\,s for \textit{faissgpu}. Consequently, \textit{symwarp}'s speedup over \textit{knncuda} increases from $6.65\times$ at 200k\_96d to $19.24\times$ at 2m\_96d, while its advantage over \textit{faissgpu} grows steadily from $1.70\times$ to $2.12\times$. These results provide further evidence that the scale-driven trend observed previously persists at fixed dimensionality, with the relative performance advantages of \textit{symwarp} becoming increasingly pronounced as dataset size increases.

\begin{table*}[t]
    \centering
    \caption{Runtime ($T$, in seconds) comparison across Deep1B subsets. Speedup (SUP) is computed relative to the \textit{knncuda} baseline. The fastest runtime for each dataset is shown in bold.}
    \label{tab:deep1b-runtime-speedup}
    \setlength{\tabcolsep}{2.6pt}
    \renewcommand{\arraystretch}{0.90}
    \footnotesize
    \resizebox{\textwidth}{!}{%
        \begin{tabular}{l rr rr rr rr}
            \toprule
            \multirow{2}{*}{Algorithm} & \multicolumn{2}{c}{200k\_96d} & \multicolumn{2}{c}{500k\_96d} & \multicolumn{2}{c}{1m\_96d} & \multicolumn{2}{c}{2m\_96d} \\
            \cmidrule(lr){2-3} \cmidrule(lr){4-5} \cmidrule(lr){6-7} \cmidrule(lr){8-9}
                & T & SUP & T & SUP & T & SUP & T & SUP \\
            \midrule
            \textit{knncuda}  & 6.25 & $1.00\times$ & 33.94 & $1.00\times$ & 151.31 & $1.00\times$ & 799.65 & $1.00\times$ \\
            \textit{faissgpu} & 1.60 & $3.91\times$ & 6.87 & $4.94\times$ & 23.98 & $6.31\times$ & 88.03 & $9.08\times$ \\
            \textit{symwarp}  & \textbf{0.94} & $6.65\times$ & \textbf{3.66} & $9.27\times$ & \textbf{11.72} & $12.91\times$ & \textbf{41.57} & $19.24\times$ \\
            \bottomrule
        \end{tabular}%
    }
    \vspace{-1.2mm}
\end{table*}

\subsubsection{Evaluation on Image-Patch Datasets}

On seven image-patch datasets with a fixed 200k instances but increasing patch size (and therefore dimensionality) from $8\times8$ (64d) to $64\times64$ (4096d) (Table~\ref{tab:image-patch-runtime-speedup}, Figure~\ref{fig:image-patch-runtime}), \textit{symwarp} consistently achieves the lowest runtime. As dimensionality increases, its runtime grows from 0.71\,s to 7.02\,s ($9.9\times$), compared with increases from 5.27\,s to 183.62\,s ($35\times$) for \textit{knncuda} and from 1.62\,s to 26.29\,s ($16\times$) for \textit{faissgpu}. Consequently, \textit{symwarp}'s speedup over \textit{knncuda} increases from $7.42\times$ at 64d to $26.16\times$ at 4096d, while its advantage over \textit{faissgpu} grows from $2.28\times$ to $3.75\times$. These results demonstrate substantially more favorable scaling of \textit{symwarp} with feature dimensionality at fixed dataset cardinality.

\begin{table*}[t]
    \centering
    \caption{Runtime ($T$, in seconds) comparison across image-patch datasets. Speedup (SUP) is computed relative to the \textit{knncuda} baseline. The fastest runtime for each dataset is shown in bold.}
    \label{tab:image-patch-runtime-speedup}
    \setlength{\tabcolsep}{2.2pt}
    \renewcommand{\arraystretch}{0.88}
    \footnotesize
    \resizebox{\textwidth}{!}{%
        \begin{tabular}{l rr rr rr rr rr rr rr}
            \toprule
            \multirow{2}{*}{Algorithm} & \multicolumn{2}{c}{8x8} & \multicolumn{2}{c}{8x16} & \multicolumn{2}{c}{16x16} & \multicolumn{2}{c}{16x32} & \multicolumn{2}{c}{32x32} & \multicolumn{2}{c}{32x64} & \multicolumn{2}{c}{64x64} \\
            \cmidrule(lr){2-3} \cmidrule(lr){4-5} \cmidrule(lr){6-7} \cmidrule(lr){8-9} \cmidrule(lr){10-11} \cmidrule(lr){12-13} \cmidrule(lr){14-15}
                & T & SUP & T & SUP & T & SUP & T & SUP & T & SUP & T & SUP & T & SUP \\
            \midrule
            \textit{knncuda}  & 5.27 & $1.00\times$ & 7.99 & $1.00\times$ & 13.50 & $1.00\times$ & 24.73 & $1.00\times$ & 47.35 & $1.00\times$ & 92.74 & $1.00\times$ & 183.62 & $1.00\times$ \\
            \textit{faissgpu} & 1.62 & $3.25\times$ & 1.84 & $4.34\times$ & 2.54 & $5.31\times$ & 3.92 & $6.31\times$ & 6.58 & $7.20\times$ & 13.45 & $6.90\times$ & 26.29 & $6.98\times$ \\
            \textit{symwarp}  & \textbf{0.71} & \textbf{$7.42\times$} & \textbf{0.81} & \textbf{$9.86\times$} & \textbf{1.00} & \textbf{$13.50\times$} & \textbf{1.45} & \textbf{$17.06\times$} & \textbf{2.37} & \textbf{$19.98\times$} & \textbf{4.10} & \textbf{$22.62\times$} & \textbf{7.02} & \textbf{$26.16\times$} \\
            \bottomrule
        \end{tabular}%
    }
    \vspace{-1.2mm}
\end{table*}

\subsubsection{Evaluation on Synthetic Datasets}

\begin{figure}[t]
\centering
\begin{minipage}[t]{.48\textwidth}
  \centering
  \includegraphics[width=1\linewidth]{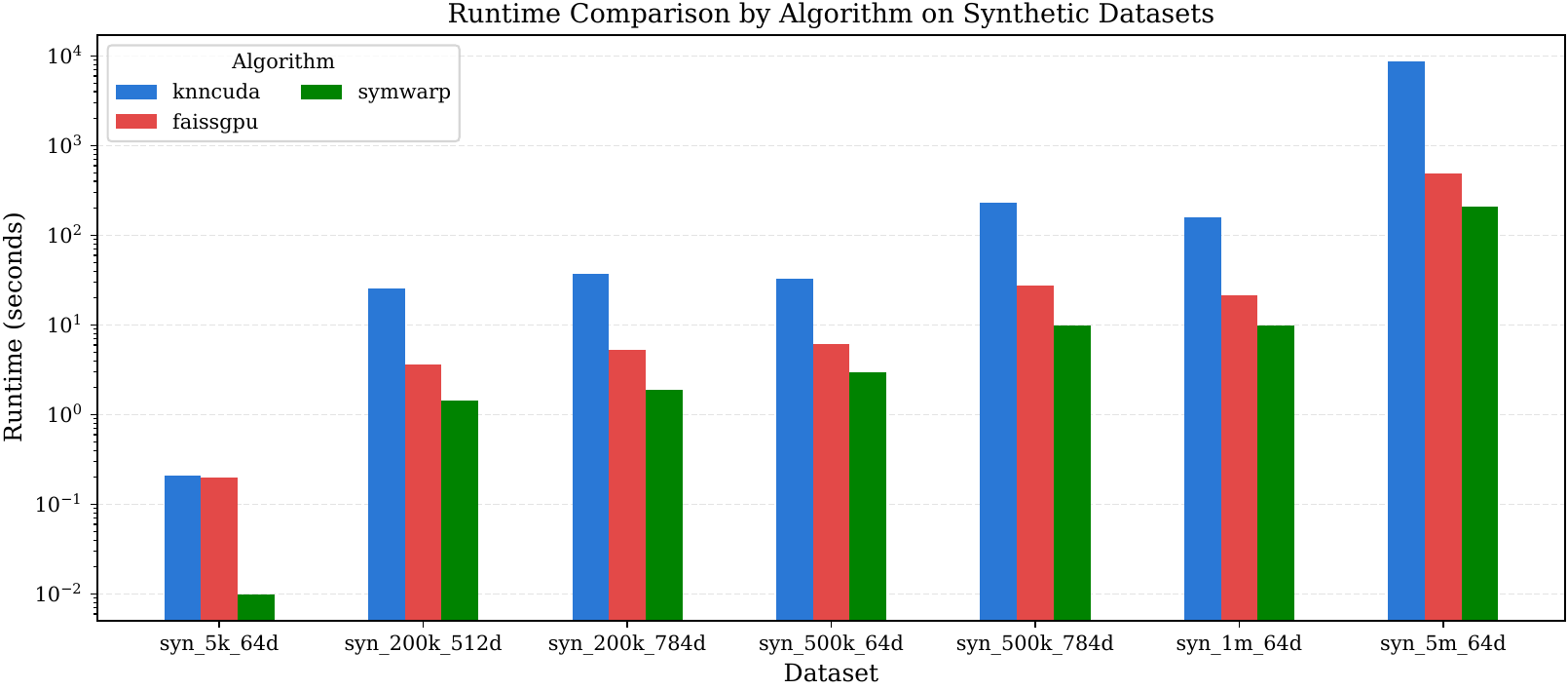}
  \caption{Runtime comparison of \textit{knncuda}, \textit{faissgpu}, and \textit{symwarp} across synthetic datasets ranging from 5K to 5M instances and spanning multiple dimensionalities ($c=64$, $k=25$). Runtime is shown on a logarithmic scale.}
  \label{fig:syn-runtime}
\end{minipage}%
\hfill
\begin{minipage}[t]{.48\textwidth}
  \centering
  \includegraphics[width=1\linewidth]{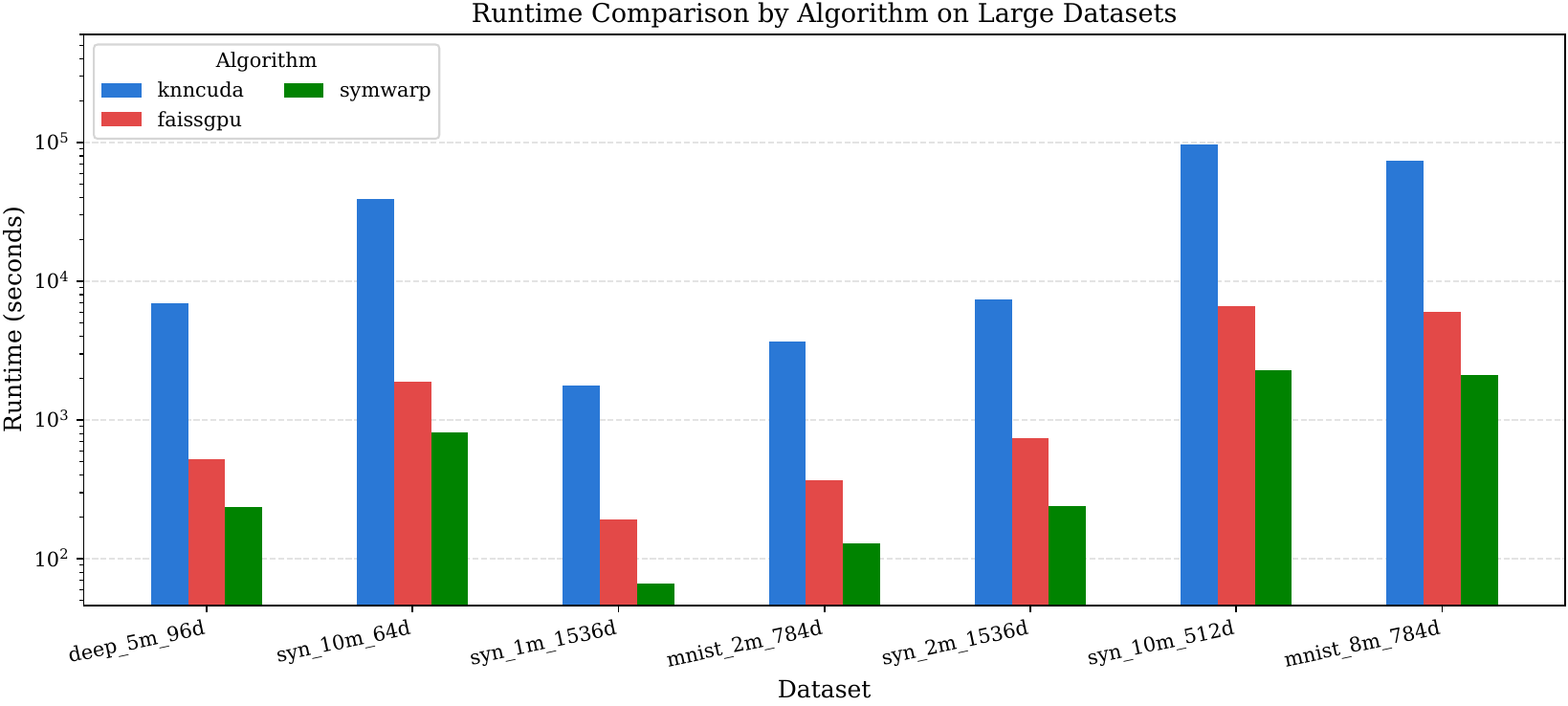}
  \caption{Runtime comparison of \textit{knncuda}, \textit{faissgpu}, and \textit{symwarp} across large datasets ranging from 1M to 10M instances and spanning multiple dimensionalities (MNIST: $c=10$, $k=25$; Deep1B: $c=80$, $k=50$; Synthetic: $c=25$, $k=32$). Runtime is reported on a logarithmic scale.}
  \label{fig:big-dataset-runtime-single}
\end{minipage}
\end{figure}

On seven synthetic datasets spanning 5K--5M points and 64--784 dimensions (Table~\ref{tab:syn-runtime-speedup}, Figure~\ref{fig:syn-runtime}), \textit{symwarp} achieves the lowest runtime on every dataset. On the smallest dataset (\texttt{syn\_5k\_64d}), all three methods complete in well under a second (0.01\,s for \textit{symwarp}, 0.20\,s for \textit{faissgpu}, and 0.21\,s for \textit{knncuda}); at this scale, runtimes are likely dominated by fixed overheads rather than by computation, so the resulting ratios ($21.00\times$ over \textit{knncuda} and $20\times$ over \textit{faissgpu}) are not representative of relative computational efficiency and are excluded from the summary ranges reported in this paper. On the remaining datasets, \textit{symwarp}'s speedup over \textit{knncuda} ranges from $11.16\times$ on \texttt{syn\_500k\_64d} to $41.25\times$ on the largest dataset evaluated, \texttt{syn\_5m\_64d}. Across the three 64-dimensional datasets, this speedup grows with dataset size ($11.16\times$, $15.97\times$, and $41.25\times$ at 500k, 1m, and 5m instances), suggesting that the advantage of \textit{symwarp} over \textit{knncuda} reflects favorable scaling behavior rather than merely a constant-factor improvement. The margin over \textit{faissgpu} varies considerably less, from $2.08\times$ to $2.80\times$. Across the three 64-dimensional datasets (500k--5m instances), it increases only slightly, from $2.08\times$ to $2.36\times$, despite a tenfold increase in dataset size. In comparison, at a fixed 200k instances, increasing dimensionality from 512d to 784d widens the margin from $2.52\times$ to $2.80\times$, and at a fixed 500k instances, increasing dimensionality from 64d to 784d widens it from $2.08\times$ to $2.77\times$. This pattern suggests that dimensionality has a somewhat stronger influence than instance count on \textit{symwarp}'s relative advantage over \textit{faissgpu}, although the differences within the evaluated range are modest.

\begin{table*}[t]
    \centering
    \caption{Runtime ($T$, in seconds) comparison on synthetic datasets. Speedup (SUP) is computed relative to the \textit{knncuda} baseline. The fastest runtime for each dataset is shown in bold.}
    \label{tab:syn-runtime-speedup}
    \setlength{\tabcolsep}{2.2pt}
    \renewcommand{\arraystretch}{0.88}
    \footnotesize
    \resizebox{\textwidth}{!}{%
        \begin{tabular}{l rr rr rr rr rr rr rr}
            \toprule
            \multirow{2}{*}{Algorithm} & \multicolumn{2}{c}{syn\_5k\_64d} & \multicolumn{2}{c}{syn\_200k\_512d} & \multicolumn{2}{c}{syn\_200k\_784d} & \multicolumn{2}{c}{syn\_500k\_64d} & \multicolumn{2}{c}{syn\_500k\_784d} & \multicolumn{2}{c}{syn\_1m\_64d} & \multicolumn{2}{c}{syn\_5m\_64d} \\
            \cmidrule(lr){2-3} \cmidrule(lr){4-5} \cmidrule(lr){6-7} \cmidrule(lr){8-9} \cmidrule(lr){10-11} \cmidrule(lr){12-13} \cmidrule(lr){14-15}
                & T & SUP & T & SUP & T & SUP & T & SUP & T & SUP & T & SUP & T & SUP \\
            \midrule
            \textit{knncuda}  & 0.21 & $1.00\times$ & 25.37 & $1.00\times$ & 37.38 & $1.00\times$ & 32.92 & $1.00\times$ & 232.48 & $1.00\times$ & 157.51 & $1.00\times$ & 8632.29 & $1.00\times$ \\
            \textit{faissgpu} & 0.20 & $1.05\times$ & 3.63 & $6.99\times$ & 5.32 & $7.03\times$ & 6.15 & $5.35\times$ & 27.62 & $8.42\times$ & 21.48 & $7.33\times$ & 494.69 & $17.45\times$ \\
            \textit{symwarp}  & \textbf{0.01} & \textbf{$21.00\times$} & \textbf{1.44} & \textbf{$17.62\times$} & \textbf{1.90} & \textbf{$19.67\times$} & \textbf{2.95} & \textbf{$11.16\times$} & \textbf{9.96} & \textbf{$23.34\times$} & \textbf{9.86} & \textbf{$15.97\times$} & \textbf{209.29} & \textbf{$41.25\times$} \\
            \bottomrule
        \end{tabular}%
    }
\end{table*}

\subsubsection{Performance on Large-Scale Datasets}

We evaluate seven large-scale datasets (Table~\ref{tab:dataset-stats}), including three derived from real-world data (\texttt{deep\_5m\_96d}, \texttt{mnist\_2m\_784d}, and \texttt{mnist\_8m\_784d}) and four synthetic datasets (\texttt{syn\_1m\_1536d}, \texttt{syn\_2m\_1536d}, \texttt{syn\_10m\_64d}, and \texttt{syn\_10m\_512d}). These datasets span 1M--10M instances, 64--1536 dimensions, and storage sizes from 1.92\,GB to 25.09\,GB.

As shown in Table~\ref{tab:big-runtime-speedup} and Figure~\ref{fig:big-dataset-runtime-single}, \textit{symwarp} achieves the lowest runtime on all seven datasets. Relative to \textit{knncuda}, its speedup ranges from $26.65\times$ on \texttt{syn\_1m\_1536d} to $48.21\times$ on \texttt{syn\_10m\_64d}. Although \textit{faissgpu} consistently outperforms \textit{knncuda}, it remains slower than \textit{symwarp} on every dataset, with the performance gap varying from approximately $2.2\times$ (\texttt{deep\_5m\_96d}) to $3.1\times$ (\texttt{syn\_2m\_1536d}).

The pair of 10M-instance synthetic datasets provides a controlled comparison of dimensionality at the largest scale evaluated. While \texttt{syn\_10m\_64d} and \texttt{syn\_10m\_512d} contain the same number of instances, increasing dimensionality from 64d to 512d widens \textit{symwarp}'s advantage over \textit{faissgpu} from $2.32\times$ to $2.89\times$. This trend is consistent with the behavior observed on the smaller synthetic datasets (Table~\ref{tab:syn-runtime-speedup}), where the gap between the two methods remains between $2.08\times$ and $2.36\times$ across the 64-dimensional datasets from 500K to 5M instances. The gap on \texttt{syn\_10m\_64d} ($2.32\times$) falls within this range, even though the large-scale synthetic runs use different parameters ($c=25$, $k=32$ rather than $c=64$, $k=25$). Together, these results suggest that dimensionality has a somewhat stronger influence than instance count on \textit{symwarp}'s relative advantage over \textit{faissgpu} within the evaluated range.

\begin{table}[t]
\centering
\captionsetup{justification=centering}
\caption{Dataset statistics and storage size.}
\label{tab:dataset-stats}
\footnotesize
\begin{tabular}{lrrr}
\toprule
Dataset & Instances & Dimensions & Size (GB) \\
\midrule
deep\_5m\_96d & 5m & 96 & 1.92 \\
syn\_10m\_64d & 10m & 64 & 2.56 \\
syn\_1m\_1536d & 1m & 1536 & 6.14 \\
mnist\_2m\_784d & 2m & 784 & 6.27 \\
syn\_2m\_1536d & 2m & 1536 & 12.29 \\
syn\_10m\_512d & 10m & 512 & 20.48 \\
mnist\_8m\_784d & 8m & 784 & 25.09 \\
\bottomrule
\end{tabular}
\end{table}

\begin{table*}[t]
    \centering
    \caption{Runtime ($T$, in seconds) comparison on large-scale datasets. Speedup (SUP) is computed relative to the \textit{knncuda} baseline. The fastest runtime for each dataset is shown in bold.}
    \label{tab:big-runtime-speedup}
    \setlength{\tabcolsep}{2.2pt}
    \renewcommand{\arraystretch}{0.88}
    \footnotesize
    \resizebox{\textwidth}{!}{%
        \begin{tabular}{l rr rr rr rr rr rr rr}
            \toprule
            \multirow{2}{*}{Algorithm} & \multicolumn{2}{c}{deep\_5m\_96d} & \multicolumn{2}{c}{syn\_10m\_64d} & \multicolumn{2}{c}{syn\_1m\_1536d} & \multicolumn{2}{c}{mnist\_2m\_784d} & \multicolumn{2}{c}{syn\_2m\_1536d} & \multicolumn{2}{c}{syn\_10m\_512d} & \multicolumn{2}{c}{mnist\_8m\_784d} \\
            \cmidrule(lr){2-3} \cmidrule(lr){4-5} \cmidrule(lr){6-7} \cmidrule(lr){8-9} \cmidrule(lr){10-11} \cmidrule(lr){12-13} \cmidrule(lr){14-15}
                & T & SUP & T & SUP & T & SUP & T & SUP & T & SUP & T & SUP & T & SUP \\
            \midrule
            \textit{knncuda}  & 6925.58 & $1.00\times$ & 39260.30 & $1.00\times$ & 1776.60 & $1.00\times$ & 3676.36 & $1.00\times$ & 7355.30 & $1.00\times$ & 97236.40 & $1.00\times$ & 74407.90 & $1.00\times$ \\
            \textit{faissgpu} & 525.83 & $13.17\times$ & 1886.65 & $20.81\times$ & 191.62 & $9.27\times$ & 368.77 & $9.97\times$ & 745.33 & $9.87\times$ & 6594.34 & $14.75\times$ & 5985.10 & $12.43\times$ \\
            \textit{symwarp}  & \textbf{236.77} & $29.25\times$ & \textbf{814.36} & $48.21\times$ & \textbf{66.66} & $26.65\times$ & \textbf{128.79} & $28.55\times$ & \textbf{240.78} & $30.55\times$ & \textbf{2278.88} & $42.67\times$ & \textbf{2121.55} & $35.07\times$ \\
            \bottomrule
        \end{tabular}%
    }
    \vspace{-2.0mm}
\end{table*}

\subsubsection{Evaluation on the Impact of Dimensionality}

\begin{figure}[t]
\centering
\begin{minipage}[t]{.48\textwidth}
  \centering
  \includegraphics[width=1\linewidth]{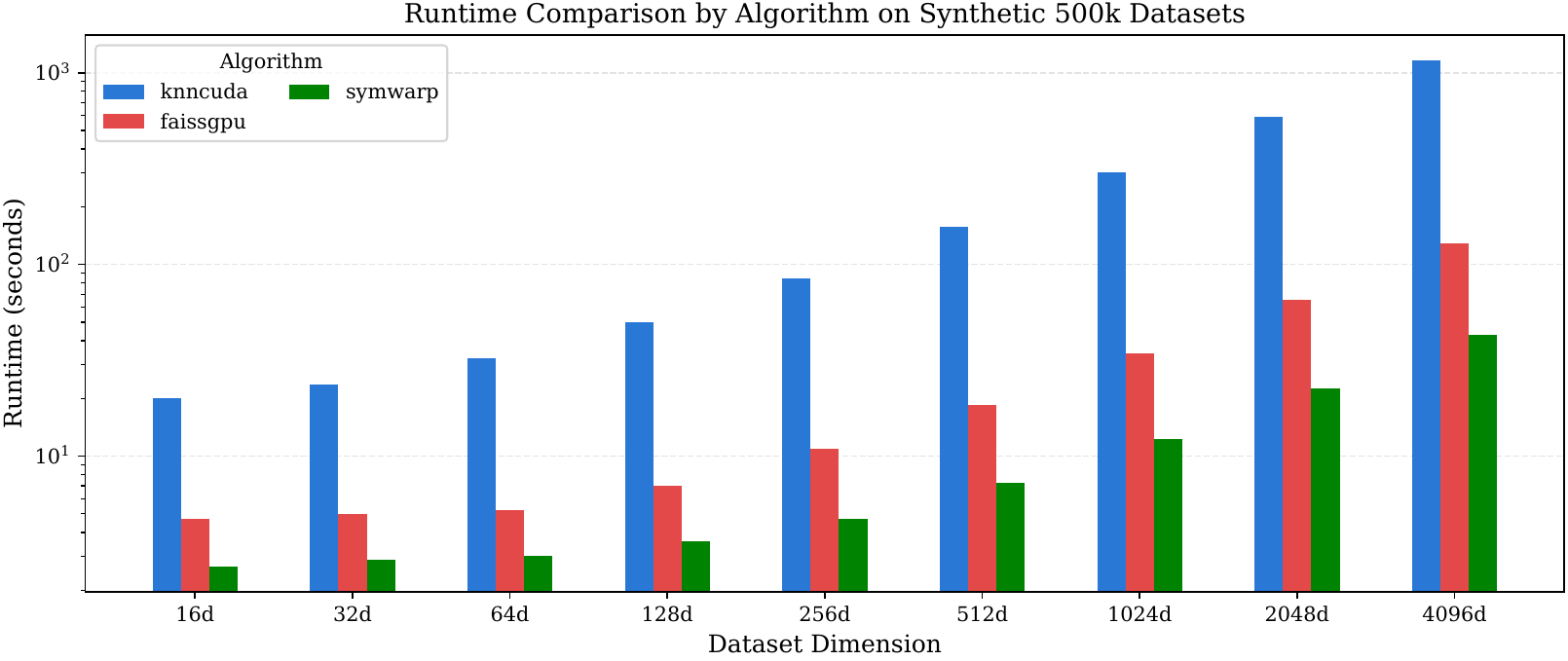}
  \caption{Runtime comparison of \textit{knncuda}, \textit{faissgpu}, and \textit{symwarp} on synthetic datasets with 500K instances and increasing dimensionality ($c=20$, $k=32$). Runtime is reported on a logarithmic scale.}
  \label{fig:syn-500k-runtime-bar}
\end{minipage}%
\hfill
\begin{minipage}[t]{.48\textwidth}
  \centering
  \includegraphics[width=1\linewidth]{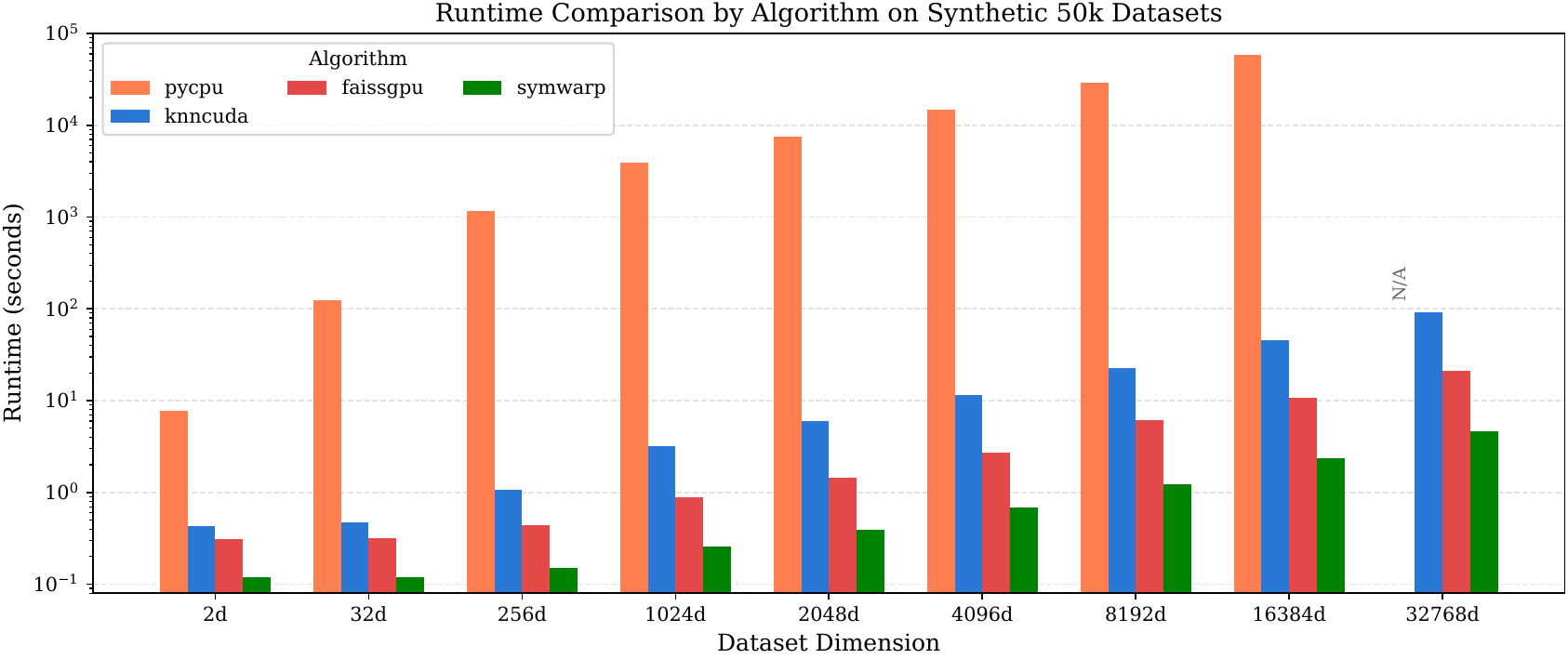}
  \caption{Runtime comparison of \textit{pycpu}, \textit{knncuda}, \textit{faissgpu}, and \textit{symwarp} on synthetic datasets with 50K instances and increasing dimensionality ($c=25$, $k=32$). Runtime is reported on a logarithmic scale.}
  \label{fig:syn50k-runtime-bar}
\end{minipage}
\end{figure}

On synthetic datasets with a fixed 500,000 instances but dimensionality ranging from 16d to 4096d (Table~\ref{tab:synthetic500k-runtime-speedup}, Figure~\ref{fig:syn-500k-runtime-bar}), \textit{symwarp} consistently achieves the lowest runtime. As dimensionality increases, its runtime grows from 2.65\,s to 42.75\,s ($16\times$), compared with increases from 20.03\,s to 1159.20\,s ($58\times$) for \textit{knncuda} and from 4.72\,s to 129.40\,s ($27\times$) for \textit{faissgpu}. Consequently, \textit{symwarp}'s speedup over \textit{knncuda} increases from $7.56\times$ at 16d to $27.12\times$ at 4096d, although the increase slows at the highest dimensionalities ($24.42\times$ at 1024d). In addition, its advantage over \textit{faissgpu} widens from $1.78\times$ to $3.03\times$; the margin remains nearly constant ($1.73$--$1.78\times$) from 16d to 64d, and almost all of the widening occurs at 128d and above. These results demonstrate more favorable scaling of \textit{symwarp} with feature dimensionality at fixed dataset cardinality, particularly once dimensionality exceeds roughly 100.

\begin{table*}[t]
    \centering
    \caption{Runtime ($T$, in seconds) comparison on synthetic datasets with 500k instances. Speedup (SUP) is computed relative to the \textit{knncuda} baseline. The fastest runtime for each dimensionality is shown in bold.}
    \label{tab:synthetic500k-runtime-speedup}
    \setlength{\tabcolsep}{2.1pt}
    \renewcommand{\arraystretch}{0.90}
    \footnotesize
    \resizebox{\textwidth}{!}{%
        \begin{tabular}{l rr rr rr rr rr rr rr rr rr}
            \toprule
            \multirow{2}{*}{Algorithm} & \multicolumn{2}{c}{16d} & \multicolumn{2}{c}{32d} & \multicolumn{2}{c}{64d} & \multicolumn{2}{c}{128d} & \multicolumn{2}{c}{256d} & \multicolumn{2}{c}{512d} & \multicolumn{2}{c}{1024d} & \multicolumn{2}{c}{2048d} & \multicolumn{2}{c}{4096d} \\
            \cmidrule(lr){2-3} \cmidrule(lr){4-5} \cmidrule(lr){6-7} \cmidrule(lr){8-9} \cmidrule(lr){10-11} \cmidrule(lr){12-13} \cmidrule(lr){14-15} \cmidrule(lr){16-17} \cmidrule(lr){18-19}
                & T & SUP & T & SUP & T & SUP & T & SUP & T & SUP & T & SUP & T & SUP & T & SUP & T & SUP \\
            \midrule
            \textit{knncuda}  & 20.03 & $1.00\times$ & 23.71 & $1.00\times$ & 32.46 & $1.00\times$ & 49.87 & $1.00\times$ & 84.85 & $1.00\times$ & 157.00 & $1.00\times$ & 300.80 & $1.00\times$ & 587.10 & $1.00\times$ & 1159.20 & $1.00\times$ \\
            \textit{faissgpu} & 4.72 & $4.24\times$ & 5.01 & $4.73\times$ & 5.25 & $6.18\times$ & 7.04 & $7.08\times$ & 10.97 & $7.73\times$ & 18.54 & $8.47\times$ & 34.34 & $8.76\times$ & 65.54 & $8.96\times$ & 129.40 & $8.96\times$ \\
            \textit{symwarp}  & \textbf{2.65} & \textbf{$7.56\times$} & \textbf{2.89} & \textbf{$8.20\times$} & \textbf{3.01} & \textbf{$10.78\times$} & \textbf{3.60} & \textbf{$13.85\times$} & \textbf{4.70} & \textbf{$18.05\times$} & \textbf{7.24} & \textbf{$21.69\times$} & \textbf{12.32} & \textbf{$24.42\times$} & \textbf{22.48} & \textbf{$26.12\times$} & \textbf{42.75} & \textbf{$27.12\times$} \\
            \bottomrule
        \end{tabular}%
    }
    \vspace{-1.2mm}
\end{table*}

\subsubsection{Comparison with the Original FastDEC Implementation}
\label{sec:pycpu-fastdec-comparison}

To facilitate comparison with the original CPU-based \textit{pycpu} implementation of FastDEC, we evaluate several comparatively small datasets derived from synthetic and real-world data.

Table~\ref{tab:syn50k-runtime-speedup} and Figure~\ref{fig:syn50k-runtime-bar} show that \textit{SymWarp-FastDEC} consistently achieves the lowest runtime on synthetic datasets containing 50,000 instances and dimensionalities ranging from 2d to 32,768d. The runtime of \textit{pycpu} increases substantially, from 7.69\,s at 2d to 58,314.25\,s at 16,384d, and it fails to complete at 32,768d. In contrast, \textit{SymWarp-FastDEC} increases only from 0.12\,s to 2.38\,s across the same range, reaching 4.65\,s at 32,768d. Consequently, its speedup over \textit{pycpu} grows from $64.08\times$ to $24{,}501.79\times$, while achieving a speedup of $19.71\times$ over \textit{knncuda} at 32,768d where \textit{pycpu} is unavailable.

For assessing the relative efficiency of GPU implementations, a more relevant comparison is against the GPU baselines. \textit{SymWarp-FastDEC}'s margin over \textit{faissgpu} widens from $2.58\times$ at 2d to a peak of $4.90\times$ at 8,192d, and remains at approximately $4.5\times$ at 16,384d and 32,768d ($4.53\times$ and $4.58\times$, respectively). This peak is the largest gap observed between the two methods across all experiments in this study, excluding the sub-second \texttt{syn\_5k\_64d} case discussed above. These results demonstrate favorable scaling of \textit{SymWarp-FastDEC} with increasing dimensionality, although its margin over \textit{faissgpu} plateaus beyond approximately 8,192d.

\begin{table*}[t]
    \centering
    \caption{Runtime ($T$, in seconds) comparison on synthetic datasets with 50k instances. Speedup (SUP) is computed relative to the \textit{pycpu} baseline. At 32768d, speedup is computed relative to \textit{knncuda} because \textit{pycpu} did not complete. The fastest runtime for each dataset is shown in bold.}
    \label{tab:syn50k-runtime-speedup}
    \setlength{\tabcolsep}{1.9pt}
    \renewcommand{\arraystretch}{0.88}
    \footnotesize
    \resizebox{\textwidth}{!}{%
        \begin{tabular}{l rr rr rr rr rr rr rr rr rr rr}
            \toprule
            \multirow{2}{*}{Algorithm} & \multicolumn{2}{c}{2d} & \multicolumn{2}{c}{32d} & \multicolumn{2}{c}{256d} & \multicolumn{2}{c}{1024d} & \multicolumn{2}{c}{2048d} & \multicolumn{2}{c}{4096d} & \multicolumn{2}{c}{8192d} & \multicolumn{2}{c}{16384d} & \multicolumn{2}{c}{32768d} \\
            \cmidrule(lr){2-3} \cmidrule(lr){4-5} \cmidrule(lr){6-7} \cmidrule(lr){8-9} \cmidrule(lr){10-11} \cmidrule(lr){12-13} \cmidrule(lr){14-15} \cmidrule(lr){16-17} \cmidrule(lr){18-19}
                & T & SUP & T & SUP & T & SUP & T & SUP & T & SUP & T & SUP & T & SUP & T & SUP & T & SUP \\
            \midrule
            \textit{pycpu}    & 7.69 & $1.00\times$ & 123.66 & $1.00\times$ & 1174.51 & $1.00\times$ & 3933.25 & $1.00\times$ & 7562.34 & $1.00\times$ & 14834.52 & $1.00\times$ & 29426.31 & $1.00\times$ & 58314.25 & $1.00\times$ & -- & -- \\
            \textit{knncuda} & 0.43 & $17.88\times$ & 0.47 & $263.11\times$ & 1.08 & $1087.51\times$ & 3.20 & $1229.14\times$ & 5.98 & $1264.61\times$ & 11.55 & $1284.37\times$ & 22.69 & $1296.88\times$ & 45.28 & $1287.86\times$ & 91.64 & $1.00\times$ \\
            \textit{faissgpu} & 0.31 & $24.81\times$ & 0.32 & $386.44\times$ & 0.44 & $2669.34\times$ & 0.88 & $4469.60\times$ & 1.44 & $5251.63\times$ & 2.73 & $5433.89\times$ & 6.08 & $4839.85\times$ & 10.77 & $5414.51\times$ & 21.29 & $4.30\times$ \\
            \textit{symwarp}  & \textbf{0.12} & \textbf{$64.08\times$} & \textbf{0.12} & \textbf{$1030.50\times$} & \textbf{0.15} & \textbf{$7830.07\times$} & \textbf{0.26} & \textbf{$15127.88\times$} & \textbf{0.39} & \textbf{$19390.62\times$} & \textbf{0.69} & \textbf{$21499.30\times$} & \textbf{1.24} & \textbf{$23730.90\times$} & \textbf{2.38} & \textbf{$24501.79\times$} & \textbf{4.65} & \textbf{$19.71\times$} \\
            \bottomrule
        \end{tabular}%
    }
    \vspace{-1.2mm}
\end{table*}

\begin{figure}[t]
\centering
\begin{minipage}[t]{.48\textwidth}
  \centering
  \includegraphics[width=1\linewidth]{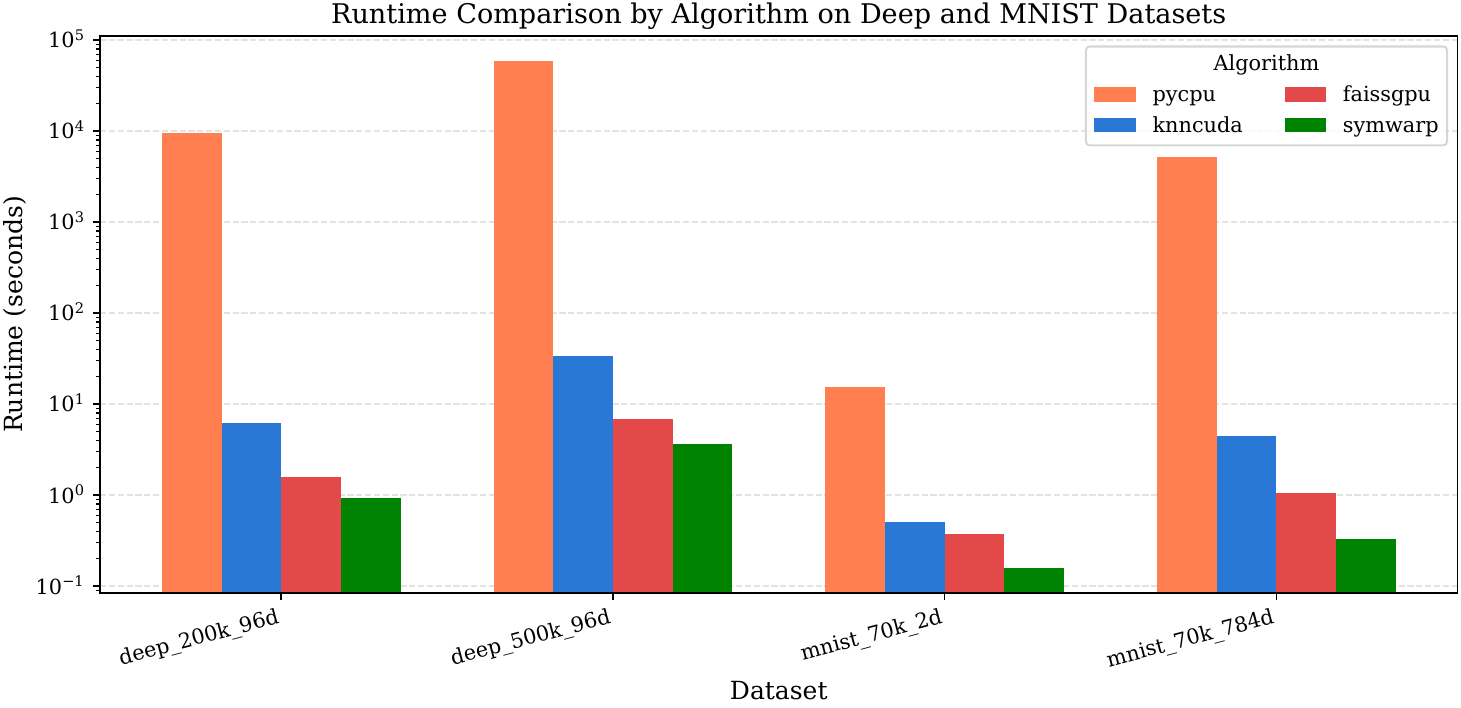}
  \caption{Runtime comparison of \textit{pycpu}, \textit{knncuda}, \textit{faissgpu}, and \textit{symwarp} on Deep1B and MNIST datasets spanning a range of dataset sizes and dimensionalities (Deep1B: $c=80$, $k=50$; MNIST: $c=10$, $k=25$). Runtime is reported on a logarithmic scale.}
  \label{fig:small-deep-mnist-runtime}
\end{minipage}%
\hfill
\begin{minipage}[t]{.48\textwidth}
  \centering
  \includegraphics[width=1\linewidth]{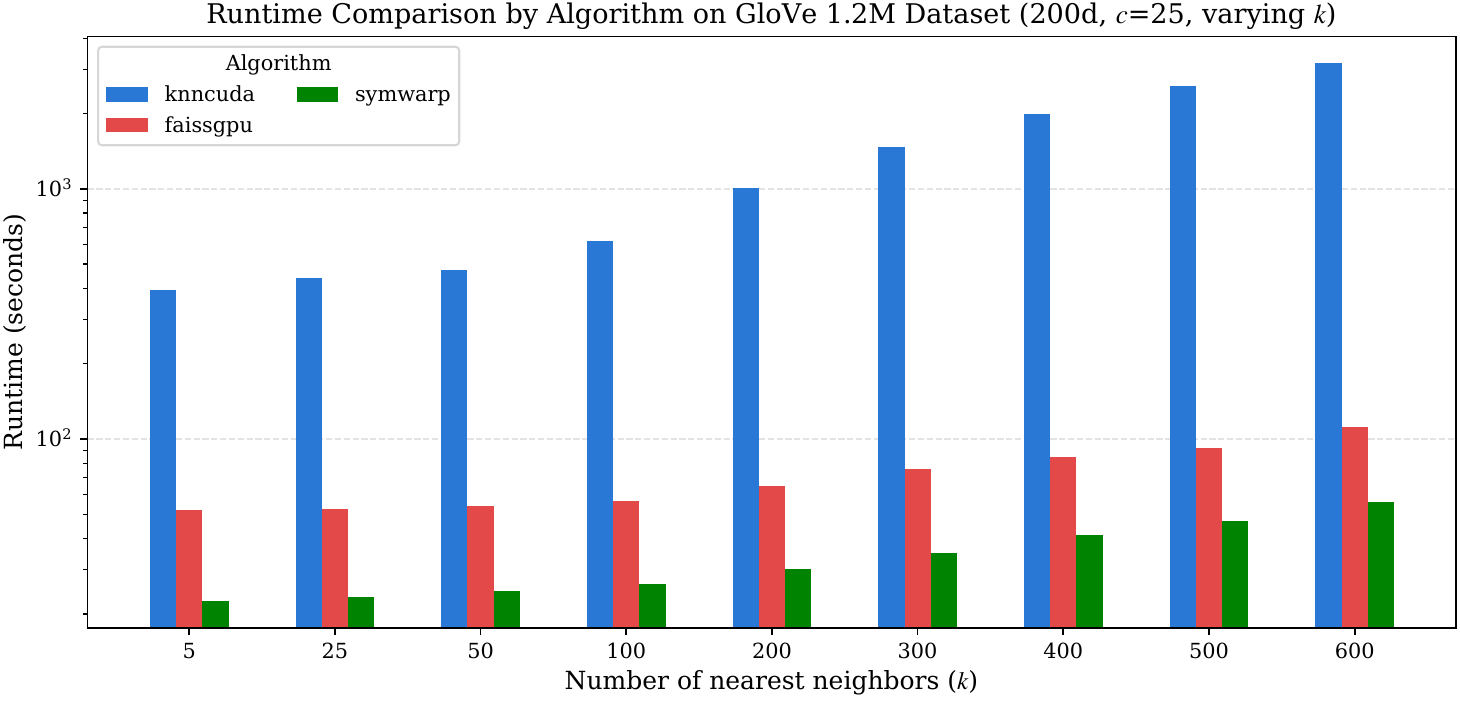}
  \caption{Impact of the number of nearest neighbors $k$ on the runtime of \textit{knncuda}, \textit{faissgpu}, and \textit{symwarp} on the GloVe dataset (1.2M instances, 200 dimensions) with the number of clusters fixed at $c=25$. Runtime is reported on a logarithmic scale.}
  \label{fig:glove-various-k-runtime}
\end{minipage}
\end{figure}

As summarized in Table~\ref{tab:deep-mnist-runtime-speedup} and illustrated in Figure~\ref{fig:small-deep-mnist-runtime}, \textit{SymWarp-FastDEC} achieves the lowest runtime across four heterogeneous benchmarks, including two Deep1B subsets and two MNIST datasets spanning different scales and dimensionalities, with speedups over \textit{pycpu} ranging from $96.94\times$ to $15{,}931.13\times$. For assessing the relative efficiency of GPU implementations, a more relevant comparison is against the GPU FastDEC variants. \textit{SymWarp-FastDEC} outperforms both \textit{knncuda} and \textit{faissgpu} on all four datasets, with its margin over \textit{faissgpu} ranging from $1.70\times$ on deep\_200k\_96d to $3.18\times$ on mnist\_70k\_784d. These results provide further evidence of a practical and consistent efficiency advantage over alternative GPU realizations across varied scales and dimensionalities.

\begin{table*}[t]
    \centering
    \caption{Runtime ($T$, in seconds) comparison on Deep1B and MNIST datasets. Speedup (SUP) is computed relative to the \textit{pycpu} baseline. The fastest runtime for each dataset is shown in bold.}
    \label{tab:deep-mnist-runtime-speedup}
    \setlength{\tabcolsep}{2.6pt}
    \renewcommand{\arraystretch}{0.90}
    \footnotesize
    \resizebox{\textwidth}{!}{%
        \begin{tabular}{l rr rr rr rr}
            \toprule
            \multirow{2}{*}{Algorithm} & \multicolumn{2}{c}{deep\_200k\_96d} & \multicolumn{2}{c}{deep\_500k\_96d} & \multicolumn{2}{c}{mnist\_70k\_2d} & \multicolumn{2}{c}{mnist\_70k\_784d} \\
            \cmidrule(lr){2-3} \cmidrule(lr){4-5} \cmidrule(lr){6-7} \cmidrule(lr){8-9}
                & T & SUP & T & SUP & T & SUP & T & SUP \\
            \midrule
            \textit{pycpu}    & 9543.81 & $1.00\times$ & 58307.94 & $1.00\times$ & 15.51 & $1.00\times$ & 5125.48 & $1.00\times$ \\
            \textit{knncuda} & 6.25 & $1527.01\times$ & 33.94 & $1717.97\times$ & 0.51 & $30.41\times$ & 4.50 & $1139.00\times$ \\
            \textit{faissgpu} & 1.60 & $5964.88\times$ & 6.87 & $8487.33\times$ & 0.37 & $41.92\times$ & 1.05 & $4881.41\times$ \\
            \textit{symwarp}  & \textbf{0.94} & \textbf{$10152.99\times$} & \textbf{3.66} & \textbf{$15931.13\times$} & \textbf{0.16} & \textbf{$96.94\times$} & \textbf{0.33} & \textbf{$15531.76\times$} \\
            \bottomrule
        \end{tabular}%
    }
    \vspace{-1.2mm}
\end{table*}

\subsubsection{Evaluation on the Impact of the Number of Nearest Neighbors}

\begin{figure}[t]
\centering
\begin{minipage}[t]{.48\textwidth}
  \centering
  \includegraphics[width=1\linewidth]{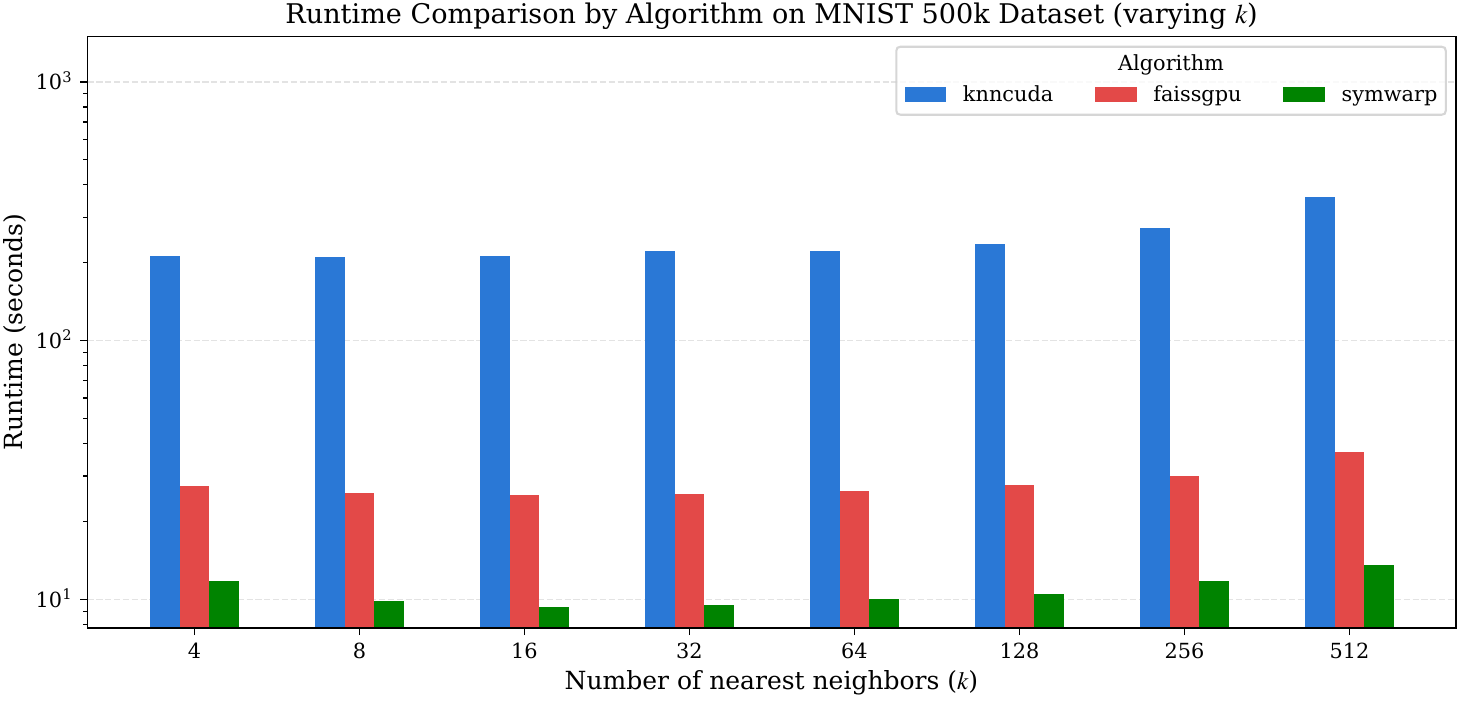}
  \caption{Impact of the number of nearest neighbors $k$ on the runtime of \textit{knncuda}, \textit{faissgpu}, and \textit{symwarp} on the MNIST dataset (500K instances, 784 dimensions) with the number of clusters fixed at $c=10$. Runtime is reported on a logarithmic scale.}
  \label{fig:mnist-500k-k-runtime}
\end{minipage}%
\hfill
\begin{minipage}[t]{.48\textwidth}
  \centering
  \includegraphics[width=1\linewidth]{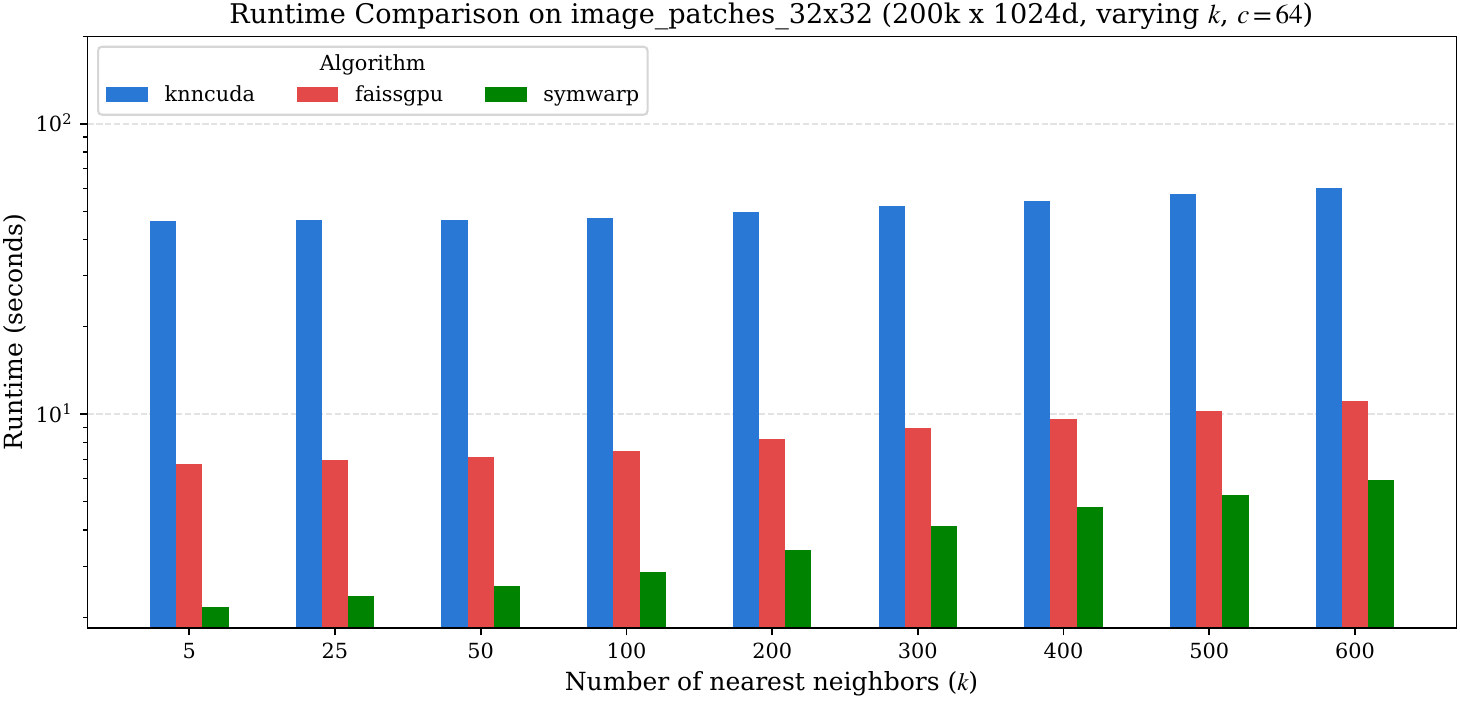}
  \caption{Impact of the number of nearest neighbors $k$ on the runtime of \textit{knncuda}, \textit{faissgpu}, and \textit{symwarp} on the 32$\times$32 image-patch dataset (200K instances, 1024 dimensions) with the number of clusters fixed at $c=64$. Runtime is reported on a logarithmic scale.}
  \label{fig:patch32x32-various-k-runtime}
\end{minipage}
\end{figure}

\begin{figure}[t]
\centering
\begin{minipage}[t]{.48\textwidth}
  \centering
  \includegraphics[width=1\linewidth]{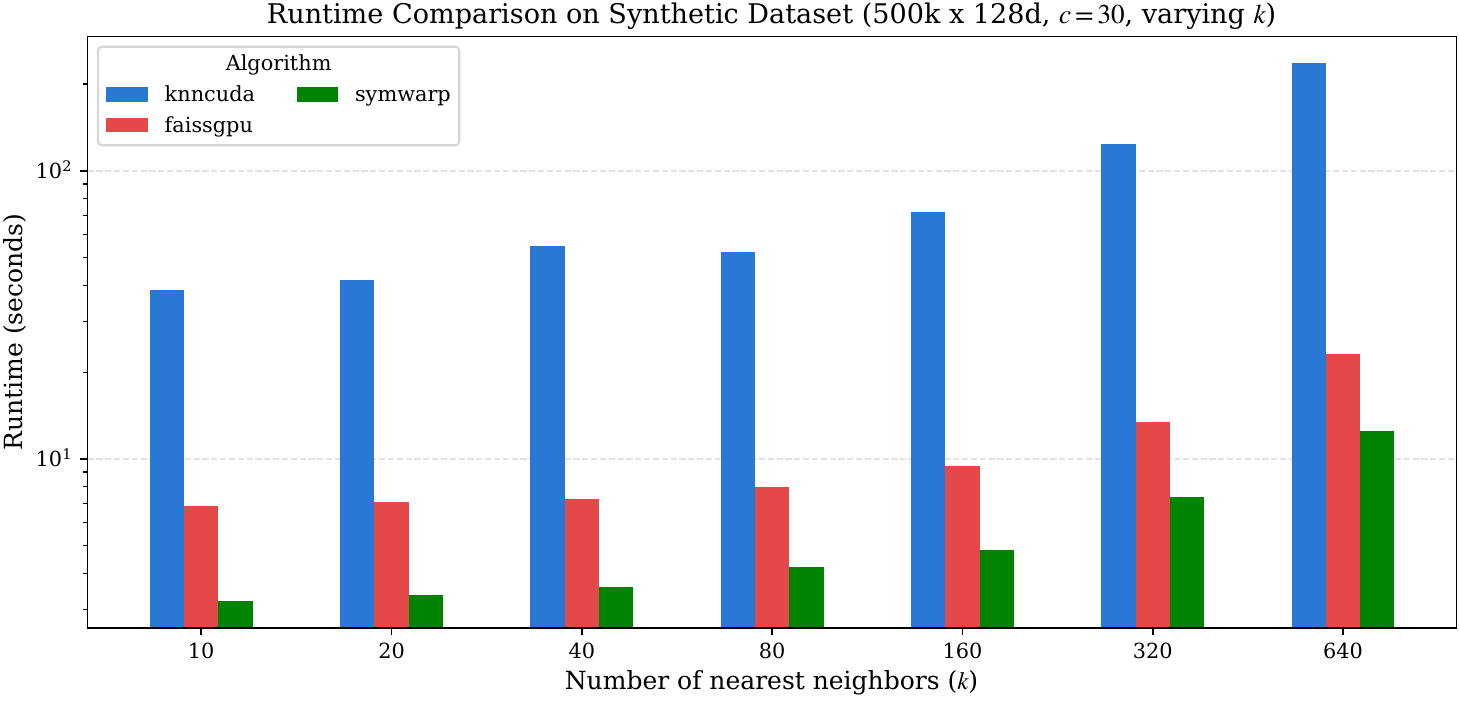}
  \caption{Impact of the number of nearest neighbors $k$ on the runtime of \textit{knncuda}, \textit{faissgpu}, and \textit{symwarp} on the synthetic dataset (500K instances, 128 dimensions) with the number of clusters fixed at $c=30$. Runtime is reported on a logarithmic scale.}
  \label{fig:syn-500k-128d-various-k-runtime}
\end{minipage}%
\hfill
\begin{minipage}[t]{.48\textwidth}
  \centering
  \includegraphics[width=1\linewidth]{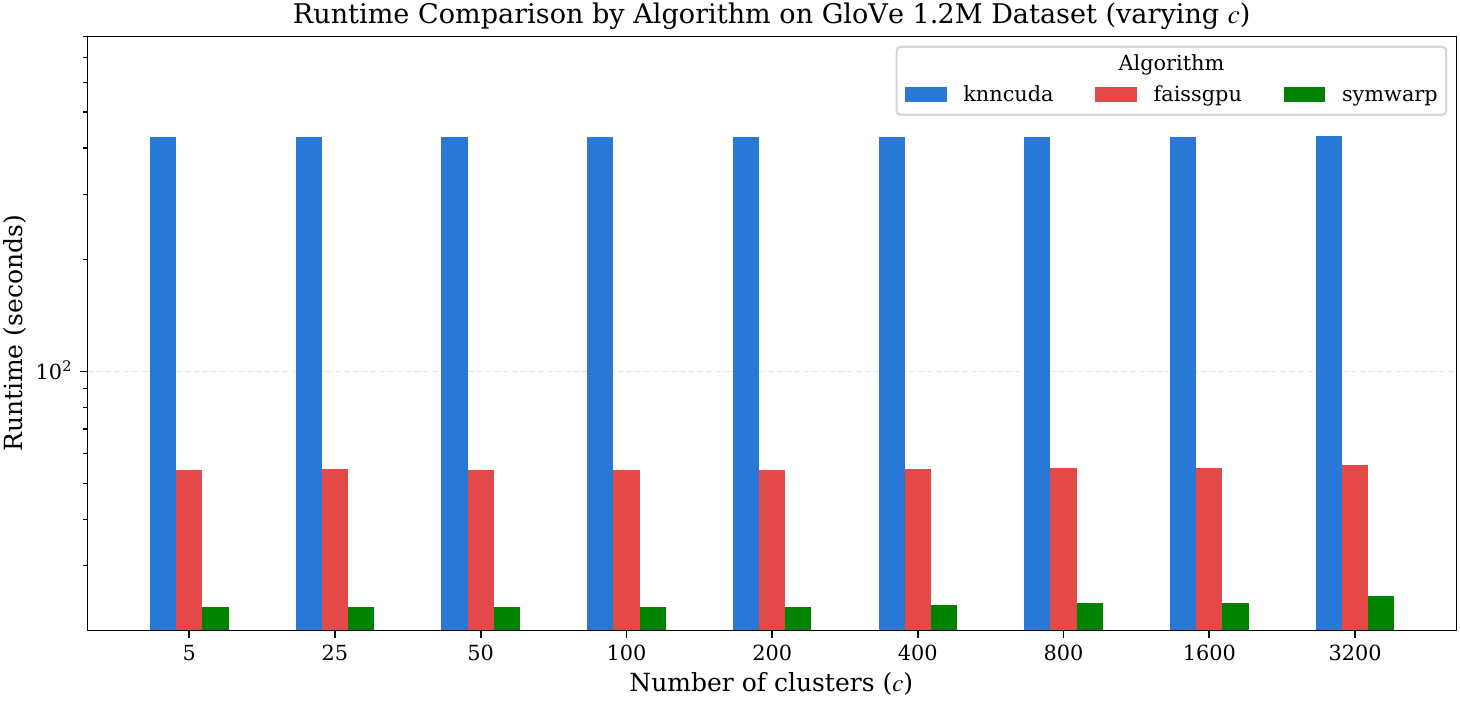}
  \caption{Impact of the number of clusters $c$ on the runtime of \textit{knncuda}, \textit{faissgpu}, and \textit{symwarp} on the GloVe dataset (1.2M instances, 200 dimensions) with the number of nearest neighbors fixed at $k=20$. Runtime is reported on a logarithmic scale.}
  \label{fig:glove-various-c-runtime}
\end{minipage}
\end{figure}

We used the same four benchmark datasets: GloVe, MNIST, image-patch, and synthetic (Tables~\ref{tab:glove-various-k-speedup}--\ref{tab:syn-500k-128d-c30-runtime-speedup}, Figures~\ref{fig:glove-various-k-runtime}--\ref{fig:syn-500k-128d-various-k-runtime}). On each dataset we fixed the number of clusters $c$ and varied the number of nearest neighbors $k$ from single digits up to $512$--$640$. Unlike $c$, the value of $k$ directly sets the size of the $k$-NN graph and the cost of top-$k$ selection, so runtime generally increases with $k$, particularly at larger neighborhood sizes. Even so, \textit{symwarp} is the fastest method for every value of $k$ on every dataset. Its speedup over \textit{knncuda} ranges from $10.1\times$ to $56.6\times$, and it is $1.8$--$3.1\times$ faster than \textit{faissgpu}. Because \textit{faissgpu} is the strongest baseline, the margin over it is the more informative measure of how efficiently \textit{symwarp} handles larger neighborhoods. On three of the four datasets (GloVe, image-patch, and synthetic), this margin narrows moderately as $k$ increases, consistent with top-$k$ maintenance accounting for a growing share of \textit{symwarp}'s runtime at large $k$; on MNIST it remains essentially constant.

On GloVe ($c=25$), the speedup over \textit{knncuda} grows steadily with $k$, from $17.5\times$ at $k=5$ to $56.6\times$ at $k=600$. Over this range, the runtime of \textit{knncuda} increases about $8\times$, while that of \textit{symwarp} increases only about $2.5\times$. The margin over \textit{faissgpu} narrows gradually, from $2.3\times$ at $k=5$ to about $2.0\times$ for $k \ge 400$.

On MNIST ($c=10$), the runtime of \textit{symwarp} is lowest and nearly flat for $8 \le k \le 128$ (9.31--10.48\,s), and remains below $14$\,s even at $k=512$. It is $18$--$27\times$ faster than \textit{knncuda} and keeps a consistent margin of $2.3$--$2.7\times$ over \textit{faissgpu} across all values of $k$.

On the image-patch dataset ($c=64$), the trend runs the other way. The speedup over \textit{knncuda} falls from $21.3\times$ at $k=5$ to $10.1\times$ at $k=600$, and the margin over \textit{faissgpu} narrows from $3.1\times$ to $1.9\times$. The runtimes of both baselines rise only modestly with $k$ (about $1.3\times$ for \textit{knncuda} and $1.7\times$ for \textit{faissgpu}), which is consistent with distance computation accounting for most of their cost on this high-dimensional dataset (1024 dimensions, 200K instances). The runtime of \textit{symwarp} increases about $2.7\times$, but from a much smaller base, and stays below $6$\,s at $k=600$.

On the synthetic dataset ($c=30$), \textit{symwarp} and \textit{faissgpu} scale similarly with $k$, although the runtime of \textit{symwarp} grows somewhat faster: from $k=10$ to $k=640$, their runtimes increase by $3.88\times$ and $3.35\times$, respectively. As a result, the margin of \textit{symwarp} over \textit{faissgpu} narrows slightly, from $2.1\times$ at $k=10$ to $1.8$--$1.9\times$ for $k \ge 320$. The runtime of \textit{knncuda} grows only moderately up to $k=160$ but rises sharply at $k=320$ and $k=640$, so the speedup of \textit{symwarp} over it increases from $12.0\times$ to $19.0\times$.

Overall, the robustness study of \citet{yang2022fastdec} showed that FastDEC maintains clustering quality superior to the competing methods evaluated for $k$ ranging from $10$ to $200$. Smaller values of $k$ within this range are therefore preferable in practice, as they make SymWarp-FastDEC substantially faster.

\begin{table*}[t]
\centering
\caption{Runtime ($T$, in seconds) Comparison on the GloVe Dataset ($1.2\mathrm{M}$ instances, 200 dimensions) with Number of Clusters Fixed at $c=25$, Varying Number of Nearest Neighbors $k$. Speedup (SUP) is computed relative to the \textit{knncuda} baseline. The fastest runtime per column is shown in bold.}
\label{tab:glove-various-k-speedup}
\setlength{\tabcolsep}{2.6pt}
\renewcommand{\arraystretch}{0.90}
\footnotesize
\resizebox{\textwidth}{!}{%
\begin{tabular}{l rr rr rr rr rr rr rr rr rr}
\toprule
\multirow{2}{*}{Algorithm} & \multicolumn{2}{c}{$k=5$} & \multicolumn{2}{c}{$k=25$} & \multicolumn{2}{c}{$k=50$} & \multicolumn{2}{c}{$k=100$} & \multicolumn{2}{c}{$k=200$} & \multicolumn{2}{c}{$k=300$} & \multicolumn{2}{c}{$k=400$} & \multicolumn{2}{c}{$k=500$} & \multicolumn{2}{c}{$k=600$} \\
\cmidrule(lr){2-3} \cmidrule(lr){4-5} \cmidrule(lr){6-7} \cmidrule(lr){8-9} \cmidrule(lr){10-11} \cmidrule(lr){12-13} \cmidrule(lr){14-15} \cmidrule(lr){16-17} \cmidrule(lr){18-19}
& T & SUP & T & SUP & T & SUP & T & SUP & T & SUP & T & SUP & T & SUP & T & SUP & T & SUP \\
\midrule
knncuda & 394.25 & $1.00\times$ & 441.31 & $1.00\times$ & 472.63 & $1.00\times$ & 619.69 & $1.00\times$ & 1007.56 & $1.00\times$ & 1471.92 & $1.00\times$ & 1991.05 & $1.00\times$ & 2569.18 & $1.00\times$ & 3175.71 & $1.00\times$ \\
faissgpu & 52.15 & $7.56\times$ & 52.54 & $8.40\times$ & 53.92 & $8.77\times$ & 56.46 & $10.98\times$ & 65.12 & $15.47\times$ & 76.12 & $19.34\times$ & 84.65 & $23.52\times$ & 92.33 & $27.83\times$ & 111.77 & $28.41\times$ \\
symwarp & \textbf{22.52} & $17.51\times$ & \textbf{23.46} & $18.81\times$ & \textbf{24.76} & $19.09\times$ & \textbf{26.41} & $23.46\times$ & \textbf{30.34} & $33.21\times$ & \textbf{35.11} & $41.92\times$ & \textbf{41.48} & $48.00\times$ & \textbf{47.22} & $54.41\times$ & \textbf{56.08} & $56.63\times$ \\
\bottomrule
\end{tabular}%
}
\vspace{-1.2mm}
\end{table*}

\begin{table*}[t]
\centering
\caption{Runtime ($T$, in seconds) Comparison on the MNIST Dataset (500K instances, 784 dimensions) with Number of Clusters Fixed at $c=10$, Varying Number of Nearest Neighbors $k$. Speedup (SUP) is computed relative to the \textit{knncuda} baseline. The fastest runtime per column is shown in bold.}
\label{tab:mnist-500k-k-speedup}
\setlength{\tabcolsep}{2.6pt}
\renewcommand{\arraystretch}{0.90}
\footnotesize
\resizebox{\textwidth}{!}{%
\begin{tabular}{l rr rr rr rr rr rr rr rr}
\toprule
\multirow{2}{*}{Algorithm} & \multicolumn{2}{c}{$k=4$} & \multicolumn{2}{c}{$k=8$} & \multicolumn{2}{c}{$k=16$} & \multicolumn{2}{c}{$k=32$} & \multicolumn{2}{c}{$k=64$} & \multicolumn{2}{c}{$k=128$} & \multicolumn{2}{c}{$k=256$} & \multicolumn{2}{c}{$k=512$} \\
\cmidrule(lr){2-3} \cmidrule(lr){4-5} \cmidrule(lr){6-7} \cmidrule(lr){8-9} \cmidrule(lr){10-11} \cmidrule(lr){12-13} \cmidrule(lr){14-15} \cmidrule(lr){16-17}
& T & SUP & T & SUP & T & SUP & T & SUP & T & SUP & T & SUP & T & SUP & T & SUP \\
\midrule
knncuda & 212.62 & $1.00\times$ & 211.32 & $1.00\times$ & 213.04 & $1.00\times$ & 222.35 & $1.00\times$ & 221.19 & $1.00\times$ & 236.52 & $1.00\times$ & 273.19 & $1.00\times$ & 359.40 & $1.00\times$ \\
faissgpu & 27.52 & $7.73\times$ & 25.74 & $8.21\times$ & 25.28 & $8.43\times$ & 25.52 & $8.71\times$ & 26.14 & $8.46\times$ & 27.60 & $8.57\times$ & 30.10 & $9.08\times$ & 37.12 & $9.68\times$ \\
symwarp & \textbf{11.76} & $18.08\times$ & \textbf{9.86} & $21.43\times$ & \textbf{9.31} & $22.88\times$ & \textbf{9.48} & $23.45\times$ & \textbf{10.04} & $22.03\times$ & \textbf{10.48} & $22.57\times$ & \textbf{11.74} & $23.27\times$ & \textbf{13.56} & $26.50\times$ \\
\bottomrule
\end{tabular}%
}
\vspace{-1.2mm}
\end{table*}

\begin{table*}[t]
\centering
\caption{Runtime ($T$, in seconds) comparison on the image-patch dataset with $200\mathrm{k}$ instances and $32 \times 32$ dimensions. The number of clusters is fixed at $c=64$, while the number of nearest neighbors $k$ is varied. Speedups (SUP) are computed relative to \texttt{knncuda}. The fastest runtime per column is shown in bold.}
\label{tab:patch32x32-c64-runtime-speedup}
\setlength{\tabcolsep}{2.2pt}
\renewcommand{\arraystretch}{0.88}
\footnotesize
\resizebox{\textwidth}{!}{%
\begin{tabular}{l rr rr rr rr rr rr rr rr rr}
\toprule
\multirow{2}{*}{Algorithm} & \multicolumn{2}{c}{$k=5$} & \multicolumn{2}{c}{$k=25$} & \multicolumn{2}{c}{$k=50$} & \multicolumn{2}{c}{$k=100$} & \multicolumn{2}{c}{$k=200$} & \multicolumn{2}{c}{$k=300$} & \multicolumn{2}{c}{$k=400$} & \multicolumn{2}{c}{$k=500$} & \multicolumn{2}{c}{$k=600$} \\
\cmidrule(lr){2-3} \cmidrule(lr){4-5} \cmidrule(lr){6-7} \cmidrule(lr){8-9} \cmidrule(lr){10-11} \cmidrule(lr){12-13} \cmidrule(lr){14-15} \cmidrule(lr){16-17} \cmidrule(lr){18-19}
& T & SUP & T & SUP & T & SUP & T & SUP & T & SUP & T & SUP & T & SUP & T & SUP & T & SUP \\
\midrule
\texttt{knncuda} & 46.19 & $1.00\times$ & 46.51 & $1.00\times$ & 46.65 & $1.00\times$ & 47.48 & $1.00\times$ & 49.50 & $1.00\times$ & 52.06 & $1.00\times$ & 54.40 & $1.00\times$ & 57.36 & $1.00\times$ & 60.06 & $1.00\times$ \\
\texttt{faissgpu} & 6.72 & $6.87\times$ & 6.96 & $6.68\times$ & 7.15 & $6.52\times$ & 7.48 & $6.35\times$ & 8.25 & $6.00\times$ & 8.94 & $5.82\times$ & 9.65 & $5.64\times$ & 10.27 & $5.59\times$ & 11.13 & $5.40\times$ \\
\texttt{symwarp} & \textbf{2.17} & $21.29\times$ & \textbf{2.37} & $19.62\times$ & \textbf{2.56} & $18.22\times$ & \textbf{2.86} & $16.60\times$ & \textbf{3.40} & $14.56\times$ & \textbf{4.13} & $12.61\times$ & \textbf{4.78} & $11.38\times$ & \textbf{5.27} & $10.88\times$ & \textbf{5.96} & $10.08\times$ \\
\bottomrule
\end{tabular}%
}
\vspace{-1.2mm}
\end{table*}

\begin{table*}[t]
\centering
\caption{Runtime ($T$, in seconds) Comparison on Synthetic Dataset (500K instances, 128 dimensions) with Number of Clusters Fixed at $c=30$, Varying Number of Nearest Neighbors $k$. Speedup (SUP) is computed relative to the \textit{knncuda} baseline. The fastest runtime per column is shown in bold.}
\label{tab:syn-500k-128d-c30-runtime-speedup}
\setlength{\tabcolsep}{2.2pt}
\renewcommand{\arraystretch}{0.88}
\footnotesize
\resizebox{\textwidth}{!}{%
\begin{tabular}{l rr rr rr rr rr rr rr}
\toprule
\multirow{2}{*}{Algorithm}
& \multicolumn{2}{c}{$k=10$}
& \multicolumn{2}{c}{$k=20$}
& \multicolumn{2}{c}{$k=40$}
& \multicolumn{2}{c}{$k=80$}
& \multicolumn{2}{c}{$k=160$}
& \multicolumn{2}{c}{$k=320$}
& \multicolumn{2}{c}{$k=640$} \\
\cmidrule(lr){2-3}\cmidrule(lr){4-5}\cmidrule(lr){6-7}\cmidrule(lr){8-9}\cmidrule(lr){10-11}\cmidrule(lr){12-13}\cmidrule(lr){14-15}
& T & SUP & T & SUP & T & SUP & T & SUP & T & SUP & T & SUP & T & SUP \\
\midrule
knncuda  & 38.55 & $1.00\times$ & 41.87 & $1.00\times$ & 54.69 & $1.00\times$ & 52.37 & $1.00\times$ & 72.14 & $1.00\times$ & 124.11 & $1.00\times$ & 236.21 & $1.00\times$ \\
faissgpu & 6.87  & $5.61\times$ & 7.08  & $5.91\times$ & 7.27  & $7.52\times$ & 8.01  & $6.54\times$ & 9.42  & $7.66\times$ & 13.42  & $9.25\times$ & 23.04  & $10.25\times$ \\
symwarp  & \textbf{3.21} & $12.01\times$ & \textbf{3.36} & $12.46\times$ & \textbf{3.59} & $15.23\times$ & \textbf{4.21} & $12.44\times$ & \textbf{4.83} & $14.94\times$ & \textbf{7.38} & $16.82\times$ & \textbf{12.45} & $18.97\times$ \\
\bottomrule
\end{tabular}%
}
\vspace{-1.2mm}
\end{table*}

\subsubsection{Evaluation on the Impact of the Number of Clusters} 

\begin{figure}[t]
\centering
\begin{minipage}[t]{.48\textwidth}
  \centering
  \includegraphics[width=1\linewidth]{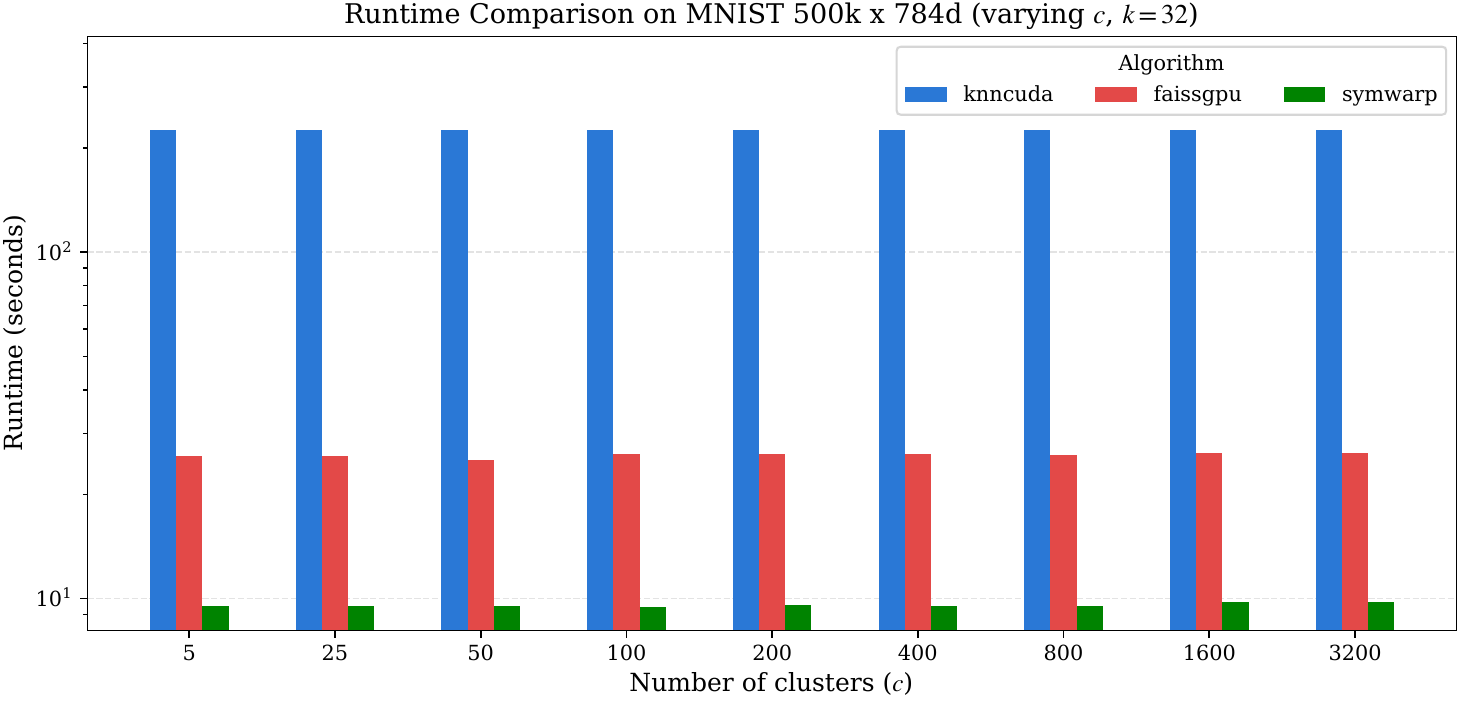}
  \caption{Impact of the number of clusters $c$ on the runtime of \textit{knncuda}, \textit{faissgpu}, and \textit{symwarp} on the MNIST dataset (500K instances, 784 dimensions) with the number of nearest neighbors fixed at $k=32$. Runtime is reported on a logarithmic scale.}
  \label{fig:mnist-various-c-runtime}
\end{minipage}%
\hfill
\begin{minipage}[t]{.48\textwidth}
  \centering
  \includegraphics[width=1\linewidth]{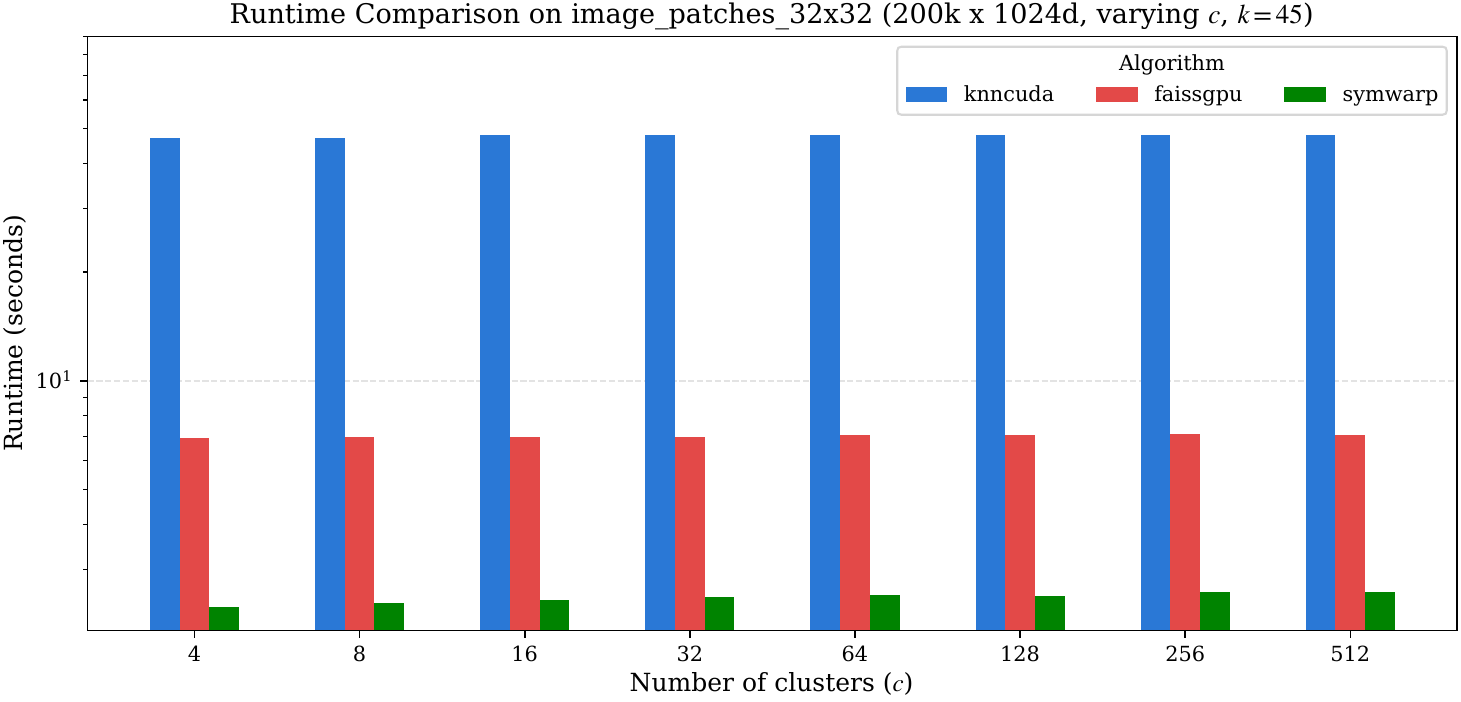}
  \caption{Impact of the number of clusters $c$ on the runtime of \textit{knncuda}, \textit{faissgpu}, and \textit{symwarp} on the 32$\times$32 image-patch dataset (200K instances, 1024 dimensions) with the number of nearest neighbors fixed at $k=45$. Runtime is reported on a logarithmic scale.}
  \label{fig:patch32x32-various-c-runtime}
\end{minipage}
\end{figure}

\begin{figure*}[t]
\centering
\vspace{-1.0mm}
\includegraphics[width=0.98\textwidth]{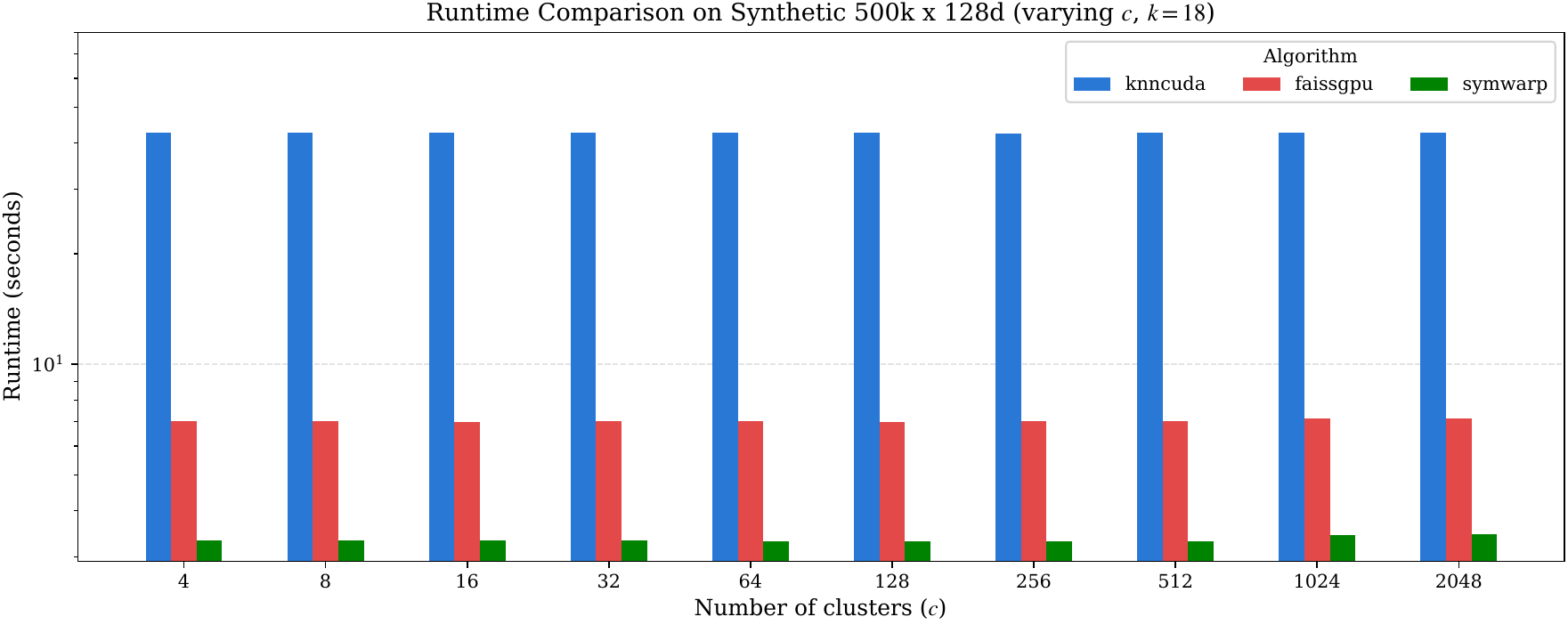}
\vspace{-1.2mm}
\caption{Impact of the number of clusters $c$ on the runtime of \textit{knncuda}, \textit{faissgpu}, and \textit{symwarp} on the synthetic dataset (500K instances, 128 dimensions) with the number of nearest neighbors fixed at $k=18$. Runtime is reported on a logarithmic scale.}
\label{fig:synth-various-c-runtime}
\vspace{-1.5mm}
\end{figure*}

Across four benchmark datasets---GloVe, MNIST, image-patch, and synthetic (Tables~\ref{tab:glove-various-c-speedup}--\ref{tab:syn500k-128d-c-runtime-speedup}, Figures~\ref{fig:glove-various-c-runtime}--\ref{fig:synth-various-c-runtime})---varying the target number of clusters $c$ over a wide range (up to $512$--$3200$, depending on the dataset) has minimal impact on runtime for any evaluated method. \textit{symwarp} remains fastest throughout, with its speedup over \textit{knncuda} staying within a narrow band (roughly $17$--$19\times$ on GloVe, $23$--$24\times$ on MNIST, $18$--$20\times$ on image-patch datasets, and $12$--$13\times$ on synthetic datasets) and a similarly stable $2.1$--$2.9\times$ margin over \textit{faissgpu} across all cluster settings. This near-flat runtime profile is consistent with the algorithm design, which reuses the $k$-NN graph constructed during the initial density estimation pass when additional DCs are required at larger values of $c$, so increasing the requested clustering resolution adds little computational cost.

\begin{table*}[t]
    \centering
    \caption{Runtime ($T$, in seconds) comparison on the GloVe $1.2\mathrm{M}\times200$d dataset with $k=20$ for varying numbers of clusters ($c$). Speedup (SUP) is reported relative to the \textit{knncuda} baseline. The best runtime for each value of $c$ is shown in bold.}
    \label{tab:glove-various-c-speedup}
    \setlength{\tabcolsep}{2.6pt}
    \renewcommand{\arraystretch}{0.90}
    \footnotesize
    \resizebox{\textwidth}{!}{%
        \begin{tabular}{l rr rr rr rr rr rr rr rr rr}
            \toprule
            \multirow{2}{*}{Algorithm} & \multicolumn{2}{c}{$c{=}5$} & \multicolumn{2}{c}{$c{=}25$} & \multicolumn{2}{c}{$c{=}50$} & \multicolumn{2}{c}{$c{=}100$} & \multicolumn{2}{c}{$c{=}200$} & \multicolumn{2}{c}{$c{=}400$} & \multicolumn{2}{c}{$c{=}800$} & \multicolumn{2}{c}{$c{=}1600$} & \multicolumn{2}{c}{$c{=}3200$} \\
            \cmidrule(lr){2-3} \cmidrule(lr){4-5} \cmidrule(lr){6-7} \cmidrule(lr){8-9} \cmidrule(lr){10-11} \cmidrule(lr){12-13} \cmidrule(lr){14-15} \cmidrule(lr){16-17} \cmidrule(lr){18-19}
            & T & SUP & T & SUP & T & SUP & T & SUP & T & SUP & T & SUP & T & SUP & T & SUP & T & SUP \\
            \midrule
            \textit{knncuda} & 427.90 & $1.00\times$ & 429.05 & $1.00\times$ & 429.33 & $1.00\times$ & 428.75 & $1.00\times$ & 428.75 & $1.00\times$ & 428.91 & $1.00\times$ & 429.16 & $1.00\times$ & 429.06 & $1.00\times$ & 430.02 & $1.00\times$ \\
            \textit{faissgpu} & 54.40 & $7.87\times$ & 54.63 & $7.85\times$ & 54.24 & $7.92\times$ & 54.25 & $7.90\times$ & 54.38 & $7.88\times$ & 54.63 & $7.85\times$ & 55.08 & $7.79\times$ & 54.98 & $7.80\times$ & 56.03 & $7.67\times$ \\
            \textit{symwarp} & \textbf{23.20} & \textbf{$18.44\times$} & \textbf{23.19} & \textbf{$18.50\times$} & \textbf{23.21} & \textbf{$18.50\times$} & \textbf{23.21} & \textbf{$18.47\times$} & \textbf{23.19} & \textbf{$18.49\times$} & \textbf{23.46} & \textbf{$18.28\times$} & \textbf{23.83} & \textbf{$18.01\times$} & \textbf{23.83} & \textbf{$18.01\times$} & \textbf{24.89} & \textbf{$17.28\times$} \\
            \bottomrule
        \end{tabular}%
    }
    \vspace{-1.2mm}
\end{table*}

\begin{table*}[t]
    \centering
    \caption{Runtime ($T$, in seconds) comparison on the MNIST $500\mathrm{k}\times784$d dataset with $k=32$ for varying numbers of clusters ($c$). Speedup (SUP) is reported relative to the \textit{knncuda} baseline. The best runtime for each value of $c$ is shown in bold.}
    \label{tab:mnist-500k-c-speedup}
    \setlength{\tabcolsep}{2.6pt}
    \renewcommand{\arraystretch}{0.90}
    \footnotesize
    \resizebox{\textwidth}{!}{%
        \begin{tabular}{l rr rr rr rr rr rr rr rr rr}
            \toprule
            \multirow{2}{*}{Algorithm} & \multicolumn{2}{c}{$c{=}5$} & \multicolumn{2}{c}{$c{=}25$} & \multicolumn{2}{c}{$c{=}50$} & \multicolumn{2}{c}{$c{=}100$} & \multicolumn{2}{c}{$c{=}200$} & \multicolumn{2}{c}{$c{=}400$} & \multicolumn{2}{c}{$c{=}800$} & \multicolumn{2}{c}{$c{=}1600$} & \multicolumn{2}{c}{$c{=}3200$} \\
            \cmidrule(lr){2-3} \cmidrule(lr){4-5} \cmidrule(lr){6-7} \cmidrule(lr){8-9} \cmidrule(lr){10-11} \cmidrule(lr){12-13} \cmidrule(lr){14-15} \cmidrule(lr){16-17} \cmidrule(lr){18-19}
            & T & SUP & T & SUP & T & SUP & T & SUP & T & SUP & T & SUP & T & SUP & T & SUP & T & SUP \\
            \midrule
            \textit{knncuda} & 225.11 & $1.00\times$ & 225.04 & $1.00\times$ & 225.19 & $1.00\times$ & 225.01 & $1.00\times$ & 225.02 & $1.00\times$ & 224.94 & $1.00\times$ & 225.21 & $1.00\times$ & 225.21 & $1.00\times$ & 225.23 & $1.00\times$ \\
            \textit{faissgpu} & 25.82 & $8.72\times$ & 25.79 & $8.73\times$ & 25.15 & $8.95\times$ & 26.14 & $8.61\times$ & 26.09 & $8.62\times$ & 26.13 & $8.61\times$ & 26.03 & $8.65\times$ & 26.36 & $8.54\times$ & 26.34 & $8.55\times$ \\
            \textit{symwarp} & \textbf{9.51} & \textbf{$23.67\times$} & \textbf{9.52} & \textbf{$23.64\times$} & \textbf{9.49} & \textbf{$23.73\times$} & \textbf{9.46} & \textbf{$23.79\times$} & \textbf{9.58} & \textbf{$23.49\times$} & \textbf{9.52} & \textbf{$23.63\times$} & \textbf{9.52} & \textbf{$23.66\times$} & \textbf{9.76} & \textbf{$23.07\times$} & \textbf{9.78} & \textbf{$23.03\times$} \\
            \bottomrule
        \end{tabular}%
    }
    \vspace{-1.2mm}
\end{table*}

\begin{table*}[t]
    \centering
    \caption{Runtime ($T$, in seconds) comparison on the $32\times32$ image-patch dataset ($200\mathrm{k}$ instances, 1024 dimensions) with $k=45$ for varying numbers of clusters ($c$). Speedup (SUP) is reported relative to the \textit{knncuda} baseline. The best runtime for each value of $c$ is shown in bold.}
    \label{tab:image-patches-runtime-speedup}
    \setlength{\tabcolsep}{1.9pt}
    \renewcommand{\arraystretch}{0.88}
    \footnotesize 
    \resizebox{\textwidth}{!}{%
        \begin{tabular}{l rr rr rr rr rr rr rr rr}
            \toprule
            \multirow{2}{*}{Algorithm} & \multicolumn{2}{c}{$c{=}4$} & \multicolumn{2}{c}{$c{=}8$} & \multicolumn{2}{c}{$c{=}16$} & \multicolumn{2}{c}{$c{=}32$} & \multicolumn{2}{c}{$c{=}64$} & \multicolumn{2}{c}{$c{=}128$} & \multicolumn{2}{c}{$c{=}256$} & \multicolumn{2}{c}{$c{=}512$} \\
            \cmidrule(lr){2-3} \cmidrule(lr){4-5} \cmidrule(lr){6-7} \cmidrule(lr){8-9} \cmidrule(lr){10-11} \cmidrule(lr){12-13} \cmidrule(lr){14-15} \cmidrule(lr){16-17}
            & T & SUP & T & SUP & T & SUP & T & SUP & T & SUP & T & SUP & T & SUP & T & SUP \\
            \midrule
            \textit{knncuda} & 46.95 & $1.00\times$ & 47.07 & $1.00\times$ & 48.02 & $1.00\times$ & 48.00 & $1.00\times$ & 47.98 & $1.00\times$ & 47.99 & $1.00\times$ & 48.01 & $1.00\times$ & 48.03 & $1.00\times$ \\
            \textit{faissgpu} & 6.92 & $6.78\times$ & 6.96 & $6.76\times$ & 6.97 & $6.89\times$ & 6.97 & $6.89\times$ & 7.05 & $6.81\times$ & 7.06 & $6.80\times$ & 7.10 & $6.76\times$ & 7.09 & $6.77\times$ \\
            \textit{symwarp} & \textbf{2.36} & \textbf{$19.89\times$} & \textbf{2.42} & \textbf{$19.45\times$} & \textbf{2.46} & \textbf{$19.52\times$} & \textbf{2.51} & \textbf{$19.12\times$} & \textbf{2.54} & \textbf{$18.89\times$} & \textbf{2.53} & \textbf{$18.97\times$} & \textbf{2.60} & \textbf{$18.47\times$} & \textbf{2.60} & \textbf{$18.47\times$} \\
            \bottomrule
        \end{tabular}%
    }
    \vspace{-1.2mm}
\end{table*}

\begin{table*}[t]
    \centering
    \caption{Runtime ($T$, in seconds) comparison on the synthetic $500\mathrm{k}\times128$d dataset with $k=18$ for varying numbers of clusters ($c$). Speedup (SUP) is reported relative to the \textit{knncuda} baseline. The best runtime for each value of $c$ is shown in bold.}
    \label{tab:syn500k-128d-c-runtime-speedup}
    \setlength{\tabcolsep}{1.9pt}
    \renewcommand{\arraystretch}{0.88}
    \footnotesize 
    \resizebox{\textwidth}{!}{%
        \begin{tabular}{l *{10}{rr}}
            \toprule
            \multirow{2}{*}{Algorithm} & \multicolumn{2}{c}{$c{=}4$} & \multicolumn{2}{c}{$c{=}8$} & \multicolumn{2}{c}{$c{=}16$} & \multicolumn{2}{c}{$c{=}32$} & \multicolumn{2}{c}{$c{=}64$} & \multicolumn{2}{c}{$c{=}128$} & \multicolumn{2}{c}{$c{=}256$} & \multicolumn{2}{c}{$c{=}512$} & \multicolumn{2}{c}{$c{=}1024$} & \multicolumn{2}{c}{$c{=}2048$} \\
            \cmidrule(lr){2-3} \cmidrule(lr){4-5} \cmidrule(lr){6-7} \cmidrule(lr){8-9} \cmidrule(lr){10-11} \cmidrule(lr){12-13} \cmidrule(lr){14-15} \cmidrule(lr){16-17} \cmidrule(lr){18-19} \cmidrule(lr){20-21}
            & T & SUP & T & SUP & T & SUP & T & SUP & T & SUP & T & SUP & T & SUP & T & SUP & T & SUP & T & SUP \\
            \midrule
            \textit{knncuda} & 42.65 & $1.00\times$ & 42.61 & $1.00\times$ & 42.57 & $1.00\times$ & 42.61 & $1.00\times$ & 42.61 & $1.00\times$ & 42.59 & $1.00\times$ & 42.56 & $1.00\times$ & 42.58 & $1.00\times$ & 42.62 & $1.00\times$ & 42.63 & $1.00\times$ \\
            \textit{faissgpu} & 7.02 & $6.08\times$ & 7.00 & $6.09\times$ & 6.99 & $6.09\times$ & 7.00 & $6.09\times$ & 7.01 & $6.08\times$ & 6.99 & $6.09\times$ & 7.00 & $6.08\times$ & 7.01 & $6.07\times$ & 7.14 & $5.97\times$ & 7.13 & $5.98\times$ \\
            \textit{symwarp} & \textbf{3.33} & \textbf{$12.81\times$} & \textbf{3.32} & \textbf{$12.83\times$} & \textbf{3.32} & \textbf{$12.82\times$} & \textbf{3.32} & \textbf{$12.83\times$} & \textbf{3.31} & \textbf{$12.87\times$} & \textbf{3.31} & \textbf{$12.87\times$} & \textbf{3.31} & \textbf{$12.86\times$} & \textbf{3.31} & \textbf{$12.86\times$} & \textbf{3.44} & \textbf{$12.39\times$} & \textbf{3.45} & \textbf{$12.36\times$} \\
            \bottomrule
        \end{tabular}%
    }
    \vspace{-1.2mm}
\end{table*}

\subsection{Output Equivalence Validation}
\label{sec:output-equivalence-validation}

Section~\ref{sec:tie-break} established that the top-$k$ tie-break in SymTri-$k$NN is resolved deterministically and quantified the prevalence of exact-distance ties in practice. However, that analysis considered only the $k$-NN layer in isolation. This section evaluates the extent to which \textit{SymWarp-FastDEC} reproduces the outputs of the original CPU reference implementation, \textit{FastDEC-pycpu}~\citep{yang2022fastdec}. Specifically, we evaluate agreement between the two pipelines at every stage of the clustering process, including neighbor-set construction, DC detection, cluster-center selection, and final cluster assignment.

Both pipelines were exercised through their own implementations, with the two corrections described at the end of this section applied to both beforehand, on seven datasets (nine dataset/configuration combinations) spanning 240--20{,}000 points and 2--784 dimensions. The evaluation includes three configurations of a real, quantized-coordinate dataset (MNIST), two of them at $k=256$, chosen to exercise the exact-tie risk identified in Section~\ref{sec:tie-break}. For each configuration, four artifacts were compared: the raw self-$k$-NN neighbor sets, the detected DC sets, the selected cluster-center point indices, and the final per-point cluster assignments. $k$-NN agreement is the percentage of points whose unordered top-$k$ neighbor set is identical in both pipelines; a single differing neighbor counts as a disagreement, making this measure stricter than mean set overlap. DC and center agreement are counts of exactly matched point indices, and partition agreement is measured by the Adjusted Rand Index (ARI)~\citep{hubert1985ari}. Because \textit{FastDEC-pycpu} becomes prohibitively expensive at larger scales, the equivalence study is restricted to smaller datasets than those used in the throughput evaluation.

\begin{table*}[t]
    \centering
    \caption{Output agreement between \textit{SymWarp-FastDEC} and \textit{FastDEC-pycpu} across nine dataset configurations. $N$ denotes the number of data points, $k$ the number of nearest neighbors per point, and $c$ the requested number of clusters. All ARI values are exactly 1 (identical partitions). $^{\dagger}$The number of DCs equals $c$, so every DC becomes a center and the significance ranking is not exercised.}
    \label{tab:output-equivalence}
    \setlength{\tabcolsep}{4pt}
    \renewcommand{\arraystretch}{0.90}
    \footnotesize 
    \resizebox{\textwidth}{!}{%
        \begin{tabular}{l rrrrrrr}
            \toprule
            Dataset & $N$ & $k$ & $c$ & $k$NN agreement (\%) & DCs matched & Centers matched & ARI \\
            \midrule
            Spiral & 312 & 7 & 3 & 100.00 & 7/7 & 3/3 & 1 \\
            Flame & 240 & 5 & 2 & 99.58 & 20/20 & 2/2 & 1 \\
            banana-ball & 2{,}000 & 11 & 3 & 99.95 & 73/73 & 3/3 & 1 \\
            R15 & 600 & 33 & 15 & 100.00 & 15/15 & 15/15 & 1 \\
            S2 & 5{,}000 & 31 & 15 & 99.30 & 17/18 & 14/15 & 1 \\
            MFCCs & 7{,}195 & 25 & 10 & 99.68 & 46/46 & 10/10 & 1 \\
            MNIST (5{,}000-row subset) & 5{,}000 & 32 & 10 & 99.98 & 13/13 & 10/10 & 1 \\
            MNIST (5{,}000-row subset)$^{\dagger}$ & 5{,}000 & 256 & 2 & 100.00 & 2/2 & 2/2 & 1 \\
            MNIST (20{,}000-row subset)$^{\dagger}$ & 20{,}000 & 256 & 9 & 99.94 & 9/9 & 9/9 & 1 \\
            \bottomrule
        \end{tabular}%
    }
    \vspace{-1.2mm}
\end{table*}

All nine configurations yielded identical final clustering partitions between \textit{SymWarp-FastDEC} and \textit{FastDEC-pycpu} (ARI $=1$ exactly), despite small differences in individual neighbor sets. $k$-NN agreement is no lower than $99.30\%$ on any configuration, and DC and center indices agree exactly everywhere except on S2. There, one DC is replaced by a neighboring point, which also changes one center, because a near-degenerate density comparison is ordered differently under FP32 and FP64 arithmetic rather than because of an exact-distance tie. The two candidate points lie on the same local density peak, and the discrepancy does not propagate to the final clustering result.

The equivalence study also uncovered two implementation defects shared by both \textit{FastDEC-pycpu} and \textit{SymWarp-FastDEC}, neither of which is related to tie breaking. First, the significance-score normalization used to rank DCs can degenerate. When too few DCs exist for the sentinel fill-in convention (Section~\ref{sub:dc-reduction}) to produce a distinct value, min-max normalization divides by a zero range, propagating NaN into the score and making center selection effectively arbitrary. In rare cases, this behavior can exclude the true global density maximum from the selected centers.

Second, a point with no CDH point in the dataset (i.e., the global density maximum) but not selected as a center leaves $CDH\_id[i]=-1$, which a naive implementation may incorrectly dereference during label propagation. Both implementations now treat a zero-range normalization factor as neutral rather than undefined and explicitly guard the $CDH\_id[i]=-1$ case during cluster assignment (Algorithm~\ref{alg:cluster_assignment}). Both fixes were applied symmetrically to the CPU and GPU implementations. Neither fix's code path is reached, on either side, in any configuration of Table~\ref{tab:output-equivalence} or in any other experiment reported here, so the unpatched original implementation produces the same outputs.

\subsection{Reduced-Precision (TF32) Option}
\label{sec:tf32-option}

All other results in this paper use strict FP32 arithmetic (Section~\ref{sec:exp-config}). Because the dominant $D$-dependent work in both SymTri-$k$NN and WP-DC is performed by cuBLAS SGEMM, \textit{SymWarp-FastDEC} can optionally enable TF32 tensor-core math for these calls. TF32 rounds GEMM inputs to 10 explicit fraction bits while accumulating in FP32; data storage, squared norms, and distance outputs remain FP32, and the search remains exhaustive. TF32 therefore changes the arithmetic used to rank candidates, not the set of candidate pairs examined. Because the norm-expansion form of the squared Euclidean distance is prone to cancellation in low dimensions, the implementation keeps FP32 when $D<8$; this guard is a safeguard rather than an accuracy guarantee for larger $D$.

Table~\ref{tab:tf32-option} reports the fit-time speedup of TF32 over FP32, measured with the protocol of Section~\ref{sec:exp-config}. It also reports the adjusted Rand index (ARI) between the complete FP32 and TF32 label vectors, obtained from separate label-export runs that reproduced the DC counts and ordered center lists of the timed runs. As expected, the benefit grows with the share of time spent in GEMM. On the fixed-size synthetic sweep, the speedup rises from $1.00\times$ at 16d to $2.23\times$ at 4096d, and the 64x64 patches and syn\_2m\_1536d reach $1.89\times$ and $1.96\times$, respectively. In contrast, deep\_2m\_96d and syn\_10m\_64d gain only $1.16\times$ and $1.13\times$, showing that a large $N$ alone does not imply a large TF32 benefit. On mnist\_70k\_2d, the dimensionality guard applies, so both builds execute the same FP32 path and produce identical partitions.

\begin{table}[t]
    \centering
    \caption{TF32 versus FP32 \textit{SymWarp-FastDEC}: fit-time speedup $S_{\mathrm{fit}}=T^{\mathrm{FP32}}/T^{\mathrm{TF32}}$ and ARI between the full FP32 and TF32 partitions. ARI compares the two precisions with each other, not with ground-truth labels. Only syn\_500k\_16d yields an identical partition.}
    \label{tab:tf32-option}
    \footnotesize
    \begin{tabular*}{0.85\textwidth}{@{\extracolsep{\fill}}lcrr@{}}
        \toprule
        Dataset & $(k,c)$ & $S_{\mathrm{fit}}$ & ARI \\
        \midrule
        mnist\_70k\_784d  & $(25,10)$ & $1.36\times$ & 0.999975 \\
        mnist\_500k\_784d & $(25,10)$ & $1.79\times$ & 0.986866 \\
        deep\_2m\_96d     & $(50,80)$ & $1.16\times$ & 0.996665 \\
        GloVe 200d        & $(45,20)$ & $1.37\times$ & 0.995019 \\
        64x64 patches     & $(25,64)$ & $1.89\times$ & 0.999719 \\
        syn\_10m\_64d     & $(32,25)$ & $1.13\times$ & 0.728461 \\
        syn\_2m\_1536d    & $(32,25)$ & $1.96\times$ & 0.971013 \\
        \midrule
        syn\_500k\_16d    & $(32,20)$ & $1.00\times$ & 1 \\
        syn\_500k\_128d   & $(32,20)$ & $1.22\times$ & 0.914592 \\
        syn\_500k\_512d   & $(32,20)$ & $1.63\times$ & 0.954741 \\
        syn\_500k\_4096d  & $(32,20)$ & $2.23\times$ & 0.949034 \\
        \bottomrule
    \end{tabular*}
\end{table}

TF32 does not preserve the FP32 output. Reduced-precision inner products perturb near-equidistant neighbor orderings and density comparisons, and these perturbations propagate to DC detection and label assignment. On the real-world datasets, agreement remains high (ARI $0.987$--$0.99998$), but no partition is identical, even where the selected centers coincide (mnist\_70k\_784d, 64x64 patches). The synthetic datasets are considerably more sensitive, with ARI as low as $0.73$ on syn\_10m\_64d despite all 25 centers being retained. This is consistent with their uniform distribution (Section~\ref{sec:datasets}): without natural cluster structure, partition boundaries are determined by small density differences. Agreement also does not increase monotonically with dimensionality. Across the 116 benchmark configurations timed in both precisions, only 70 select the same set of centers.

We therefore keep FP32 as the default configuration, under which the equivalence results of Section~\ref{sec:output-equivalence-validation} hold. TF32 is a separate, opt-in throughput mode that is most useful for high-dimensional workloads; when it is used, partition stability and clustering quality should be validated for the target data, and FP32 quality results should not be assumed to carry over.

\section{Conclusions and Future Work}
\label{sec:conclusions}

In this paper, we presented SymWarp-FastDEC\footnote{SymWarp-FastDEC is released at \url{https://github.com/cyberbull66/symwarp-fastdec}.}, a GPU-accelerated re-architecture of FastDEC that accelerates its two dominant stages: exact self-$k$-NN-based density estimation and dominance-component (DC) reduction. The key enablers are (i) SymTri-$k$NN, an index-free exact GPU self-$k$-NN algorithm that exploits the symmetry of the self-query setting via upper-triangular traversal and warp-cooperative top-$k$ maintenance, and (ii) WP-DC, a warp-parallel DC reduction kernel that decouples DC reduction from any auxiliary index or graph structure. Across diverse real-world and synthetic benchmarks, SymWarp-FastDEC achieves approximately $1.7\times$--$4.9\times$ end-to-end speedup over a FastDEC implementation employing FAISS GpuIndexFlatL2 for exact $k$-NN search. Under the FP32 configuration and equivalence protocol used in this study, it produces clustering outcomes identical to the original FastDEC (ARI $=1$ exactly on all equivalence benchmarks). Consequently, it can serve as a drop-in replacement (same interface and parameters) for the original FastDEC in that setting. This end-to-end advantage over the pipeline built on the heavily optimized and widely deployed exact GPU $k$-NN baseline FAISS GpuIndexFlatL2 holds across all evaluated benchmarks and generally grows with dimensionality. All experiments were conducted on a single NVIDIA RTX A6000; while the design principles are architecture-independent, the relative performance against FAISS may vary across GPU generations.

Beyond performance, the proposed design strengthens FastDEC's practicality on modern high-dimensional embeddings by preserving full-dimensional vectors and avoiding lossy dimensionality reduction. Furthermore, because SymTri-$k$NN is index-free, its performance is largely unaffected by the degradation that many search indexes experience in high-dimensional spaces while still providing exact results. The resulting pipeline provides an efficient, exact, and scalable baseline for large-scale density-peak clustering, extending the practical applicability of exact density-peak methods to substantially larger datasets in domains such as representation learning, semantic search, image analysis, anomaly detection, and data mining.

Future work includes: (1) extending the single-GPU design to multi-GPU execution for the $\Theta(N^2)$ density-estimation stage, including overlap of computation and communication via CUDA streams; (2) reducing GPU-memory pressure through more compact neighbor and metadata layouts, as well as out-of-core or chunked execution for very large $N$; (3) broadening support for additional distance and similarity measures and reduced-precision modes, building on the TF32 option of Section~\ref{sec:tf32-option} with workload-specific validation of partition stability, while preserving FastDEC's exactness guarantees where required; (4) enhancing the pipeline to better support streaming or continually growing datasets, including incremental updates to neighborhood structures and cluster assignments under bounded memory; (5) leveraging SymTri-$k$NN and WP-DC as reusable building blocks for accelerating other self-similarity and density-based analytics pipelines that share the same computational structure; and (6) evaluating SymWarp-FastDEC across additional GPU architectures and generations.

\backmatter

\section*{Declarations}

\bmhead{Funding}
This research was supported by the University of Waikato.

\bmhead{Competing interests}
The authors have no relevant financial or non-financial interests to disclose.

\bmhead{Data availability}
The MNIST, MNIST8M, GloVe, and Deep1B datasets analyzed in this study are publicly available from the sources cited in Section~\ref{sec:datasets}; the subsets used in our experiments were sampled from these datasets. The image-patch datasets and the synthetic datasets were generated as described in Section~\ref{sec:datasets}. The generated datasets, the sampled subsets, and the raw runtime measurements supporting the results of this study are available from the corresponding author on reasonable request. The source images from which the image patches were extracted are publicly available from the Open Images Dataset \citep{kuznetsova2020open}.

\bmhead{Code availability}
The source code of SymWarp-FastDEC, together with the FastDEC-knncuda and FastDEC-faissgpu baseline implementations used in our experiments, is publicly available at \url{https://github.com/cyberbull66/symwarp-fastdec}.

\bibliography{sn-bibliography} 

\end{document}